\documentclass[journal]{IEEEtran}
\ifCLASSINFOpdf
\else
\fi
\usepackage{amsmath}
\usepackage{algorithmic}

\usepackage{array}

\usepackage{microtype}                 
\PassOptionsToPackage{warn}{textcomp}  
\usepackage{textcomp}                  
\usepackage{mathptmx}                  
\usepackage{times}                     
\usepackage{cite}                      
\usepackage{tabu}                      
\usepackage{booktabs}                  
\usepackage{subfig}
\usepackage{times}
\usepackage{epsfig}
\usepackage{graphicx}
\usepackage{amsmath}
\usepackage{amssymb}
\usepackage{algorithm}
\usepackage{color}
\usepackage{algorithmic}
\usepackage{bbm}
\usepackage{bm}
\usepackage{multirow}
\usepackage{comment}
\usepackage{picins}
\usepackage[table]{xcolor}
\usepackage[switch]{lineno}
\newcommand\norm[1]{\left\lVert#1\right\rVert}
\definecolor{myc1}{gray}{0.8}

\newcommand{\gu}[1]{{\textcolor{black}{#1}}}
\newcommand{\guSec}[1]{{\textcolor{black}{#1}}}
\newcommand{\guThi}[1]{{\textcolor{black}{#1}}}

\usepackage[breaklinks=true,bookmarks=false]{hyperref}

\begin{document}
%
\title{\gu{Blur Removal via Blurred-Noisy Image Pair}}
%
%
%

\author{Chunzhi Gu,
        Xuequan Lu,~\IEEEmembership{Member,~IEEE,}
        Ying He,~\IEEEmembership{Member,~IEEE,}
        and~Chao~Zhang,~\IEEEmembership{Member,~IEEE}
\thanks{C. Gu and C. Zhang* are with the School of Engineering, University of Fukui, Fukui, Japan (e-mails: gu-cz@monju.fuis.u-fukui.ac.jp; zhang@u-fukui.ac.jp).}
\thanks{X. Lu is with School of Information Techonolgy, Deakin University, Geelong, VIC 3216, Australia (e-mail: xuequan.lu@deakin.edu.au).}
\thanks{Y. He is with School of Computer Science and Engineering, Nanyang Technological University, Singapore (e-mail: YHe@ntu.edu.sg).}
}

\maketitle

\begin{abstract}
Complex blur such as the mixup of space-variant and space-invariant blur, which is hard to model mathematically, widely exists in real images. In this paper, we propose a novel image deblurring method that does not need to estimate blur kernels. We utilize a pair of images that can be easily acquired in low-light situations: (1) a blurred image taken with low shutter speed and low ISO noise; and (2) a noisy image captured with high shutter speed and high ISO noise. Slicing the blurred image into patches, we extend the Gaussian mixture model (GMM) to model the underlying intensity distribution of each patch using the corresponding patches in the noisy image. We compute patch correspondences by analyzing the optical flow between the two images. The Expectation Maximization (EM) algorithm is utilized to estimate the parameters of GMM. To preserve sharp features, we add an additional bilateral term to the objective function in the M-step. We eventually add a detail layer to the deblurred image for refinement. Extensive experiments on both synthetic and real-world data demonstrate that our method outperforms state-of-the-art techniques, in terms of robustness, visual quality, and quantitative metrics.
\end{abstract}


%


\section{Introduction}

\IEEEPARstart{I}{t} is prevalent to adopt image deblurring techniques to recover quality images from blurry images. A common situation is capturing photos in dimly-lit environments (e.g., photographing moving objects in a night scene), where one can hardly get sharp and bright photos. Most likely, the taken photos are dark or blurry, depending on the camera settings and object conditions. Though a lower shutter speed can effectively increase brightness, it almost inevitably leads to blur. On the other hand, increasing the shutter speed makes the camera sensor or film exposed to limited light, resulting in dark photos. Setting a high ISO for increasing brightness is a trade-off way to obtain bright photos. Nevertheless, a higher gain setting amplifies noise which may even worsen the photo quality. Recovering quality photos from such captured blurry photos remains challenging and can hardly be resolved by the existing image deblurring techniques. 

\begin{figure}[tb]
\centering
\setcounter{subfigure}{0}
\subfloat{
\includegraphics[width=25mm,scale=0.5]{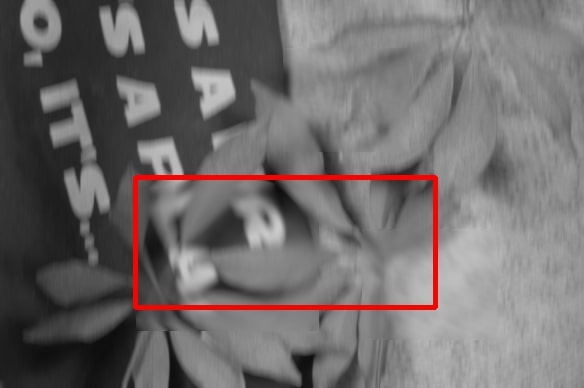}
\includegraphics[width=25mm,scale=0.5]{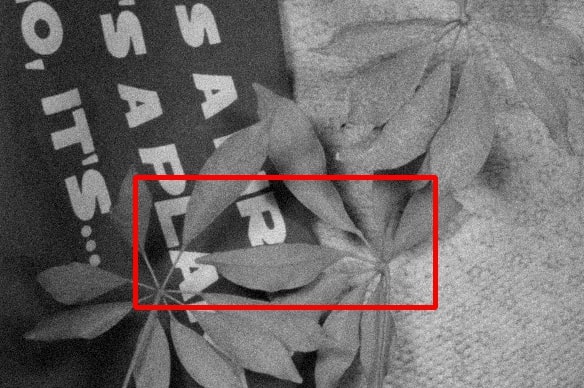}
\includegraphics[width=25mm,scale=0.5]{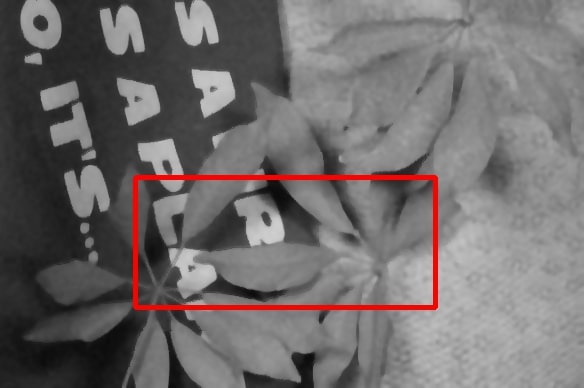}
}\\[0.1ex]

\subfloat{
\includegraphics[width=25mm,scale=0.2]{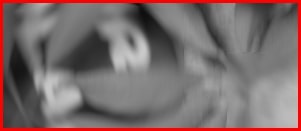}
\includegraphics[width=25mm,scale=0.2]{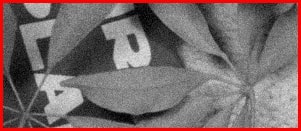}
\includegraphics[width=25mm,scale=0.2]{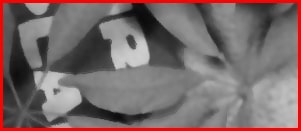}
}

\subfloat{
\includegraphics[width=25mm,scale=0.5]{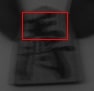}
\includegraphics[width=25mm,scale=0.5]{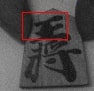}
\includegraphics[width=25mm,scale=0.5]{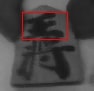}
}\\[-2.1ex]

\setcounter{subfigure}{0}
\hspace{-2.3mm}
\subfloat[Blurred image]{
\includegraphics[width=25mm,scale=0.2]{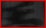}}
\subfloat[Noisy image]{
\includegraphics[width=25mm,scale=0.2]{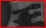}}
\subfloat[Deblurred (a)]{
\includegraphics[width=25mm,scale=0.2]{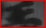}}

\caption{Debluring synthetic (row 1) and real-world (row 2) images. Note that (b) is in a different view from (a) and (c). In the synthetic example: (a) is blurred by a mixup of two types of blur and (b) contains synthetic Gaussian noise ($\sigma = 10$). In the real-world example: (a) is taken with the shutter speed of 1 second and ISO of 100 (both the camera and the object are moving); (b) is taken with the shutter speed of 1/20 second and ISO of 1600, and further enhanced.}
\label{fig:first page}
\end{figure}

\begin{figure*}[t]
\begin{center}
\includegraphics[width=1.0\linewidth]{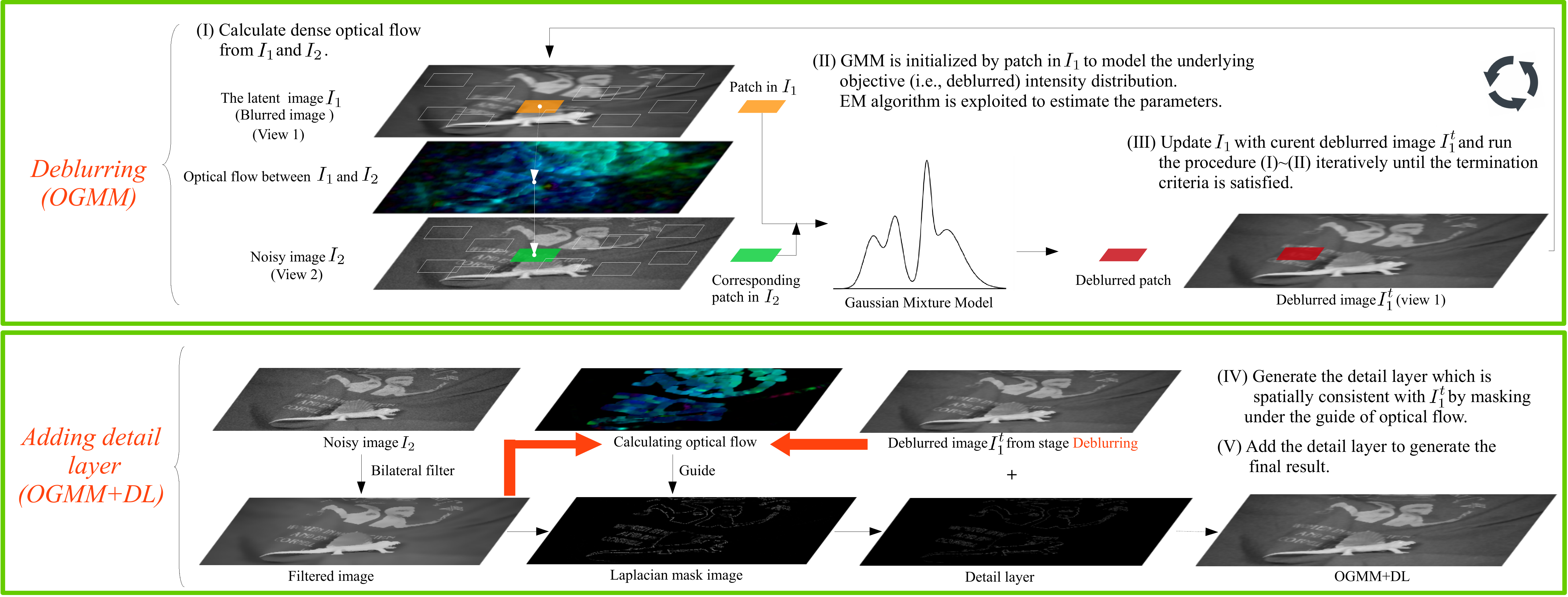}
\end{center}
\caption{Overview of our image deblurring approach.
The \textit{Deblurring} stage is an iterative procedure. The deblurred image ${I_1^{t+1}}$ at the $(t+1)$-th iteration is updated based on the deblurred result ${I_1^{t}}$ at the $t$-th iteration. In the \textit{Adding Detail Layer} stage, the detail layer can be extracted using the Laplacian mask image. The spatial inconsistency between $I_1^t$ and the detail layer is solved by the updated optical flow. }
\label{fig:introduction}
\end{figure*}

Removing blur from blurred images to achieve latent sharp images has been widely studied \cite{wang2014recent,7479956}. Many approaches \cite{joshi2008psf,cho2009fast} estimate the blur kernels using salient features. Such methods may fail when images are not bright enough to get sufficient features such as edges. In fact, it is difficult to model blur in real photos in some cases, because of the mix of different types of blur (i.e., complex blur). As a result, deblurring methods based on blur kernel estimation have limited performance in handling complex blur.  

Compared with a single image, multiple images often show more information that can be utilized for deblurring. In this work, we attempt to exploit a pair of blurred/noisy images which can be easily obtained by changing the shutter speed and ISO settings. The noisy image contains complementary pixel information to the blurred image. Deblurring using a pair of blurred/noisy images has been sparsely treated so far \guSec{\cite{yuan2007image, arun2016hand}, and a close work to ours is Yuan  \textit{et al.}} \cite{yuan2007image} in which the noisy image is used to add details lost in the deblurring process with deconvolution. Despite its convincing deblurring performance, it still suffers from two major issues: (1) the image pair is strictly constrained to be taken from the same view, which is to avoid the misalignment of pixels between two images; and (2) the blur kernel in their paper is supposed to be a single type (linear motion blur) caused by camera shake. These two constraints diminish its practical use, for example, the moving objects in row 2 of Fig. \ref{fig:first page}. As such, their technique has limited performance in the cases without meeting the above assumptions.

To overcome the above limitations, we propose a novel approach for image deblurring, with the easily obtained noisy/blurred image pair. Specifically, we first slice the blurred image (in view 1) into patches and each patch is guided to its corresponding patch in the noisy image (in view 2) by optical flow. To remove the blur and recover the latent sharp image, we extend the Gaussian Mixture Model (GMM) to model the intensity distribution of the latent sharp image, with its parameters estimated by the Expectation-Maximization (EM) algorithm. We further add a bilateral term to the objective function in the M-step of EM, to better preserve sharp features in images. We alternately update the optical flow and perform the EM algorithm for several rounds, to achieve desired deblurring results. Therefore, we refer to our method as optical flow guided GMM (OGMM). To further retain details, we add a detail layer (the denoised sharp features) back to the deblurred image, according to the pixel correspondences from the final estimated optical flow. Our method is free of blur kernels, and it can handle complex blur which is challenging for kernel estimation based techniques (e.g., space-variant blur in the first row of Fig. \ref{fig:first page}). 

The main contributions of this paper are threefold. First, we propose a novel deblurring approach called optical flow guided GMM (OGMM) with a pair of blurred/noisy images as input. Second, we formulate deblurring as a parameter estimation problem, and derive an EM algorithm to optimize the involved parameters. Eventually, a bilateral term is added to the objective function of the M-step in EM to better preserve sharp features, and a detail layer is extracted to enhance the details in the deblurred image. Instead of kernel estimation or deconvolution, we make full use of the noisy image taken in a different view for deblurring.

\section{Related work}

\textbf{Blind deblurring for single image.} Blind deblurring aims to accurately estimate the unknown blur kernel, based on which deconvolution is performed to recover the corresponding sharp image. There are several types of methods for blind deblurring, such as maximum a posterior (MAP)  \cite{krishnan2011blind,xu2010two}, variational Bayes \cite{fergus2006removing,zhang2010denoising,whyte2012non}, edge prediction  \cite{hirsch2010efficient,joshi2008psf} \guSec{and sparse coding \cite{zhang2014group, zha2020benchmark}.} For the MAP based methods, various strategies are presented to cope with the problem revealed by Levin \textit{et al.} \cite{levin2009understanding} that the failure of na\"{i}ve MAP may occur because it favors no-blur explanations. Marginal distributions are considered to be maximized over all possible images \cite{levin2009understanding,fergus2006removing}. Image regularizations are introduced into the MAP framework \cite{krishnan2011blind, xu2013unnatural, shan2008high} to retain salient image structures. The state-of-the-art methods for 
single image deblurring also depend on rich information hidden in the blur. \gu{Bai \textit{et al.} \cite{bai2018graph} used reweighted graph total variation as prior to reconstruct a skeleton patch from the blurry observation. Kheradmand \textit{et al.} \cite{kheradmand2014general} introduced a novel data adaptive objective function to handle challenging motion blur PSFs. \guSec{Zhang \textit{et al.} \cite{zhang2014group} established an unified framework for natural image restoration using group-based sparse representation.}} Yan \textit{et al.} \cite{yan2017image} proposed an image prior named Extreme Channels Prior (ECP) to help the uniform kernel estimation based on the observation that the values of bright channel pixels are likely to decrease. Hu  \textit{et al.} \cite{hu2014deblurring} utilized light streaks in the images taken in low light situations as constraints for estimating the blur model, but it only succeeds when the light streak is large. Single image blind deblurring usually encounters the bottleneck that the useful information for kernel estimation is insufficient, and can hardly output a proper blur model in real cases.
\gu{Recently, deep learning based deblurring methods have also been explored \cite{xu2017motion, nah2017deep, kupyn2018deblurgan}. Xu \textit{et al.} \cite{xu2017motion} emphasized the use of CNN on motion kernel estimation, which performs well mostly on linear motion kernels. To further free the algorithm from the restrictions of blur kernel estimation, Nah et al. \cite{nah2017deep} and Kupyn et al. \cite{kupyn2018deblurgan} proposed the networks that can handle complex blur kernels. }

\textbf{Multiple images deblurring.} Efforts have been made to multiple images deblurring \cite{cai2009blind,hee2014gyro,yuan2007image,zhang2013multi,sroubek2012robust,chen2008robust,zhu2012deconvolving,li2011exploring}. The superiority of deblurring with multiple images lies in the complementary information provided in those images. Hee \textit{et al.} \cite{hee2014gyro} introduced a Gyro-Based method to cope with handshake blur caused by camera motion. Multiple blurred images can provide necessary frequency components which are missing due to blur. However, it can hardly handle object movement. Cai \textit{et al.} \cite{cai2009blind} aligned multiple motion blurred frames accurately and show promising results with their tight framelet system. Li \textit{et al.} \cite{li2011exploring} used two well-aligned blurred images to better estimate the blur kernel.
Zhang \textit{et al.} \cite{zhang2013multi} estimate the latent sharp image with given multiple blurry and/or noisy images by designing a penalty function which can balance the effects of observations with varying quality and avoid local minimal. However, they assume a single type of linear motion blur or uniform blur. \textit{In fact, none of the above approaches can handle complex blur.} \guSec{To pursue better solutions in solving non-uniform blur, other methods \cite{arun2016hand, zhuo2010robust,cho2012registration, zhang2014multi, delbracio2015burst, delbracio2015removing} have been proposed, in which they leveraged different forms of multiple observations, e.g., a blurred-noisy pair  \cite{arun2016hand}, a blurred-flash pair \cite{zhuo2010robust} and multiple blurred images \cite{cho2012registration,zhang2014multi,delbracio2015burst, delbracio2015removing}. Arun \textit{et al.} \cite{arun2016hand} exploited the gradient information on the noisy image to iteratively estimate the latent sharp image, and they seemed to show the results on the data with less intense blurs, according to the provided visual experiments. Cho \textit{et al.} \cite{cho2012registration} proposed a registration based deblurring strategy, with the constraint that the dominant blur directions of the two input blurred images are required to be approximately orthogonal. In essence, \cite{zhuo2010robust,cho2012registration, zhang2014multi, delbracio2015burst, delbracio2015removing} can hardly handle the blur arising from object movement, which are less powerful than our method. } 

\textbf{Patch based GMM framework.} Gaussian mixture model has been widely exploited in image restoration tasks \cite{zha2017image,teodoro2016image2,zoran2011learning,xu2015patch,teodoro2016image,sun2014good},\guThi{\cite{
dong2011image,zha2020image1,zha2020image2}} and point cloud processing tasks \cite{Lu2017,Lu2018,LU2019}. In \cite{zoran2011learning}, Gaussian mixture priors are learned from a set of natural images. By maximizing the expected patch log likelihood, an image without distortion can be reconstructed with priors. The learned patch group Gaussian mixture model (PG-GMM) by Xu \textit{et al.} \cite{xu2015patch}, providing dictionaries and regularization parameters, achieves a high denoising performance. The study by Zoran \textit{et al.} \cite{zoran2012natural} gives a comprehensive analysis that modeling natural images by GMM is effective in log likelihood scores, denoising performance and sample quality. However, GMM based learning methods commonly suffer from huge computational time and a massive dataset. We exploit GMM in a different way, which relates the patches in the noisy image with the patches in the latent image of the blurred image according to dense optical flow. In other words, we attempt to model the intensity distribution in each patch instead of learning patch based image priors to restore images.

\section{Method}
Fig. \ref{fig:introduction} illustrates the overview of our method, which consists of two stages: deblurring and adding detail layer. The latter can be viewed as post-processing or refinement. We first adopt optical flow \cite{farneback2003two} to find the corresponding patches between the blurred image and the noisy image. We then formulate the image deblurring problem under the framework of GMM, and adopt the EM algorithm \cite{dempster1977maximum} to optimize the involved parameters. We further add a bilateral term to the objective function in the M-step, to prevent sharp features being smoothed out. Optical flow update and the EM algorithm are alternately called, to achieve the best deblurring results. Finally, we extract a detail layer from the noisy image and add it to the deblurred image, to better preserve the details.

\subsection{Patch Correspondence}
\label{sec:patchcorrespondence}
The blurred image $I_1$ is decomposed into a set of overlapping square patches $C=\begin{Bmatrix}c_1, ..., c_i, ..., c_P\end{Bmatrix}$, where ${c_i} \in {R^{M}}$ and $M=s_1 \times s_1$. $P$ is the number of the patches, and $s_1$ denotes the patch size in $I_1$, and $M$ is the number of pixels in each patch. The \gu{multiset} of pixel intensities in an arbitrary patch from $I_1$ is denoted as $X$ ($X \in R^M$), and $x_m$ denotes \guSec{the $m$-th} pixel intensity in $X$. We \guSec{adopt} the dense optical flow (DOF) \cite{farneback2003two} to find $c_i$'s corresponding patch $d_j$ in the noisy image $I_2$. Note that for brighter and clearer visualization purposes, in the case of real images, brightness and contrast of $I_2$ are obtained by adjusting gain, bias, and gamma correction parameters. Here, patch $c_i$ has correspondence to patch $d_j$ if the two center pixels of $c_i$ and $d_j$ are connected with respect to the DOF field. The set of corresponding patches in $I_2$ can then be denoted as $D=\begin{Bmatrix}d_1, ..., d_j, ..., d_P\end{Bmatrix}$, where ${d_j} \in {R^{K}}$, $K=s_2 \times s_2$. $s_2$ is the patch size in $I_2$. \guSec{We empirically set $s_2>s_1$ to ensure sufficient training data for GMM.} The pixel intensity \gu{multiset} of an arbitrary $d_j$ is indicated as $Y$, $Y \in R^K$, and $y_k$ is \guSec{the $k$-th} pixel intensity in $Y$.

\subsection{The Probabilistic Model}
Our key idea is to model the underlying distribution of pixel intensities $X$ with the noisy observation $Y$. \textit{We use $X=\{x_m\}$ to denote the corresponding latent pixel intensities, for slight notation misuse}. \gu{We directly model the pixel intensities instead of the whole patch because the latter will suffer from significantly larger computation burden and noticeable interference due to increased size.} To relate $X$ with $Y$, we assume that $y_k$ follows a GMM whose centroids are $\{x_m\}$. That is, the GMM with those centroids can generate the noisy observations. Thus, we formulate the deblurring problem under the GMM probabilistic framework. \gu{Inspired by Myronenko et al. \cite{myronenko2010point}, we define the probability density function of $y_k$ as} 
\begin{equation}
\label{eq:eq1}
p(y_k)=(1-\omega)\sum_{m=1}^M\frac{1}{M}p(y_k|x_m)+\omega\frac{1}{K},
\end{equation}
where $p(y_k|x_m) = \frac{1}{({2 \pi{\sigma}_m^2})^{\frac{d}{2} }} e^{-\frac{||y_k-x_m||^2}{2{\sigma}_m^2}}$ denotes the $m$-th Gaussian component, and $d$ is the dimension of $x_m$ and $y_k$ ($d=1$ for gray image). An additional uniform distribution $\frac{1}{K}$ accounts for the noise, with a weight $\omega$. $\bm{\sigma}^2 = \begin{Bmatrix}{\sigma}_1^2, ..., {\sigma}_m^2, ..., {\sigma}_M^2\end{Bmatrix}$ stands for the covariances and $\frac{1}{M}$ represents the equal membership probability for all the Gaussian components. The centroids of the GMM model are initialized by $X$. We next need to find the centroids and covariances that can best explain the distribution of $Y$. 

\subsection{EM optimization}
The centroids and covariances of the GMM can be estimated by minimizing the \textit{negative} log-likelihood function \cite{bishop1995neural}.
\begin{equation}
\label{eq:eq2}
E(X,\bm{\sigma}^2)=-\sum_{k=1}^K\log(\frac{1-\omega}{M}\sum_{m=1}^Mp(y_k|x_m)+\omega\frac{1}{K}).
\end{equation}

We use the expectation-maximization (EM) algorithm \cite{dempster1977maximum} to solve Eq. (\ref{eq:eq2}). The EM algorithm consists of two steps: E-step and M-step. E-step and M-step are alternately called for multiple iterations to achieve decent estimations.

\textbf{E-step.} The posterior probability $p^{old}(x_m|y_k)$ is calculated based on Bayes' theorem and the parameters in the previous iteration. $p^{old}_{mk}$ represents $p^{old}(x_m|y_k)$ for simplicity.
\begin{equation}
\label{eq:eq3}
p^{old}_{mk} = \frac{ e^{-\frac{||y_k-x_m||^2}{2 {\sigma}_m^2}}}{\sum_{m=1}^M e^{-\frac{||y_k-x_m||^2}{2 {\sigma}_m^2}}+\frac{\omega M({2 \pi {\sigma}_m^2})^{\frac{d}{2} }}{(1-\omega)K}}.
\end{equation}

\textbf{M-step.} The M-step is to update the involved parameters ($X$ and $\bm{\sigma}^2$) based on the computed posteriors. This is equivalent to minimizing the upper bound of Eq. (\ref{eq:eq2}). ``$new$'' means calculating the posterior probability with the parameters to be estimated in the current iteration.
\begin{equation}
\begin{aligned}
\label{eq:eq4}
Q(X,\bm{\sigma}^2) &= -\sum \limits_{k=1}^{K} \sum \limits_{m=1}^{M} p^{old}_{mk} \log \frac{\left(\frac{1-\omega}{M}p^{new}(y_k \vert x_m)+\frac{\omega}{MK}\right)} { \left(\frac{1-\omega}{M}p^{old}(y_k \vert x_m)+\frac{\omega}{MK}\right)p^{old}_{mk}}\\
&\propto
\sum \limits_{k=1}^{K}\sum \limits_{m=1}^{M} p^{old}_{mk}\frac{\norm{y_k-x_m}^2}{2 {\sigma}_m^2} +\sum \limits_{k=1}^{K}\sum \limits_{m=1}^{M}\frac{p^{old}_{mk}}{2}\log  {\sigma}_m^2.
\end{aligned}
\end{equation}

\subsection{Bilateral Term}
\label{sec:Bilateral term}
\begin{figure}[t]
	\centering
\setcounter{subfigure}{0}
{
\includegraphics[width=38mm,scale=1.0]{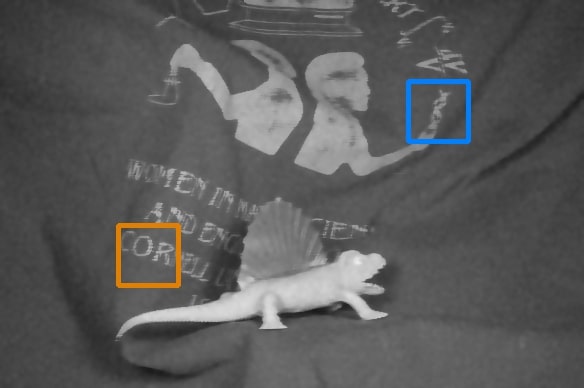}
 }
 \hfill
 {
\includegraphics[width=38mm,scale=1.0]{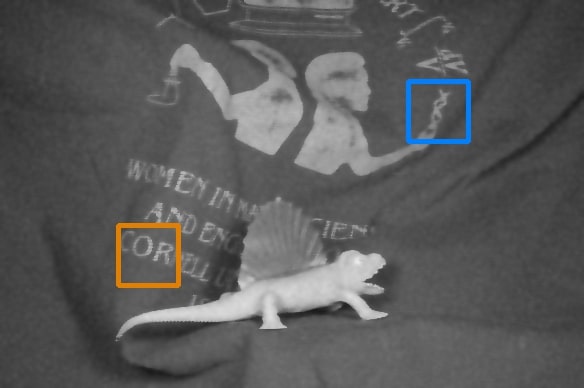}
}
\setcounter{subfigure}{0}
\subfloat[Without bilateral term]
{
\includegraphics[width=18mm,scale=1.0]{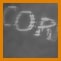}
 
\includegraphics[width=18mm,scale=1.0]{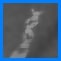}
\label{subfloat:without_bilateral_term}
}
\hfill
\subfloat[With bilateral term]
{
\includegraphics[width=18mm,scale=1.0]{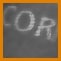}
\includegraphics[width=18mm,scale=1.0]{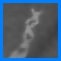}
}
\label{subfloat:with_bilateral_term}

\caption{ Deblurred results with and without the bilateral term: (a) without the bilateral term (i.e., $\lambda = 1.0$ in Eq. (\ref{eq:eq6})); (b) with the bilateral term ($\lambda = 0.75 $ in Eq. (\ref{eq:eq6})). }
\label{fig:bilateral comparison}
\end{figure}

Eq. (\ref{eq:eq4}) can be treated as a data term, which in this work is to numerically approximate $Y$ with $X$. However, this data term only takes the pixel intensity distribution into account, without considering the spatial information. As illustrated in Fig. \ref{fig:bilateral comparison}(a), sharp edges would become coarse (e.g., discontinuity) in the iteration of EM. To overcome this problem, we propose to add a bilateral term to the objective function in M-step. Inspired by the bilateral filter \cite{tomasi1998bilateral}, we define the bilateral term $B$ as
\begin{equation}
\label{eq:eq5}
B(X) = \sum_{m'\in N(m)}\norm{x_m-x_{m'}}^2e^{-\frac{d_{m'}^2}{{2\sigma}_d^2}}e^{-\frac{l_{m'}^2}{{2\sigma}_l^2}},
\end{equation}
where $m' \in N(m)$ denotes a neighbour pixel with its intensity equals to $x_{m'}$. $d_{m'}$ and 
$l_{m'}$ are the spatial distance and the difference of intensity value between the neighbour pixel $m'$ and the center pixel $m$, respectively. $\sigma_d$ and 
$\sigma_l$ are constants to control the degree of smoothness.

Redefining Eq. (\ref{eq:eq4}) as $D(X,\bm{\sigma}^2)$ and weighing it with $\lambda$, the final objective function can be written as
\begin{equation}
\label{eq:eq6}
Q(X,\bm{\sigma}^2) = \lambda D(X,\bm{\sigma}^2) + (1-\lambda )B(X).
\end{equation}
We next need to minimize Eq. (6), to solve the involved parameters.

\subsection{Minimization}
\label{sec:Minimization}

In this section we explain how to solve the optimum solutions of $x_m$ and $ {\sigma}_m^2$. We first take the partial derivation of Eq. (\ref{eq:eq6}) with respect to $x_m$,  

\begin{equation}
\label{eq:eq7}
\begin{aligned}
\frac{\partial Q(X, {\bm{\sigma}}^2)}{\partial x_m} = &\frac{\lambda}{{\sigma}_m^2}\sum \limits_{k=1}^{K}p^{old}_{mk}(x_m-y_{k}) +(1-\lambda)\sum_{m'\in N(m)} \frac{2(x_m-x_{m'})}{e^{\frac{d_{m'}^2}{{2\sigma}_d^2}}e^{\frac{l_{m'}^2}{2\sigma_l^2}}}.
\end{aligned}
\end{equation}

Given the mathematical difficulty to directly equate Eq. (\ref{eq:eq7}) to zero and solve $x_m$, due to the $x_m$ in $l_{m'}^2$, we apply gradient descent to approximate the optimum $x_m$ instead. Gradient descent shares the huge advantage in its efficiency to iteratively update the value of $x_m$ with the manual step value. In each gradient descent step $x_m$ is updated as 

\begin{equation}
\label{eq:eq8}
\begin{split}
x_{m}^{q+1} =  x_{m}^{q} + \alpha\left(\lambda \frac{\sum \limits_{k=1}^{K}p^{old}_{mk}(y_{k}-x_{m}^{q})}{\sum\limits_{k=1}^{K}p^{old}_{mk}} -\mu\sum_{m'\in N(m)} \frac{x_{m}^{q}-x_{m'}^{q}}{e^{\frac{d_{m'}^2}{{2\sigma}_d^2}}e^{\frac{l_{m'}^2}{{2\sigma}_l^2}}}\right),
\end{split}
\end{equation}
where $\mu =  2\frac{ {\sigma}_m^2(1-\lambda)}{\sum_{k=1}^{K}p^{old}_{mk}}$, $q+1$ and $q$ denote the $(q+1)$-th and $q$-th gradient descent iteration respectively. $\alpha$ represents the gradient descent step, which is set to  0.1 in our experiments. Eq. (\ref{eq:eq8}) is obtained by empirically scaling the gradient of Eq. (\ref{eq:eq7}) by ${ {\sigma}_m^2/\sum_{k=1}^{K}p^{old}_{mk}}$ for the convenience of better controlling the gradient descent.
$\mu$ plays a role in adjusting the proportion of the bilateral term. In general, we take $\mu \in [0.1,1.0]$ as a controllable parameter for simplicity. Notice that $p^{old}_{mk}$ is not updated in the gradient descent iteration for lowering computational burden.

$x_m$ is updated to $x_{m}'$ after the gradient descent meets the termination criterion (in the experiment, the iteration stops if the iteration number reaches 50). We take the partial derivative of Eq. (\ref{eq:eq6}) with respect to $ {\sigma}_m^2$. By solving ${\partial Q}/{\partial  {\sigma}_m^2}=0$, $ {\sigma}_m^2$ is updated as

\begin{equation}
\label{eq:eq9}
\begin{split}
{ {\sigma}_m^2}' = \left(\sum_{k=1}^K p^{old}_{mk} \Vert x_{m}'-y_k\Vert^2 \right) /\sum_{k=1}^{K}p^{old}_{mk}.
\end{split}
\end{equation}

\begin{figure}[t]
\centering
  \includegraphics[angle=90,origin=b,width=0.8\linewidth]{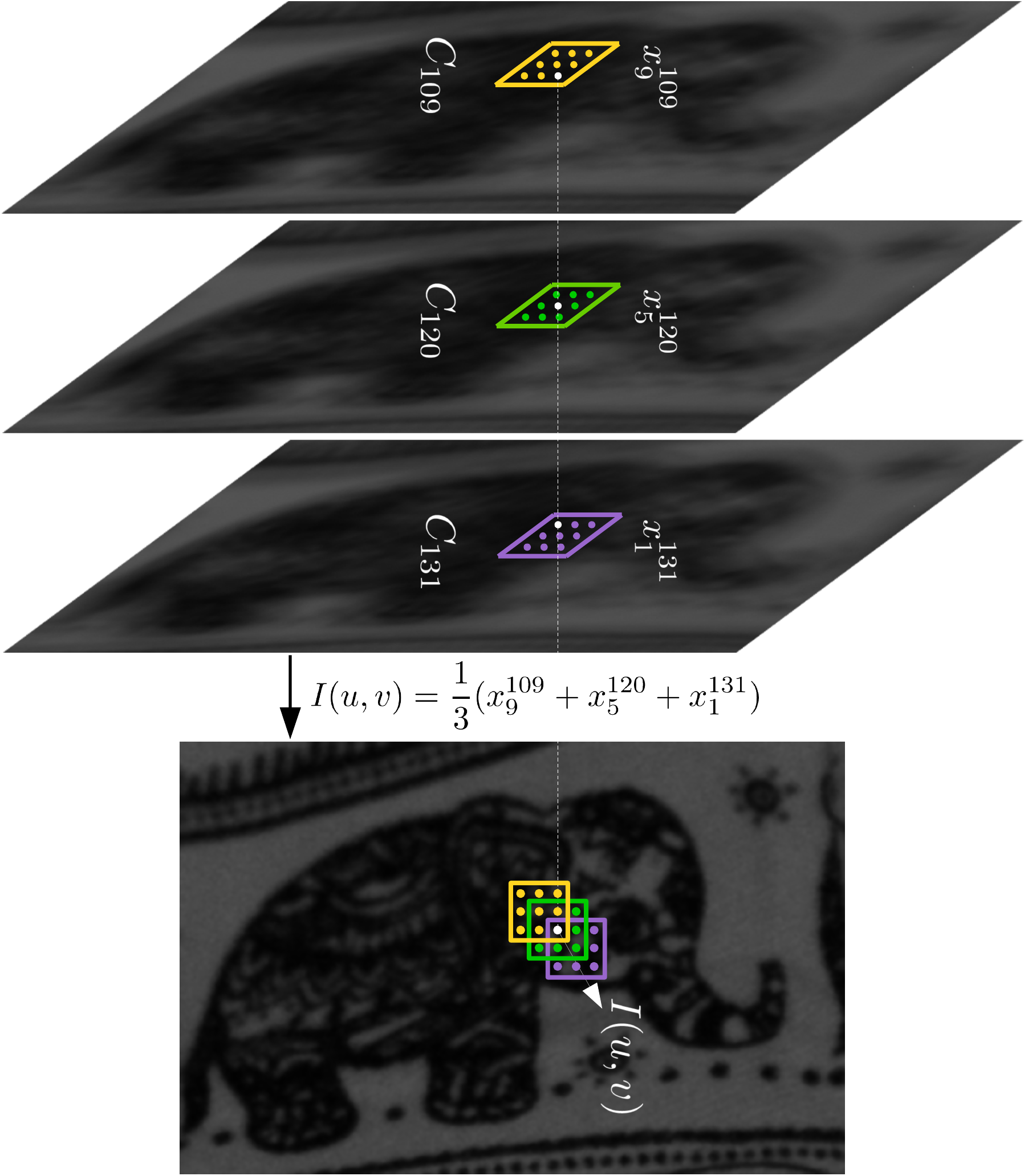}
   \caption{Illustration of how the pixel value $x_m$ of an arbitrary pixel $m$ located at $(u,v)$ is calculated from multiple overlapped patches. Here, three overlapped patches $c_i$, (here $i=109,120,131$) are assumed, which  contain pixel intensity valued $x_9$, $x_5$ and $x_1$, respectively. The final pixel intensity $x_m$ located at $(u,v)$ is calculated as the average of $x_9^{109}, x_5^{120}$, and $x_1^{131}$.  }
\label{fig:overlapping patches}
\end{figure}

Notice that the step size for slicing $I_1$ into patches should be small so that a certain pixel can be updated in different GMM models due to overlapped patches. The final output value of a certain pixel $I_{u,v}$ is calculated by simply averaging all the updated values located at $(u,v)$,
\begin{equation}
\label{eq:eq10}
I(u,v)=\frac{\sum_{i,m}{\mathbbm{1}(pos_m^i=(u,v))x_m^{i}}}{\sum_{i,m}{\mathbbm{1}(pos_m^i=(u,v))}},
\end{equation}
where $pos_m^i$ denotes the position of pixel $m$ in the $i$-th patch, $x_m^{i}$ denotes the pixel intensity of $m$ in the $i$-th patch. $\mathbbm{1}$ is an indicator function. See Fig. \ref{fig:overlapping patches} for an example.

\subsection{Optical Flow Update}
\label{sec:opticalFLowUpdate}
Blur hinders accurate estimation of optical flow, which can possibly lead to inaccurate matches in finding patch correspondences. 
To mitigate this issue, we alternate  optical flow and the EM algorithm for multiple iterations. $I_1^t$ denotes the deblurred result in the $t$-th iteration ($T$ times in total), and is used to compute the optical flow in the $(t+1)$-th iteration. Updating optical flow increases the confidence of the patch correspondences.

\begin{figure}[t]
	\centering
\setcounter{subfigure}{0}
{
\includegraphics[width=38mm,scale=1.0]{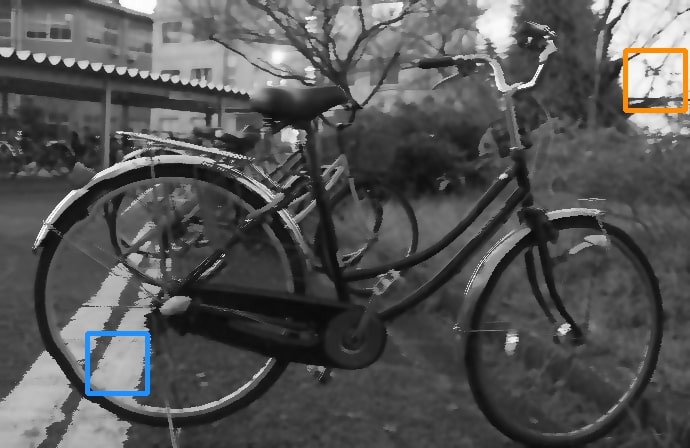}}
 \hfill
 {
\includegraphics[width=38mm,scale=1.0]{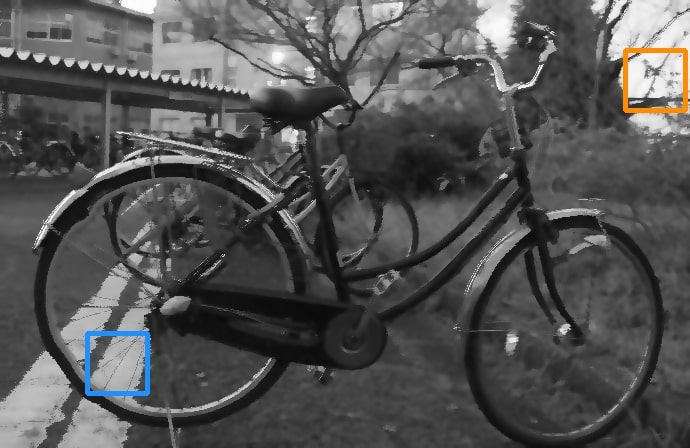}
}

\setcounter{subfigure}{0}
\subfloat[Without detail layer]
{
\includegraphics[width=18mm,scale=1.0]{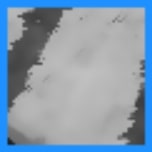}
\includegraphics[width=18mm,scale=1.0]{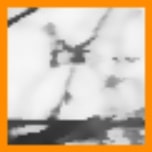}
\label{subfloat:without detail layer}
}
\hfill
\subfloat[With detail layer]
{
\includegraphics[width=18mm,scale=1.0]{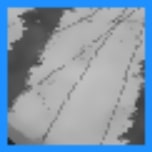}

\includegraphics[width=18mm,scale=1.0]{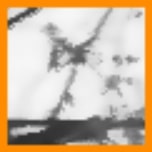}
}
\label{subfloat:with detail layer}
\caption{Deblurred results with and without adding the detail layer: (a) without adding the detail layer, (b) with adding the detail layer.}
\label{fig:detail layer comparison}
\end{figure}

\subsection{Detail Layer}
We extract the sharp features from $I_2$ and add it back to ${I_1^T}$ to further preserve the details. A similar idea has been used in \cite{yuan2007image}. Since the noise in $I_2$ can negatively affect the quality of the detail layer, we apply the bilateral filter \cite{tomasi1998bilateral} to $I_2$ at first.
We then obtain a mask $I_m$ by applying the Laplacian filter \cite{burt1983laplacian} to $I_2$, to select the retained details. Since the $I_2$ and $I_1^T$ are in different views, we use the DOF filed between them to find the spatial correspondence (see Sec. \ref{sec:patchcorrespondence}). 
We can observe from Fig. \ref{fig:detail layer comparison} that the details are better recovered by adding the detail layer. 
\begin{algorithm}[tb]
\caption{Adding detail layer (DL)}
\begin{algorithmic}[1]
\label{alg: algorithm 1}
\renewcommand{\algorithmicrequire}{\textbf{Input:}}
\renewcommand{\algorithmicensure}{\textbf{Output:}}
\REQUIRE Deblurred image ${I_1^T}$, enhanced noisy image $I_2$, constant threshold $\tau \in [10,150]$, detail weight $\eta \in [0.1,0.5]$
\ENSURE  Deblurred image with sharp features added
\STATE Apply bilateral filter to $I_2$
\STATE Apply Laplacian filter to $I_2$ to obtain mask image $I_m$
\FOR{every pixel located at $(u,v)$ in ${I_1^T}$}
    \STATE{Find the correspondence according to vector in DOF field from $(u,v)$ in ${I_1^T}$ to $(u',v')$ in $I_2$}
      \IF{$I_m(u',v') > \tau$}
        \STATE ${I_1^T}(u,v) \gets (1-\eta) {I_1^T}(u,v)+\eta I_2(u',v')$
      \ENDIF
    \ENDFOR
\RETURN Updated $I_1^{T}$
\end{algorithmic} 
\end{algorithm}
The algorithm of adding details is listed in Alg. \ref{alg: algorithm 1}. The proposed deblurring algorithm is summarized in Alg. \ref{alg: algorithm 2}.

\begin{algorithm}[tb]
\caption{Image deblurring (OGMM+DL)}
\begin{algorithmic}[1]
\label{alg: algorithm 2}
\renewcommand{\algorithmicrequire}{\textbf{Input:}}
\renewcommand{\algorithmicensure}{\textbf{Output:}}
\REQUIRE Blurred image $I_1$, enhanced noisy image $I_2$, iteration times $T$, termination number $\gamma$
\ENSURE  Deblurred image  ${I_1^T}$
\STATE Parameters setting: $ {\sigma}_m^2 \in [100,500]$, $\omega = 0.02$, $\lambda \in [0.75,0.8]$
\FOR{$t=1$ to $T$}
\STATE Update optical flow and find corresponding patches with respect to $I_1^{t-1}$ and $I_2$ ($I_1^0=I_1$)
\FOR{Each patch in $I_1^{t-1}$}
\STATE Initialize centroids by $X$
\REPEAT
\STATE E-step: update each $p^{old}_{mk}$ by Eq. (\ref{eq:eq3})
\STATE M-step: update each $x_m$ and $ {\sigma}_m^2$ by Eq. (\ref{eq:eq8}) and Eq. (\ref{eq:eq9})
\UNTIL $\gamma$ is reached
\ENDFOR
\STATE Obtain $I_1^{t}$ via Eq. (\ref{eq:eq10})
\ENDFOR
\STATE Add details to ${I_1^T}$ by Alg. \ref{alg: algorithm 1}
\RETURN ${I_1^T}$
\end{algorithmic} 
\end{algorithm}


\begin{figure}[htb]
\setcounter{subfigure}{0}
\subfloat[Blurred IMG]{\includegraphics[origin=b,width=0.25\linewidth]{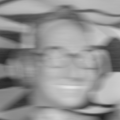}}
\hspace{5pt}
\begin{minipage}{0.95\linewidth}
\subfloat[$s1=3, s2=5$]{
\includegraphics[width=0.22\linewidth]{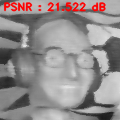}}
\subfloat[$s1=3, s2=7$]{
\includegraphics[width=0.22\linewidth]{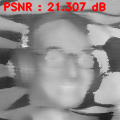}}
\subfloat[$s1=3, s2=9$]{
\includegraphics[width=0.22\linewidth]{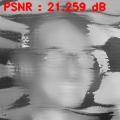}}
   
\subfloat[$s1=5, s2=7$]{
\includegraphics[width=0.22\linewidth]{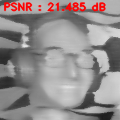}}
\subfloat[$s1=5, s2=9$]{
\includegraphics[width=0.22\linewidth]{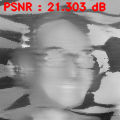}}
\subfloat[$s1=7, s2=9$]{
\includegraphics[width=0.22\linewidth]{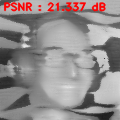}}
\end{minipage}

\hspace{7pt}
\subfloat[$\omega = 0.02$]{
\includegraphics[width=0.21\linewidth]{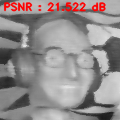}}
\subfloat[$\omega = 0.5$]{
\includegraphics[width=0.21\linewidth]{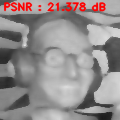}}
\subfloat[$\omega = 0.8$]{
\includegraphics[width=0.21\linewidth]{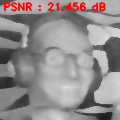}}
\subfloat[\guSec{$s1=3, s2=3$}]{
\includegraphics[width=0.21\linewidth]{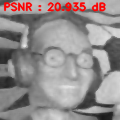}}

\caption{\gu{Comparison of different parameter settings ($s_2, s_1, \omega$) on the deblurring results of (a). (b)$\sim$(d), (e)$\sim$(g), and (h)$\sim$(j) explain the influence of $s_2$, $s_1$ and $\omega$, respectively. \guSec{(k) describes the deblurring result under equal patch size ($s_1 = s_2$).} }}
\label{fig:parameterSettings}
\end{figure}

\begin{figure}[htb]
\setcounter{subfigure}{0}
\subfloat[\guThi{Blurred IMG}]{\includegraphics[origin=b,width=0.25\linewidth]{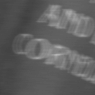}}
\hspace{5pt}
\begin{minipage}{0.95\linewidth}
\subfloat[\guThi{Gd:10, EM:1}]{
\includegraphics[width=0.22\linewidth]{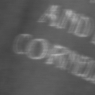}}
\subfloat[\guThi{Gd:30, EM:1}]{
\includegraphics[width=0.22\linewidth]{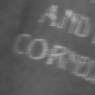}}
\subfloat[\guThi{Gd:50, EM:1}]{
\includegraphics[width=0.22\linewidth]{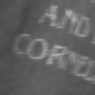}}
   
\subfloat[\guThi{Gd:10, EM:2}]{
\includegraphics[width=0.22\linewidth]{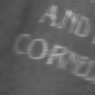}}
\subfloat[\guThi{Gd:30, EM:2}]{
\includegraphics[width=0.22\linewidth]{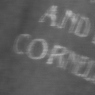}}
\subfloat[\guThi{Gd:50, EM:2}]{
\includegraphics[width=0.22\linewidth]{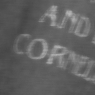}}

\subfloat[\guThi{Gd:10, EM:3}]{
\includegraphics[width=0.22\linewidth]{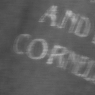}}
\subfloat[\guThi{Gd:30, EM:3}]{
\includegraphics[width=0.22\linewidth]{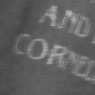}}
\subfloat[\guThi{Gd:50, EM:3}]{
\includegraphics[width=0.22\linewidth]{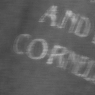}}
\end{minipage}

\caption{\guThi{An example of intermediate deblurred results produced at different gradient descent iterations in each round of EM update. Gd and EM denote the gradient descent iteration in one EM update, respectively. }}
\label{fig:intermedaiteResults}
\end{figure}

\gu{\subsection{Key Parameters, Time Complexity and Convergence}
\label{sec:parameter Analysis}
To gain insights into the impact of parameters on OGMM, we analyze the influence of the key parameters (i.e., patch size in $I_1$: $s_1$, patch size in $I_2$: $s_2$, noise weight: $\omega$) on the deblurring results (Fig. \ref{fig:parameterSettings}). We also give a brief discussion on time complexity.
}

\gu{We first discuss the effect of patch size (Fig. \ref{fig:parameterSettings}(b)$\sim$(g)). $s_1$ and $s_2$ directly relate to $M$ and $K$, respectively, as mentioned in Sec. \ref{sec:patchcorrespondence}. $s_2$ is required to be larger than $s_1$ to ensure a sufficient number of pixels in the corresponding noisy patch for running OGMM. \guSec{Specifically, an equal patch size ($s_1 = s_2$) or a larger $s_1$  may weaken the ability of our method to capture a decent intensity distribution. The low PSNR in Fig. \ref{fig:parameterSettings}(k) accounts for a less decent outcome for the equivalent patch size.} It can be observed from Fig. \ref{fig:parameterSettings}(b)$\sim$(g) that the increase of $s_1$/$M$ and $s_2$/$K$ leads to worse deblurred results. This is mainly because the larger $s_2$/$K$ is, the less relevant information in the noisy patch which corresponds to the blurred patch is involved. In addition, a larger $s_1$/$M$ contains more blurred pixels. If the estimated optical flow is not sufficiently accurate, it will lead to unnatural results. Based on these observations, we empirically set $s_1=3, s_2=5$ as it achieves decent deblurring results in general. As for $\omega$, which accounts for ``outliers'', we empirically found that $\omega =0.02$ is a good choice in terms of both deblurring and preserving details. A greater $\omega$ will degrade the performance, as shown in Fig. \ref{fig:parameterSettings}(h)$\sim$(j).}

\begin{figure}[t]
\centering
  \includegraphics[width=0.45\linewidth]{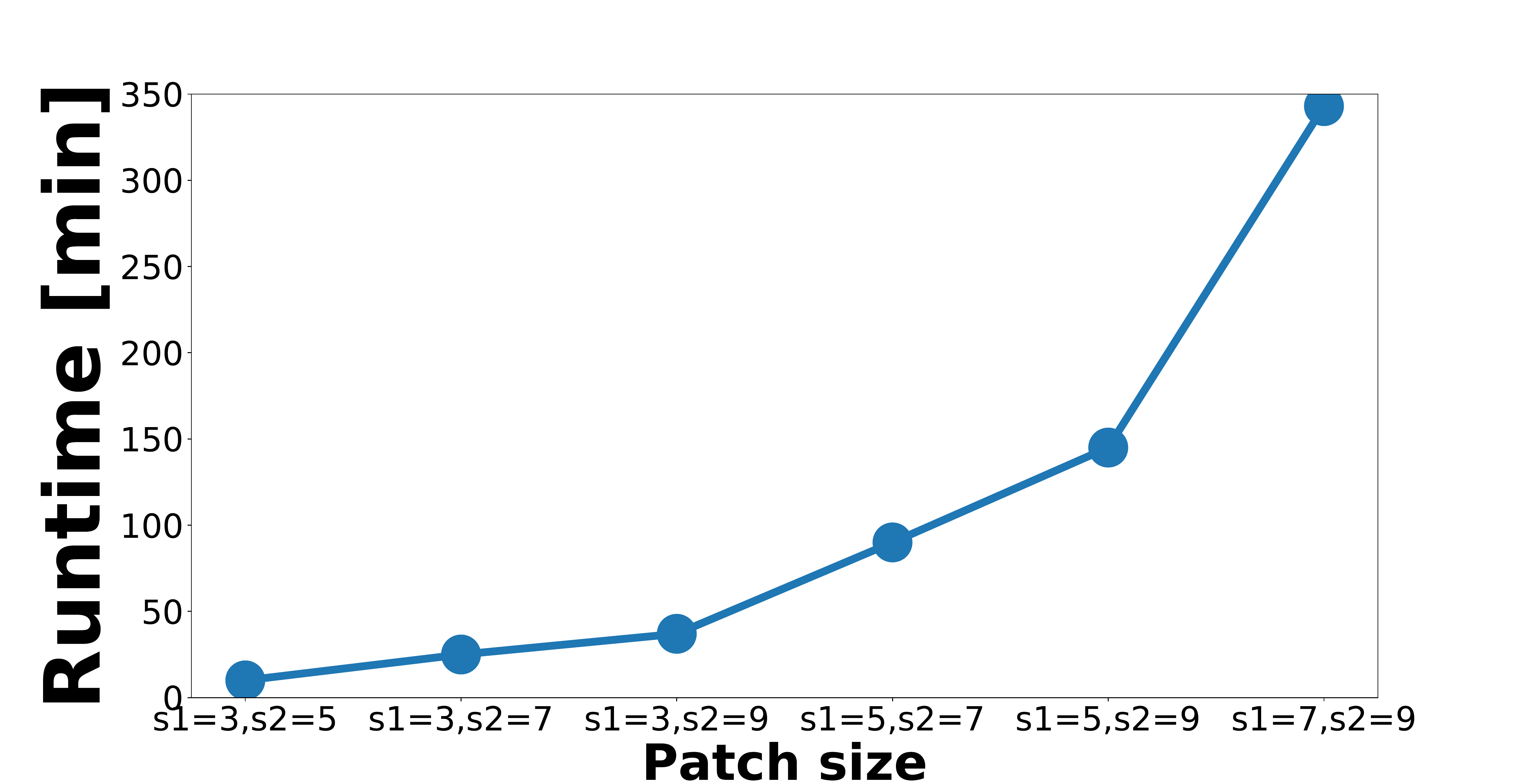}
    \includegraphics[width=0.45\linewidth]{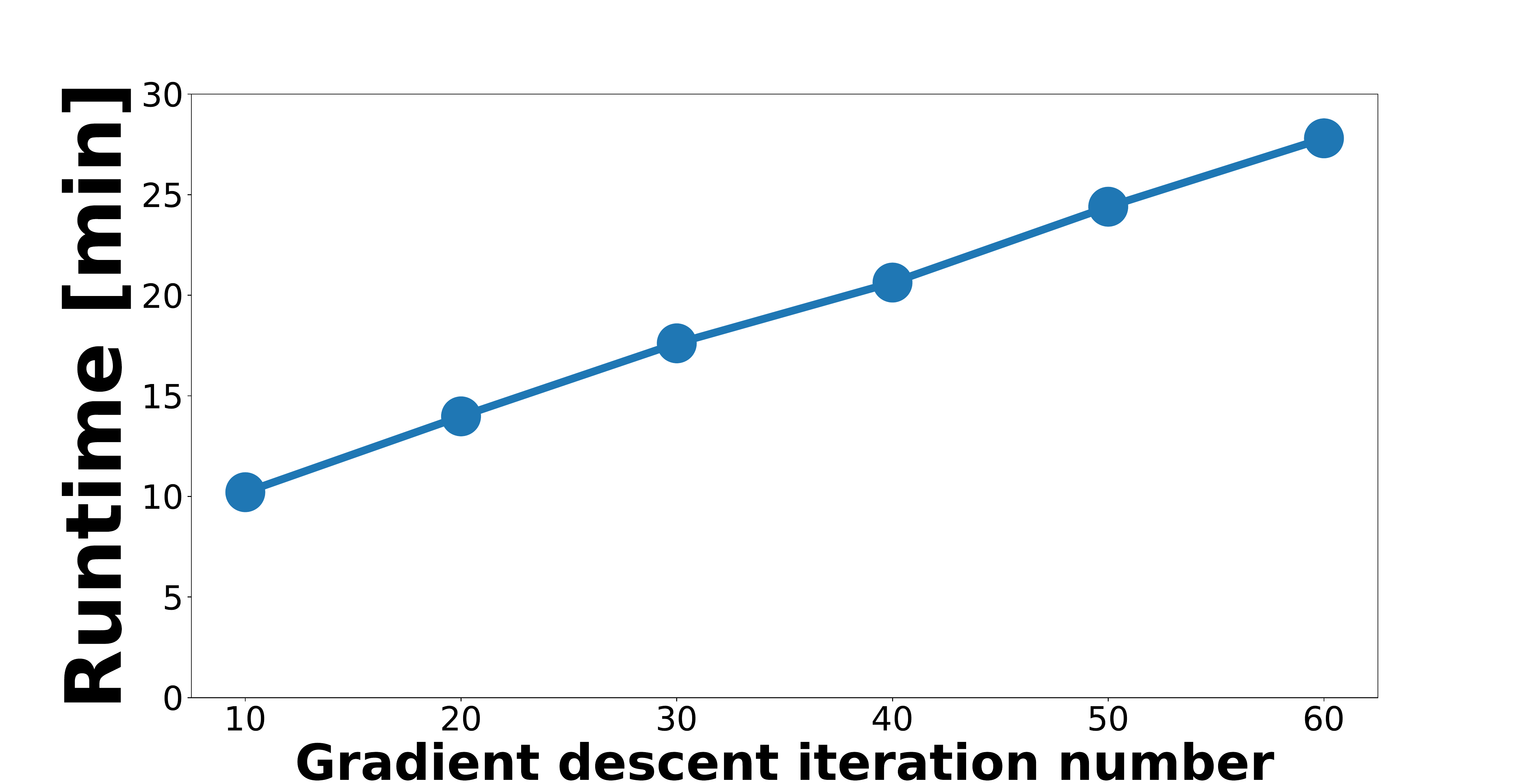}
   \caption{\gu{The running time with respect to the patch size (left) and the number of iterations (right).} } 
\label{fig:timeComplexity}
\end{figure}

\begin{figure*}[t]
\centering
    \subfloat[$s_1=3, s_2=5, \omega = 0.02$]{\includegraphics[width=0.167\linewidth]{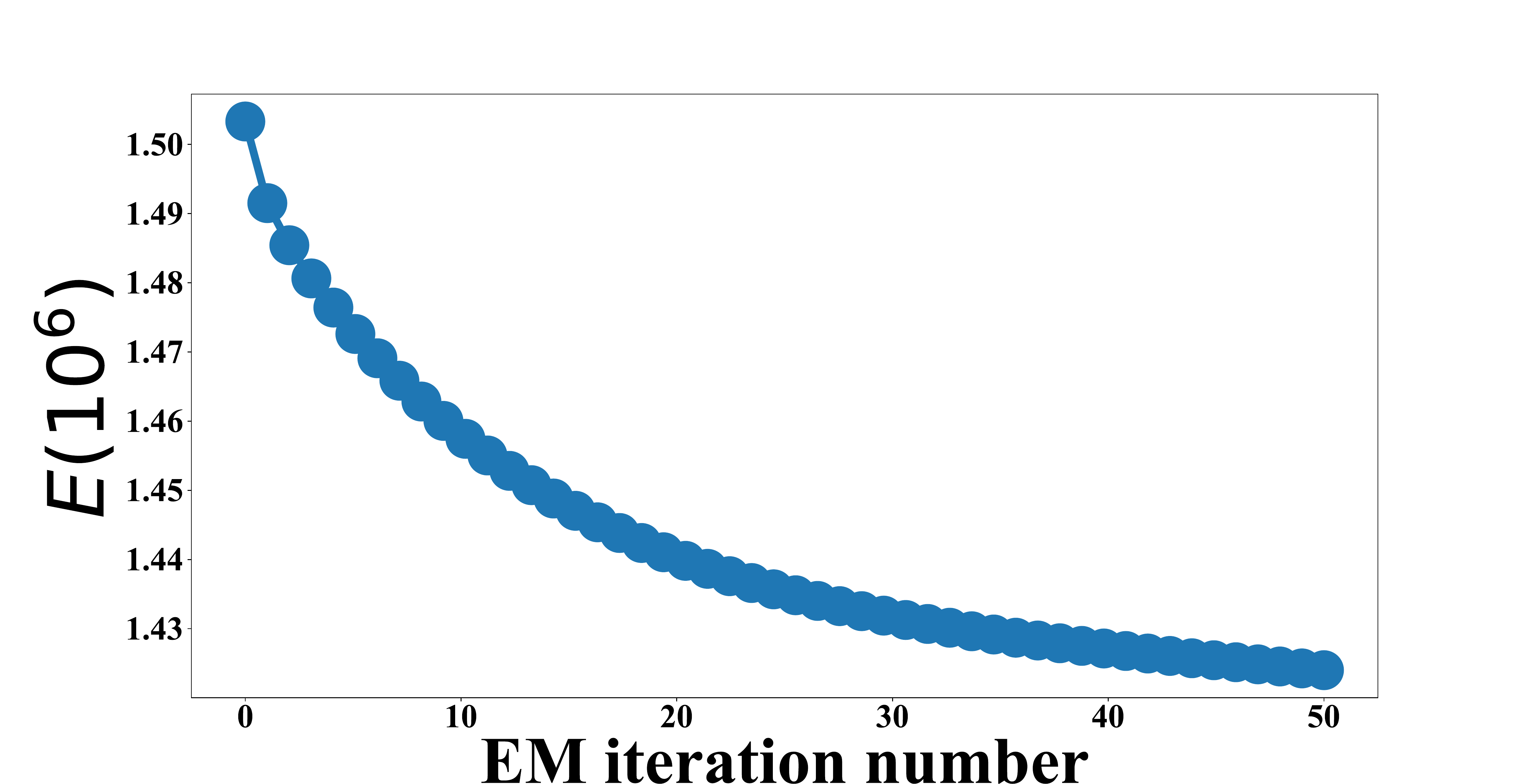}}
    \subfloat[$s_1=3, s_2=7, \omega = 0.02$]{\includegraphics[width=0.167\linewidth]{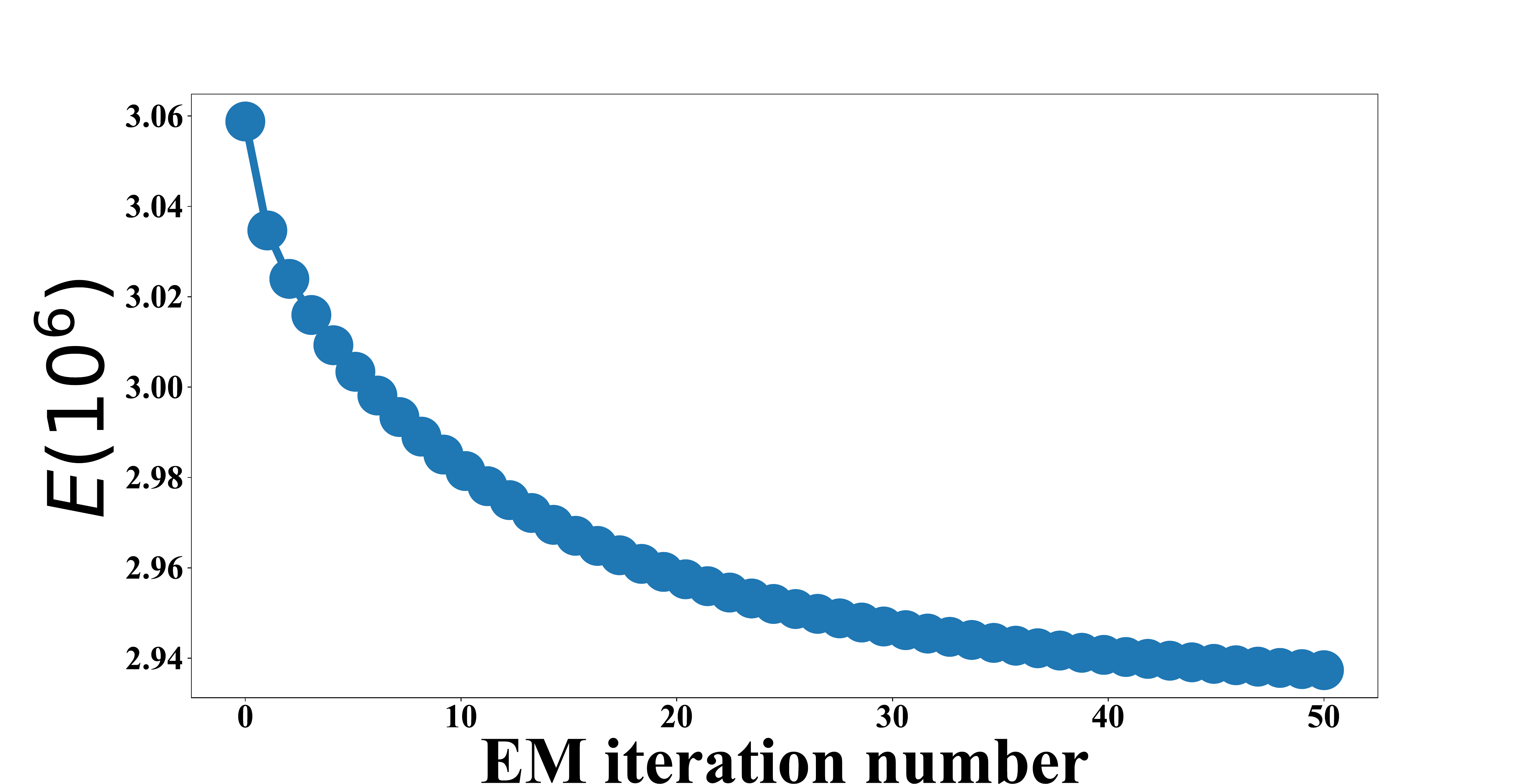}}
    \subfloat[$s_1=3, s_2=9, \omega = 0.02$]{\includegraphics[width=0.167\linewidth]{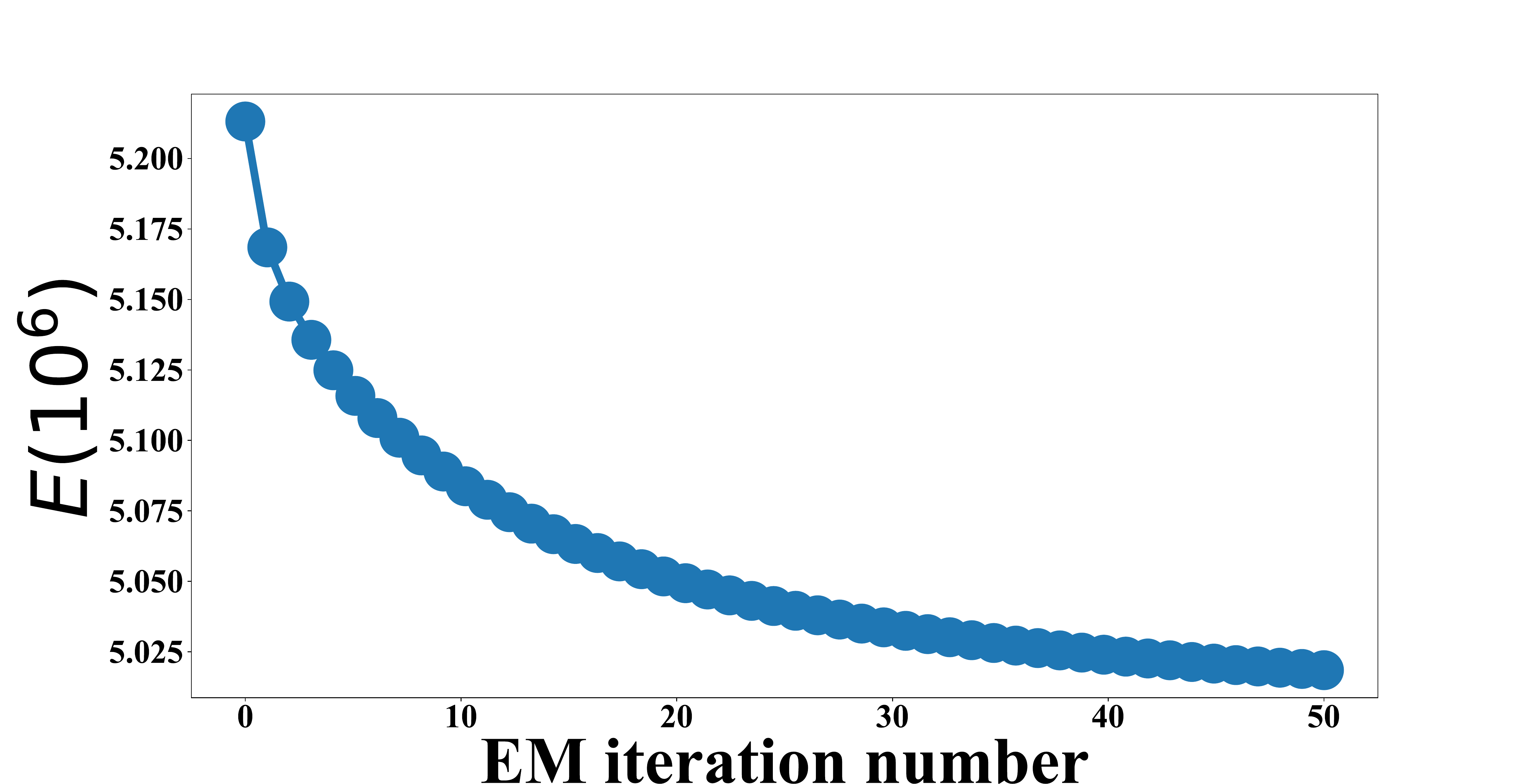}}
    \subfloat[$s_1=5, s_2=7, \omega = 0.02$]{\includegraphics[width=0.167\linewidth]{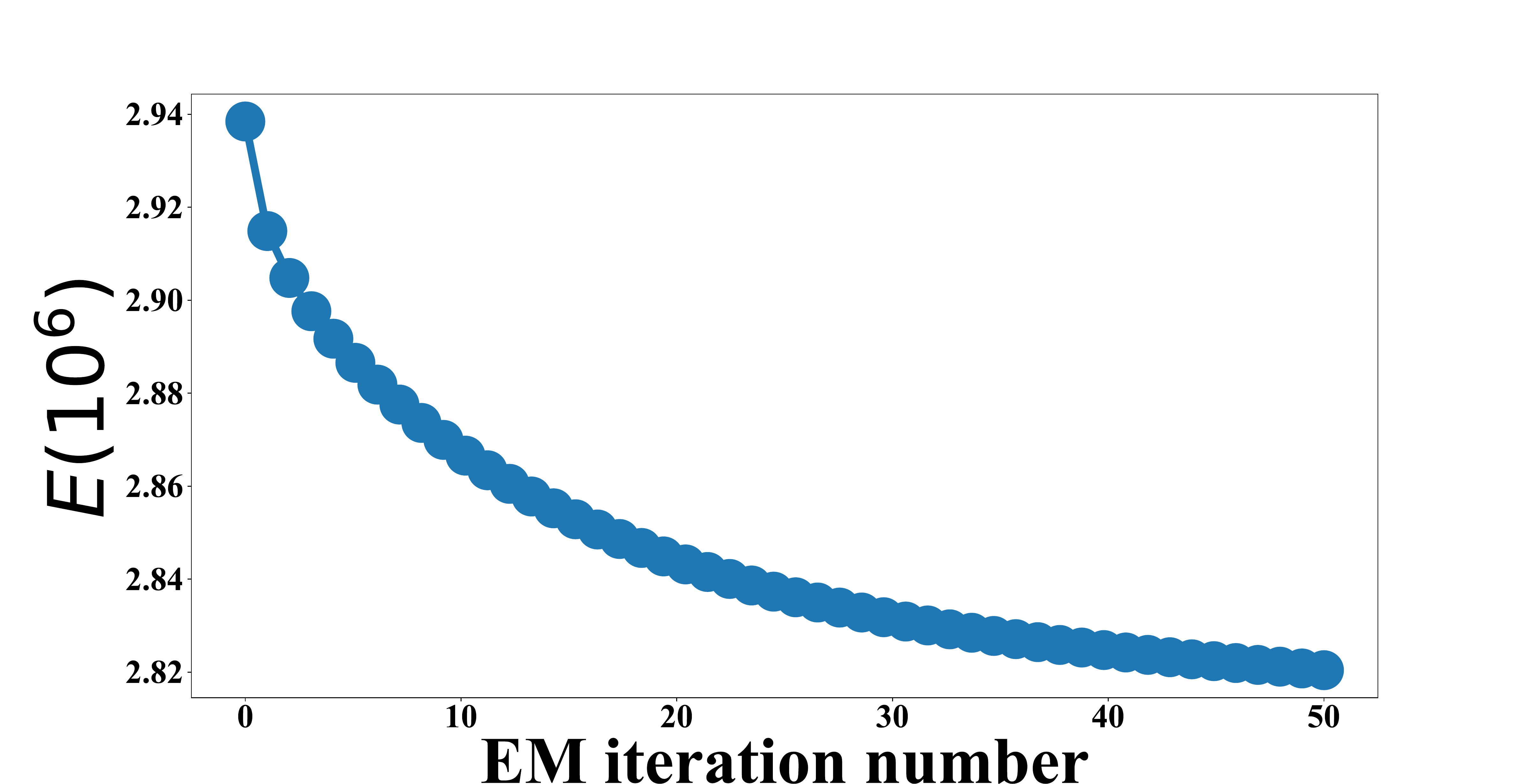}}
    \subfloat[$s_1=3, s_2=5, \omega=0.5$]{\includegraphics[width=0.167\linewidth]{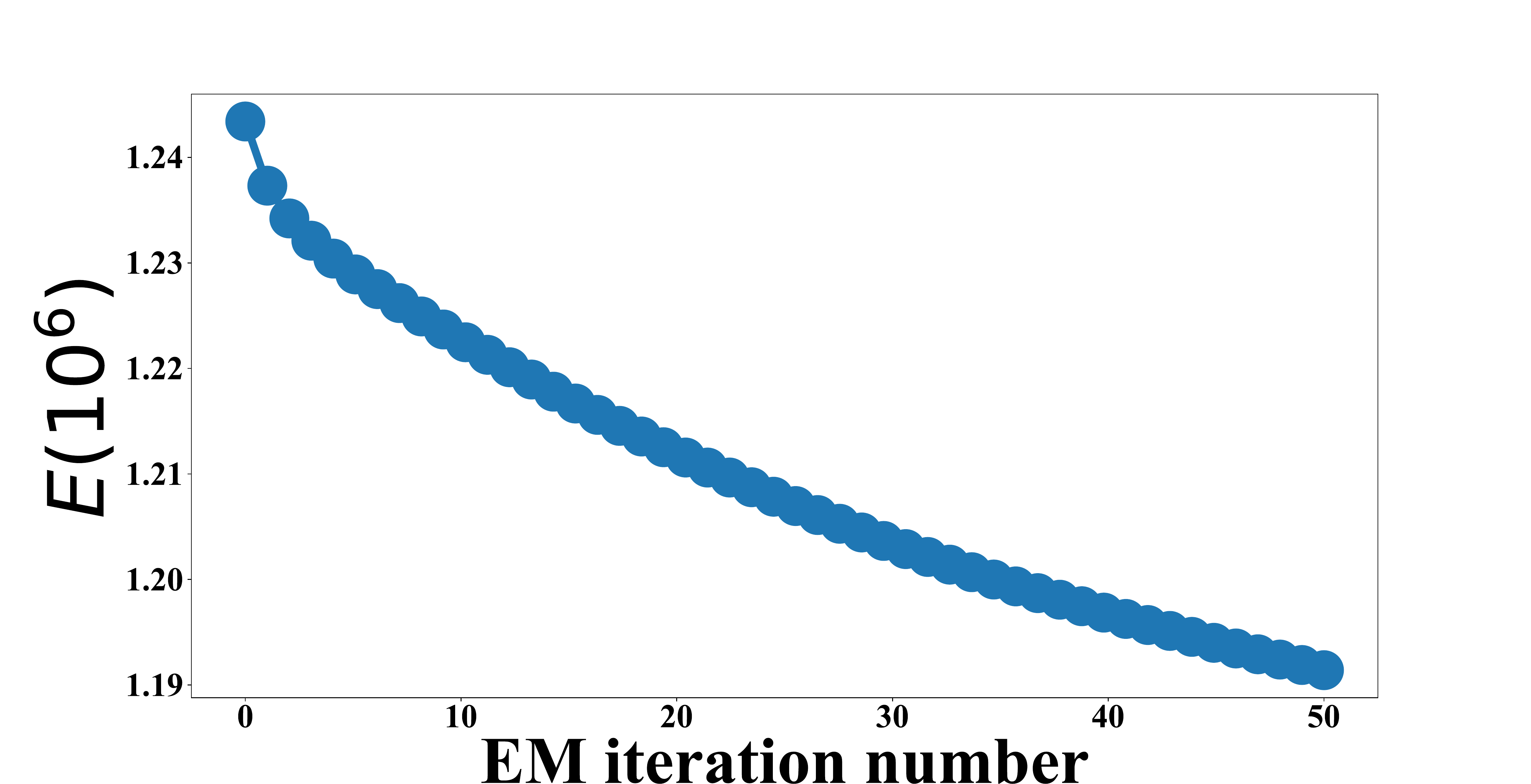}}
    \subfloat[$s_1=3, s_2=5, \omega=0.8$]{\includegraphics[width=0.167\linewidth]{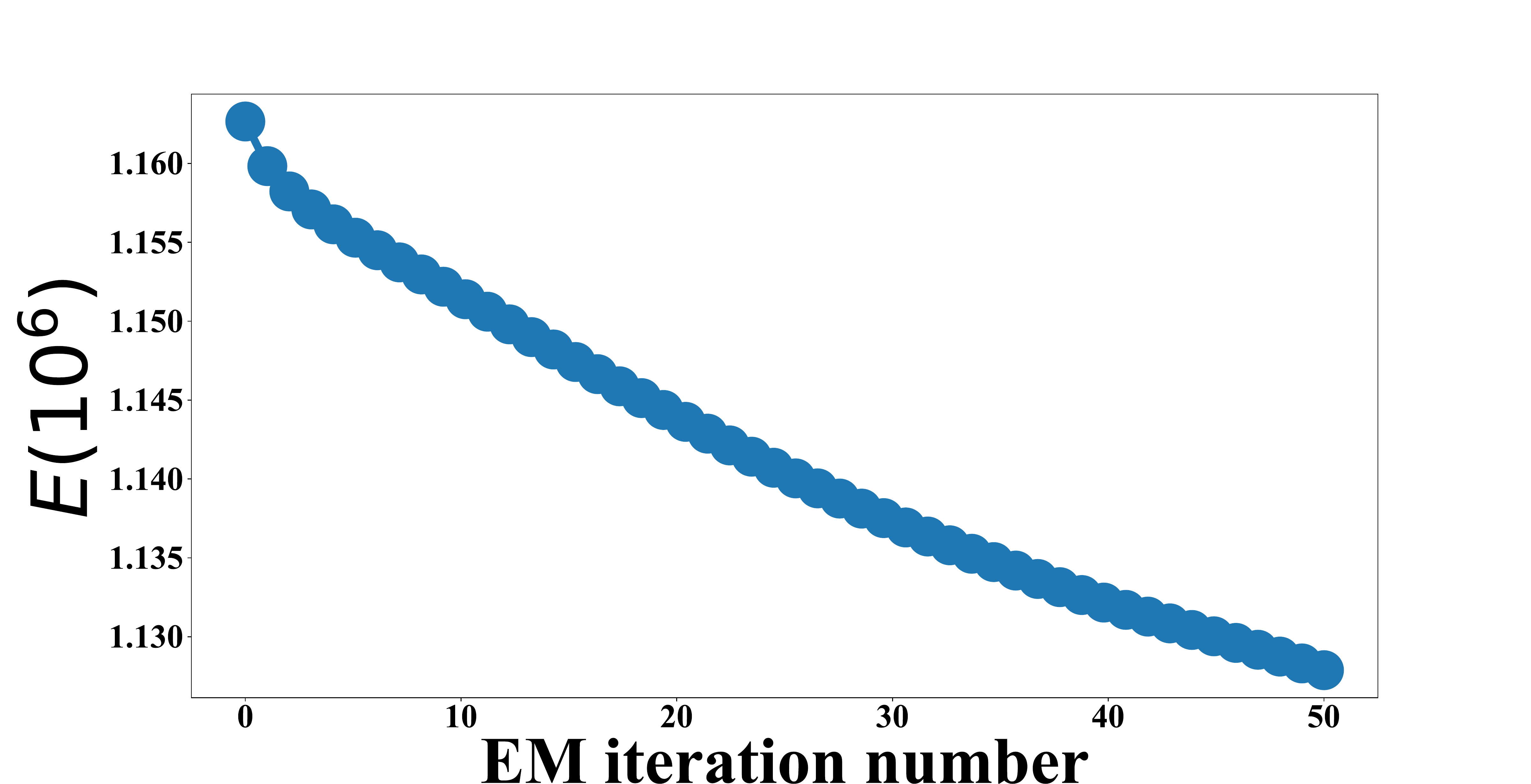}}
   \caption{\gu{Curves of Eq. \ref{eq:eq2} with respect to the EM iterations under different parameter settings.}}
\label{fig:EMlikelihood}
\end{figure*}

\gu{Similar to other EM-based algorithms, such as \cite{xu2015patch,luo2016adaptive}, the complexity of our algorithm under EM framework mainly depends on the termination criterion and the size of data (patch size). \gu{Fig. \ref{fig:timeComplexity} shows the changes of runtime with respect to the patch size (Fig. \ref{fig:parameterSettings}(b)$\sim$(g)) and the number of iterations (Fig. \ref{fig:parameterSettings}(b)). We observe that the runtime has an exponential-like relationship with the patch size and it is linearly proportional to the number of iterations.
In addition, the curves of Eq. \eqref{eq:eq2} with respect to the number of EM iterations with different parameter settings are shown in Fig. \ref{fig:EMlikelihood}. It can be seen from Fig. \ref{fig:EMlikelihood}(a,b,c) that the energy of Eq. \eqref{eq:eq2} declines drastically in the first few iterations, and tend to be steady (i.e., converged) with increasing iterations. In practice, we empirically set the EM iteration number to 1 to 5, depending on the size of the image. In each iteration, 50 rounds of gradient descent are called. \guThi{Fig. \ref{fig:intermedaiteResults} shows an example of intermediate deblurred results produced at different gradient descent iterations in each round of EM update. It can be seen that the first EM update contributes the most to removing the blur ((Fig. \ref{fig:intermedaiteResults}(b)$\sim$(d))), comparing to the following updates (Fig. \ref{fig:intermedaiteResults}(e)$\sim$(j)).} Also, a greater $\omega$ would lead to a slower convergence (Fig. \ref{fig:EMlikelihood}(e,f)). } } 

\begin{figure}[htb]
\centering
\setcounter{subfigure}{0}
\begin{minipage}{0.70\linewidth}
\subfloat[\textit{BlurType1}]{
   \includegraphics[width=0.28\linewidth]{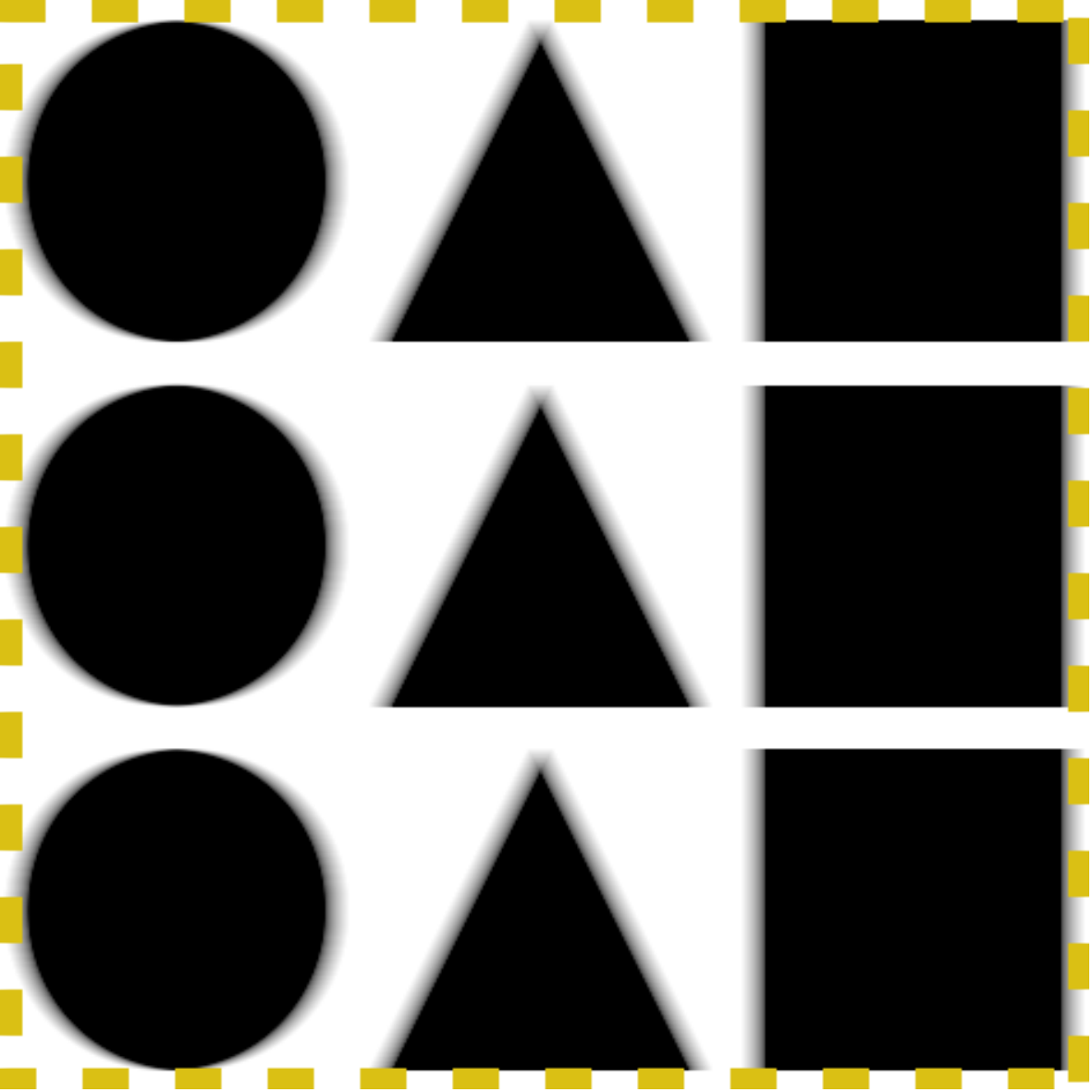}}
\subfloat[\textit{BlurType2}]{
   \includegraphics[width=0.28\linewidth]{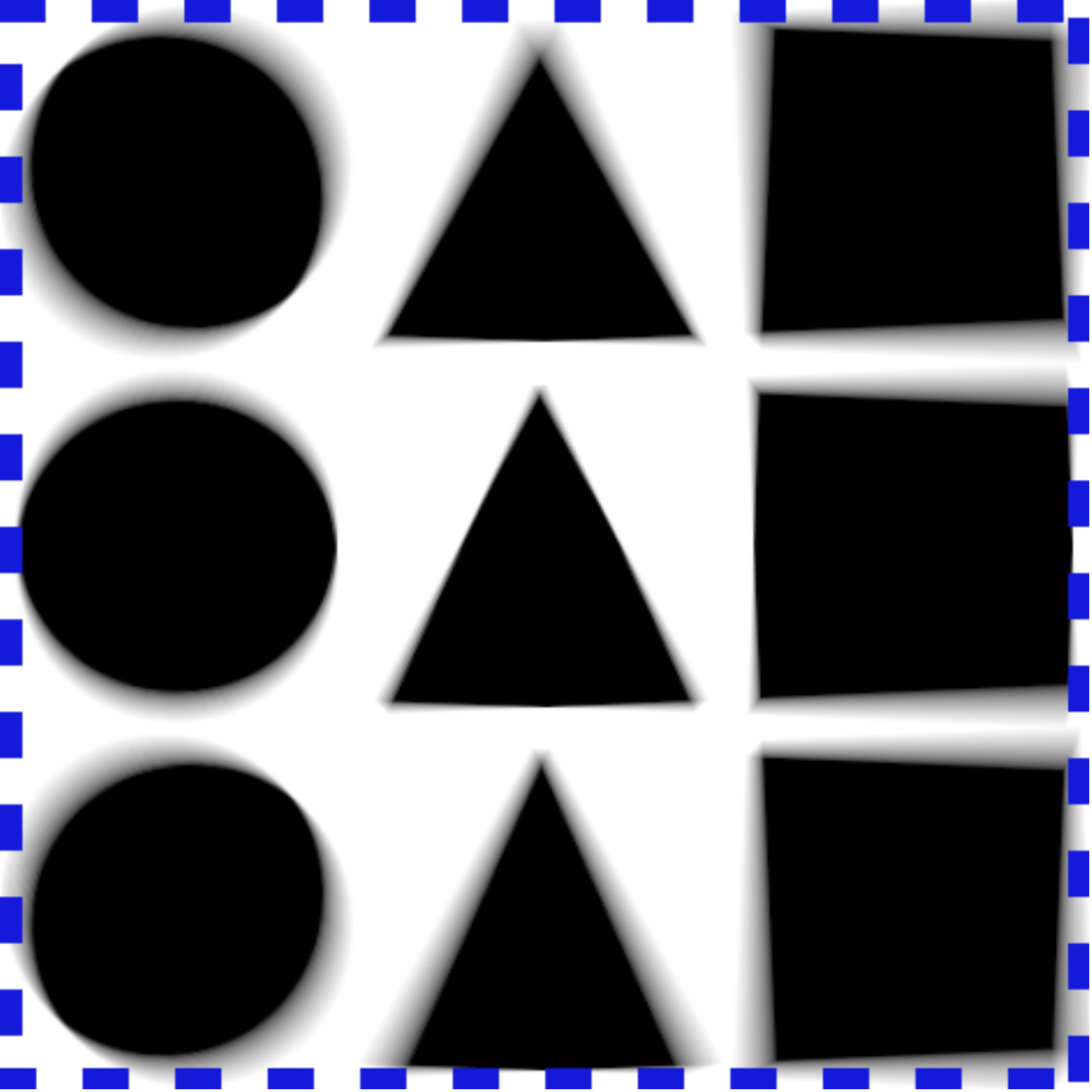}}
\subfloat[\textit{BlurType3}]{
   \includegraphics[width=0.28\linewidth]{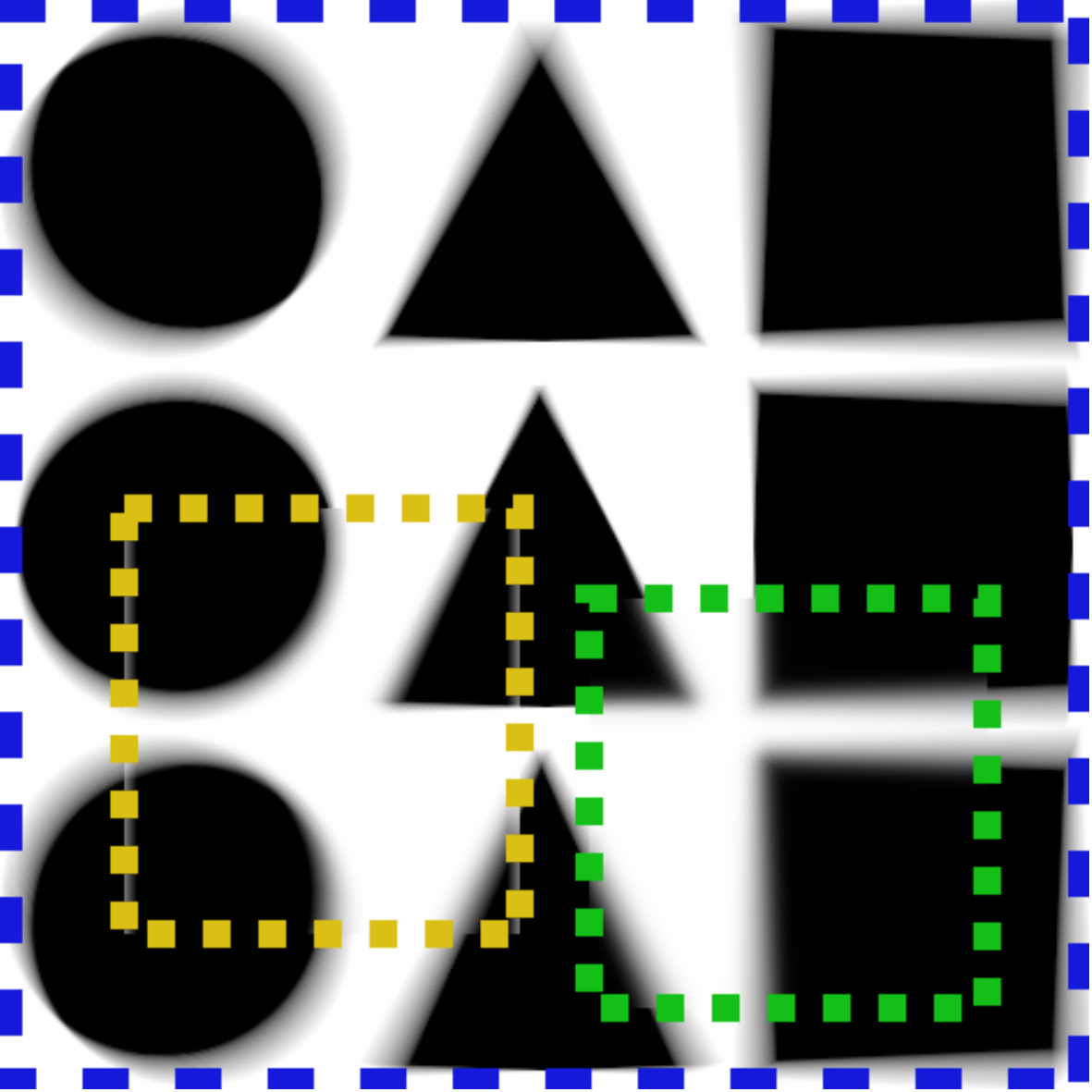}}
   
\subfloat[\textit{BlurType4}]{
   \includegraphics[width=0.28\linewidth]{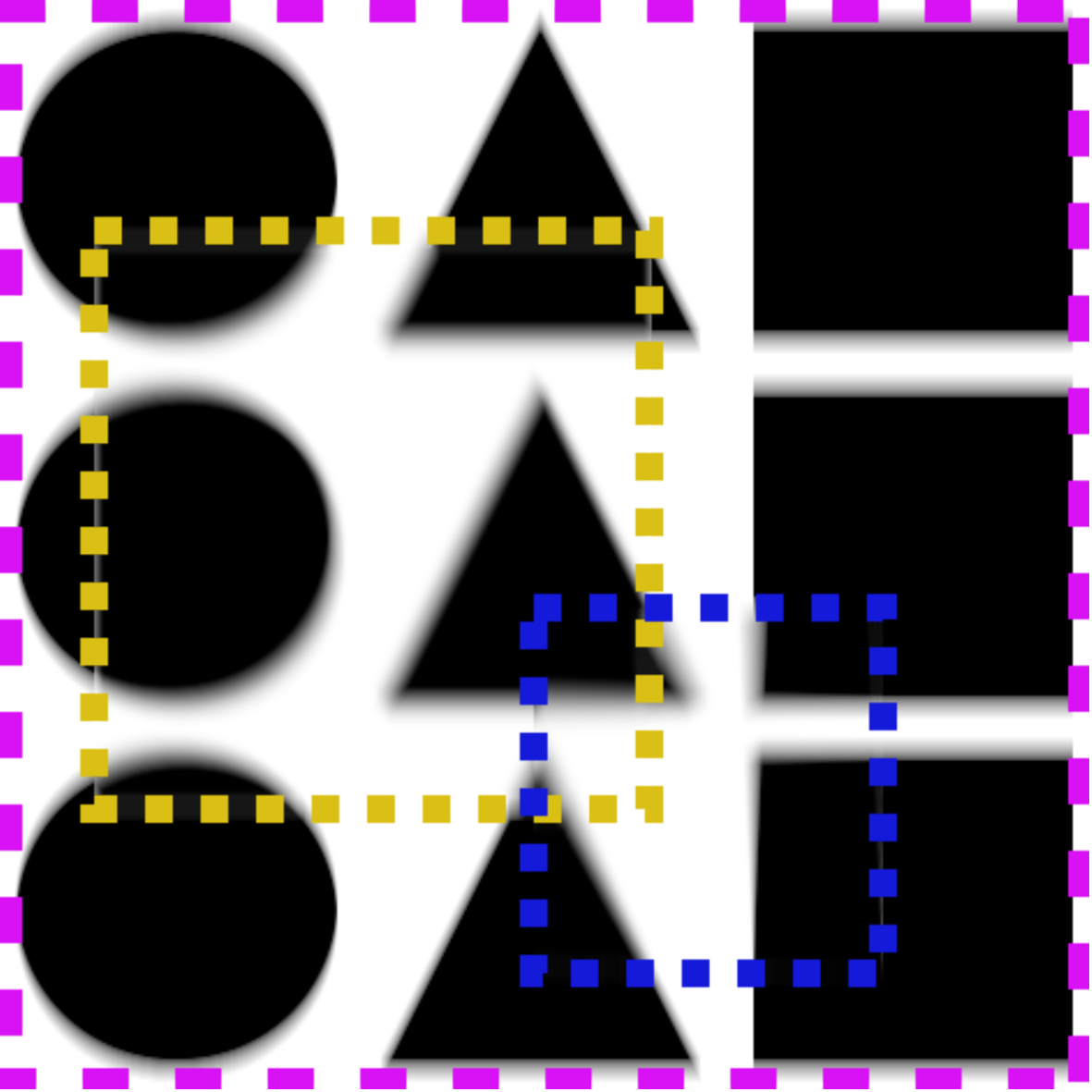}}
\subfloat[\textit{BlurType5}]{
   \includegraphics[width=0.28\linewidth]{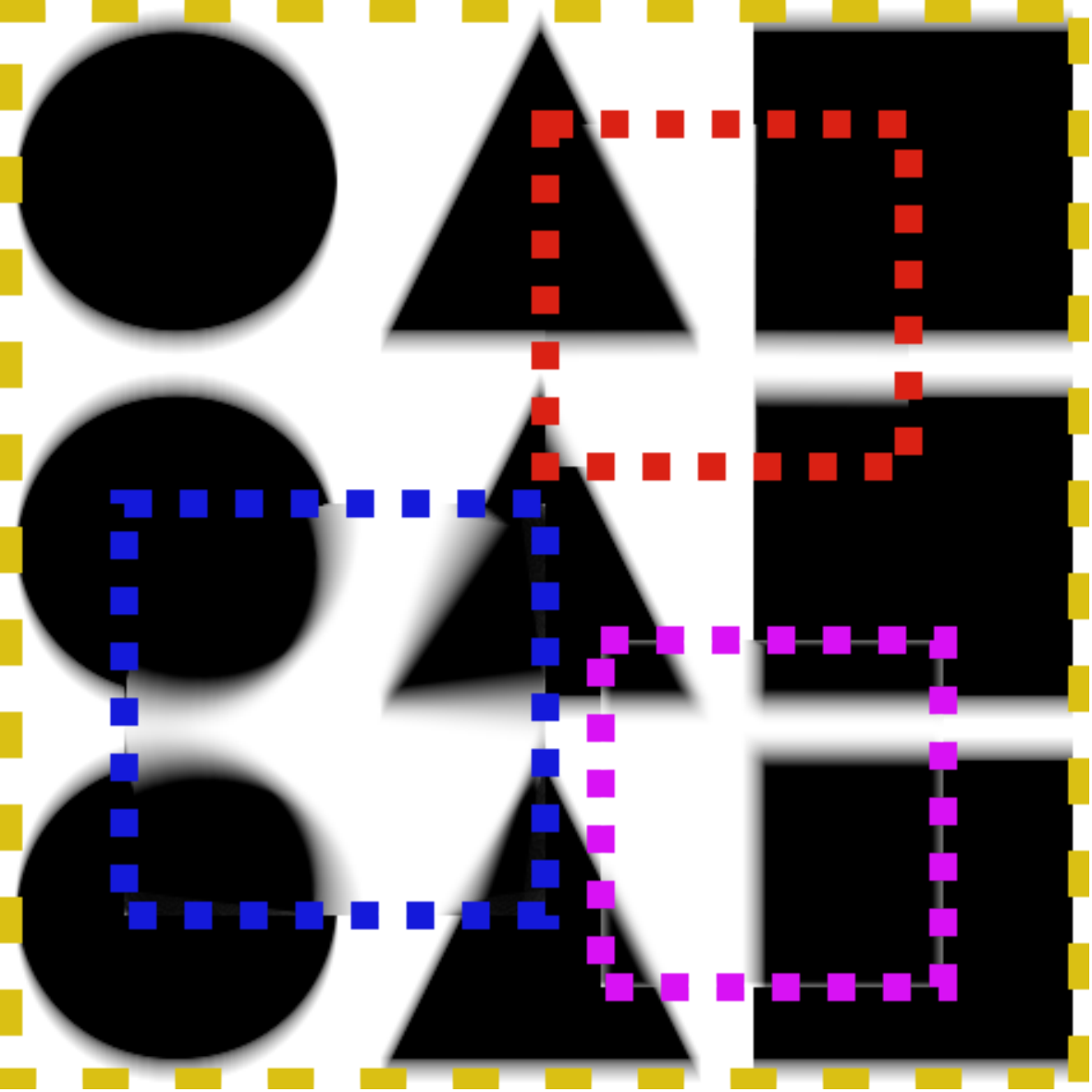}}
\subfloat[\textit{BlurType6}]{
   \includegraphics[width=0.28\linewidth]{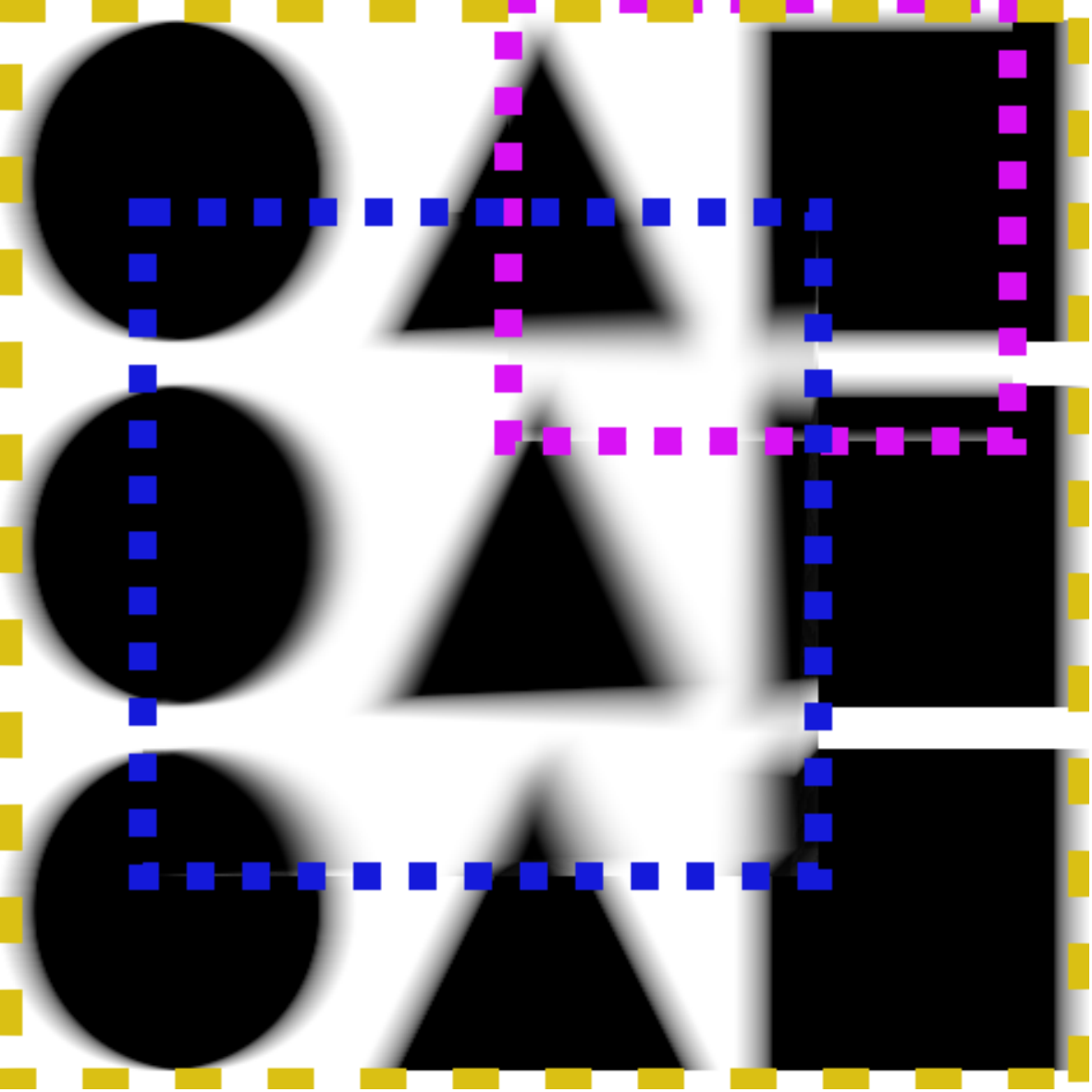}}
   \end{minipage}
 \subfloat{
\includegraphics[angle=-90,origin=b,width=0.15\linewidth]{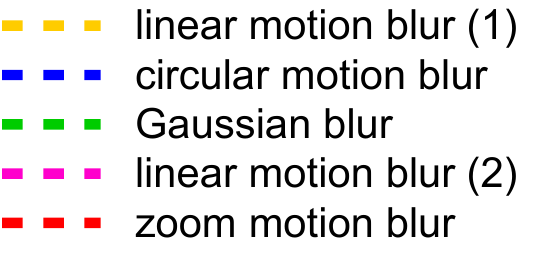}}

\caption{Visualization of the six blur types in Tab. \ref{syntheticTable}. Dotted rectangles in different colors represent the regions corrupted with different types of blur. \textit{BlurType3} to \textit{BlurType6} are complex blur, generated by fusing multiple types of blur. }
\label{fig:BlurVisualization}
\end{figure}

\begin{table*}[htb]
\centering
\caption
{Comparisons of average PSNR/SSIM/MSE on test images \cite{baker2011database} which are corrupted with different types of synthetic blur. \textit{BlurType1} is linear motion blur, \textit{BlurType2} is circular motion blur, and \textit{BlurType3} to \textit{BlurType6} are complex blur mixed with multiple types of blur. See Fig. \ref{fig:BlurVisualization} for illustration.  }
\label{syntheticTable}
\begin{tabular}{llll}
\hline
 & \multicolumn{1}{c}{\textit{BlurType1}}                    & \multicolumn{1}{c}{\textit{BlurType2}}                    & \multicolumn{1}{c}{\textit{BlurType3}}                    \\ \hline
RL  \cite{richardson1972bayesian}    & 17.787 / 0.804 / 1244.060         & 18.766 / 0.822 / 1013.496          & 19.413 / 0.838 / 908.450          \\
Deconvblind   \cite{holmes1995light}      & 21.910 / 0.913 / 522.051         & 22.736 / 0.916 / 416.793          & 22.900 / 0.917 / 405.950          \\
Whyte  \cite{whyte2012non}    &  23.739 / 0.901 / 846.817        &      22.147 / 0.879 / 603.632      &   21.687 / 0.866 / 684.806    \\
\gu{GFRS  \cite{kheradmand2014general}}    &  \gu{12.174 / 0.473 / 4305.645}   &   \gu{13.531 / 0.550 / 3254.577}      &   \gu{14.338 / 0.598 / 2758.658}    \\
\gu{GBBD  \cite{bai2018graph}}    &  \gu{23.823 / 0.921 / 496.064}     &  \gu{23.128 / 0.906 / 409.134}    &   \gu{22.687 / 0.902 / 451.751}   \\
\guThi{Registration-1\cite{farneback2003two}, \cite{buades2005non}}   & \guThi{21.180 / 0.775 / 755.719}  & \guThi{20.565 / 0.769 / 794.853}  & \guThi{21.297 / 0.815 / 702.828} \\
\guThi{Registration-2\cite{artin2016geometric,buades2005non}}  & \guThi{21.667 / 0.860 / 578.877}  & \guThi{20.178 / 0.825 / 813.343}  & \guThi{20.257 / 0.817 / 802.075}  \\
SBD-single \cite{zhang2013multi}   & 25.354 / 0.956 / 253.032         & 24.313 / 0.932 / 276.382          & 24.199 / 0.931 / 286.646          \\
SBD-multi \cite{zhang2013multi}   & 12.354 / 0.471 /  4356.241        & 12.728 / 0.499 / 3822.419         & 11.833 / 0.452 / 4521.26        \\
GCRL+DL  \cite{yuan2007image}      &  20.939 / 0.883 / 614.729&      21.578 / 0.877 / 524.715    &       21.930 / 0.877 / 526.811  \\
OGMM                 & \textbf{27.905 / 0.966 / 144.208}   & 25.046 / 0.944 / 230.982          & 25.295 / 0.947 / 225.836          \\
OGMM+DL              & 27.716 / 0.966 / 141.942 & \textbf{25.133 / 0.945 / 226.747} & \textbf{25.410 / 0.948 / 220.014} \\ \hline
                     & \multicolumn{1}{c}{\textit{BlurType4}}                    & \multicolumn{1}{c}{\textit{BlurType5}}                    & \multicolumn{1}{c}{\textit{BlurType6}}                    \\ \hline
RL   \cite{richardson1972bayesian}    & 19.863 / 0.856 / 826.026          & 19.499 / 0.845 / 891.206          &   19.357 / 0.833 / 886.731                                \\
Deconvblind  \cite{holmes1995light}       & 23.606 / 0.931 / 354.657          & 23.153 / 0.924 / 381.380          & 22.765 / 0.914 / 410.282      \\
Whyte  \cite{whyte2012non}     &   22.346 / 0.876 / 786.165             &  22.634 / 0.886 / 656.215           &     20.770 / 0.857 / 1011.524         \\
\gu{GFRS  \cite{kheradmand2014general}}    &  \gu{14.145 / 0.594 / 2887.265}        &      \gu{13.753 / 0.570 / 3168.338}      &   \gu{14.526 / 0.598 / 2602.047}    \\
\gu{GBBD  \cite{bai2018graph}}    &  \gu{23.658 / 0.907 / 402.519}   &      \gu{22.145 / 0.879 / 502.416}      &   \gu{19.950 / 0.806 / 731.409}  \\
\guThi{Registration-1\cite{farneback2003two}, \cite{buades2005non}}   & \guThi{21.917 / 0.822 / 659.675}  & \guThi{21.926 / 0.823 / 662.619}  & \guThi{21.033 / 0.812 / 728.168} \\
\guThi{Registration-2\cite{artin2016geometric,buades2005non}} & \guThi{21.761 / 0.881 / 514.607}  & \guThi{20.385 / 0.840 / 704.858}  & \guThi{18.314 / 0.771 / 1122.865}  \\
SBD-single \cite{zhang2013multi}    & 24.927 / 0.947 / 241.660          & 24.629 / 0.942 / 255.831          & 23.644 / 0.920 / 340.248       \\
SBD-multi \cite{zhang2013multi}   & 11.537 / 0.446 / 4748.240         & 11.938 / 0.470 / 4348.459         &   9.895 / 0.333 / 6891.413\\
GCRL+DL \cite{yuan2007image}      &    21.798 / 0.884 / 504.584      &      21.722 / 0.881 / 515.390     &   21.494 / 0.874 / 539.748     \\
OGMM                 & 26.277 / 0.958 / 188.181          & 25.937 / 0.954 / 203.715          &    24.724 / 0.940 / 257.072                               \\
OGMM+DL              & \textbf{26.453 / 0.959 / 180.985} & \textbf{26.087 / 0.955 / 196.959} &  \textbf{24.811 / 0.942 / 252.102}                                 \\ \hline
\end{tabular}
\end{table*}


\begin{figure*}
\centering
{
\includegraphics[width=0.13\linewidth]{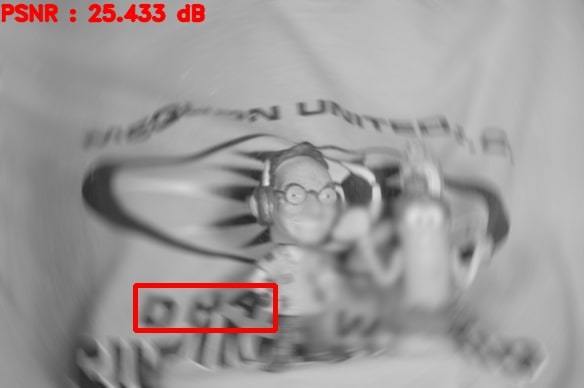}
\includegraphics[width=0.13\linewidth]{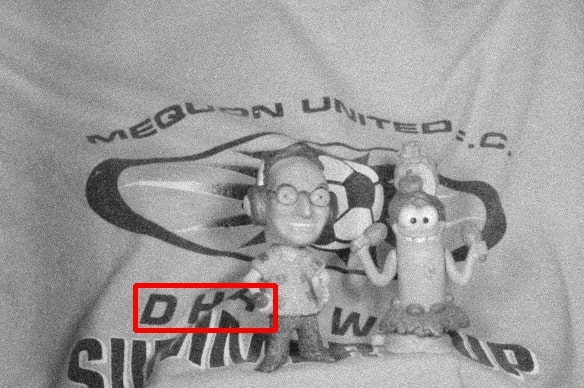}
\includegraphics[width=0.13\linewidth]{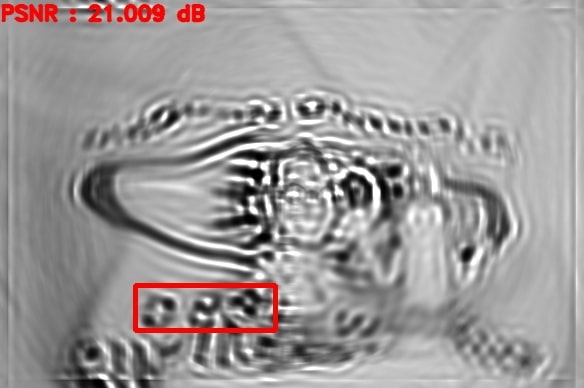}
\includegraphics[width=0.13\linewidth]{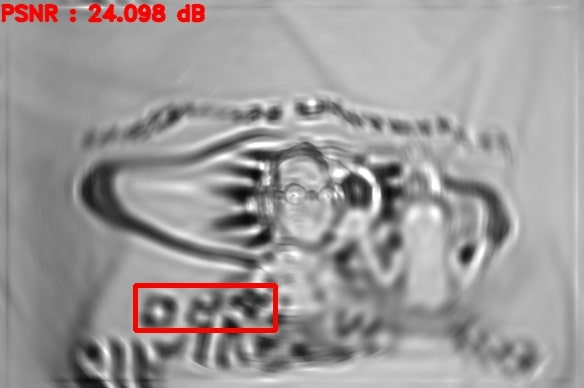}
\includegraphics[width=0.13\linewidth]{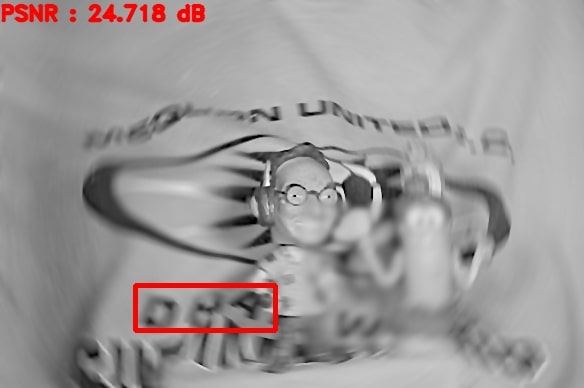}
\includegraphics[width=0.13\linewidth]{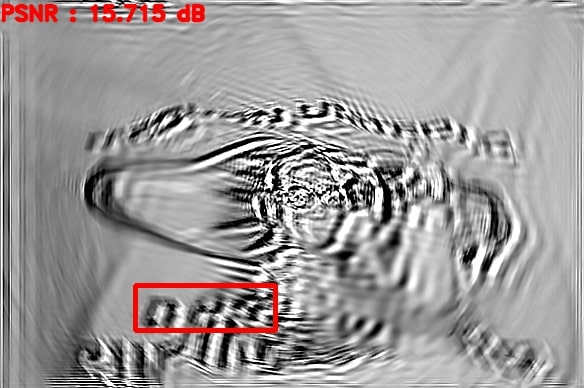}
\includegraphics[width=0.13\linewidth]{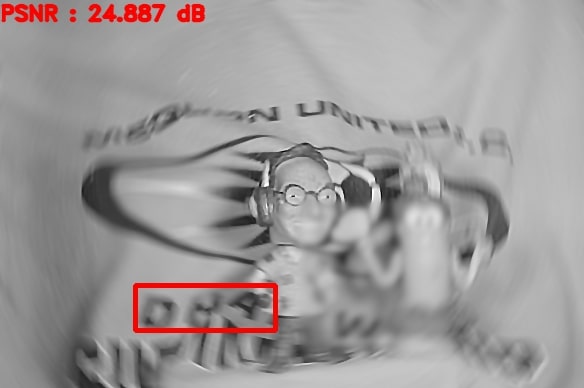}
}\\[-2.1ex]
\setcounter{subfigure}{0}
{\subfloat[Blurred image]{
\includegraphics[width=0.13\linewidth]{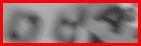}}
\subfloat[Noisy image]{
\includegraphics[width=0.13\linewidth]{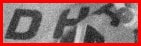}}
\subfloat[RL \cite{richardson1972bayesian}]{
\includegraphics[width=0.13\linewidth]{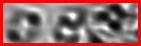}}
\subfloat[Deconvblind \cite{holmes1995light}]{
\includegraphics[width=0.13\linewidth]{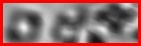}}
\subfloat[Whyte \cite{whyte2012non}]{
\includegraphics[width=0.13\linewidth]{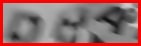}}
\subfloat[\gu{GFRS \cite{kheradmand2014general}}]{
\includegraphics[width=0.13\linewidth]{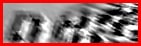}}
\subfloat[\gu{GBBD \cite{bai2018graph}}]{
\includegraphics[width=0.13\linewidth]{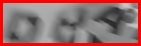}}\hspace{0.3mm}
}

{
\includegraphics[width=0.13\linewidth]{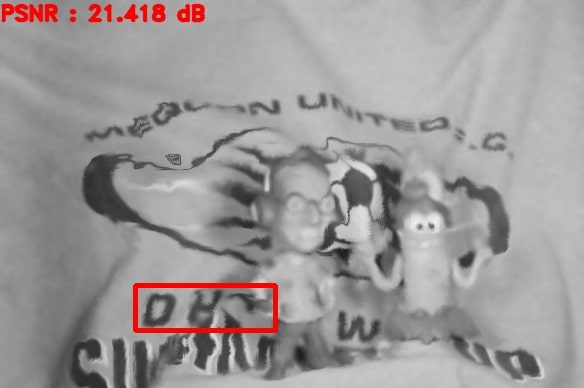}
\includegraphics[width=0.13\linewidth]{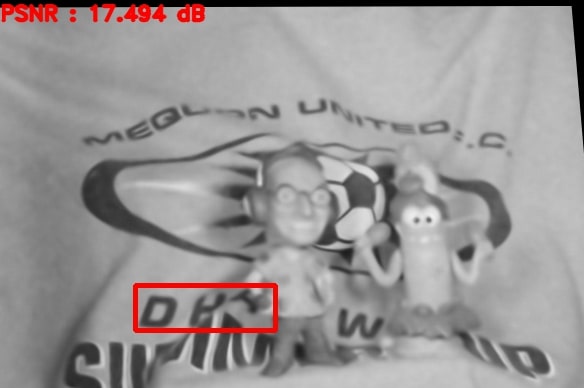}
\includegraphics[width=0.13\linewidth]{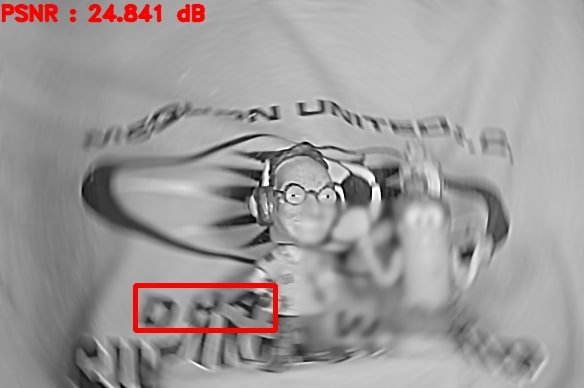}
\includegraphics[width=0.13\linewidth]{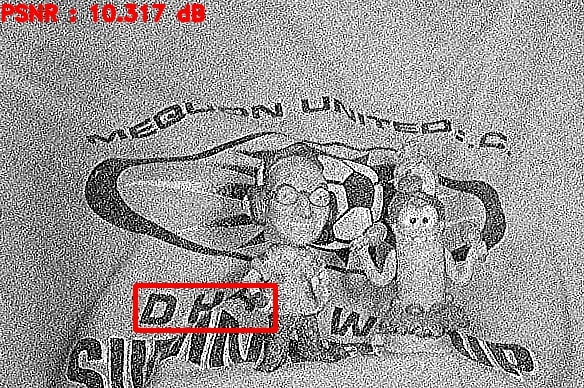}
\includegraphics[width=0.13\linewidth]{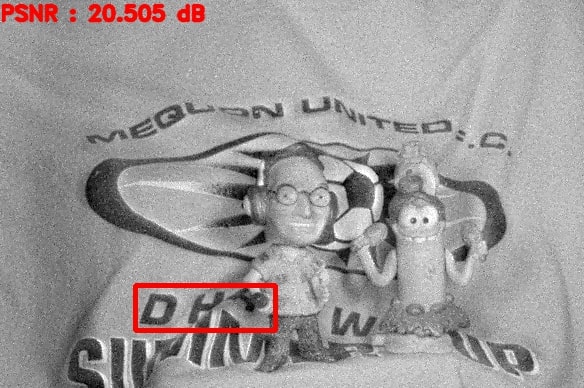}
\includegraphics[width=0.13\linewidth]{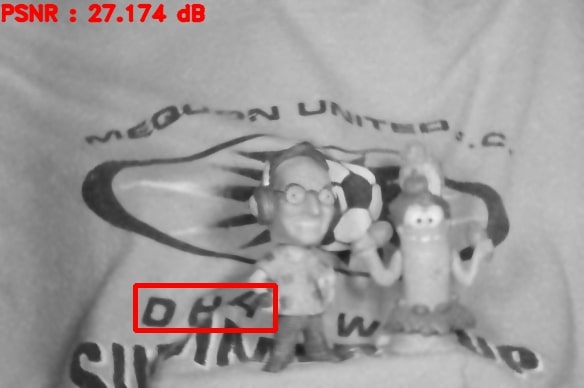}
\includegraphics[width=0.13\linewidth]{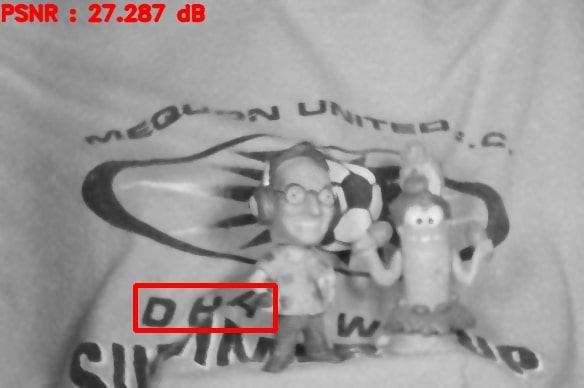}}\\[-2.1ex]

\subfloat[\guThi{Registration-1}]{
\includegraphics[width=0.13\linewidth]{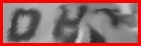}}
\subfloat[\guThi{Registration-2}]{
\includegraphics[width=0.13\linewidth]{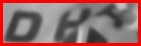}}
\subfloat[SBD-single \cite{zhang2013multi}]{
\includegraphics[width=0.13\linewidth]{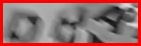}}
\subfloat[SBD-multi \cite{zhang2013multi}]{
\includegraphics[width=0.13\linewidth]{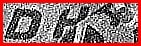}}
\subfloat[GCRL+DL \cite{yuan2007image}]{
\includegraphics[width=0.13\linewidth]{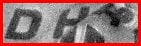}}
\subfloat[OGMM]{
\includegraphics[width=0.13\linewidth]{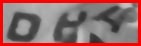}}
\subfloat[OGMM+DL]{
\includegraphics[width=0.13\linewidth]{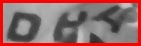}}\hspace{1.3mm}

   \caption{Visual comparison on image \textit{Mequon} in dataset \cite{baker2011database} with synthetic blur (\textit{BlurType3} in Fig. \ref{fig:BlurVisualization}). (b) is added with Gaussian noise ($\sigma = 10)$. }
\label{fig:ComplexBlurComparison1}
\end{figure*}

\begin{figure*}
\centering
{
\includegraphics[width=0.13\linewidth]{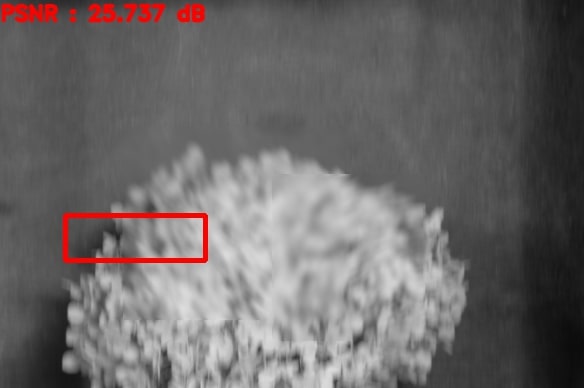}
\includegraphics[width=0.13\linewidth]{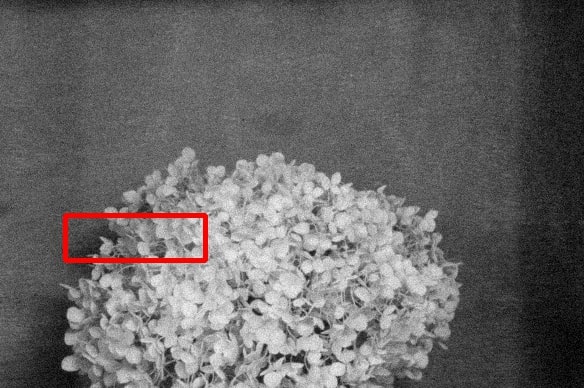}
\includegraphics[width=0.13\linewidth]{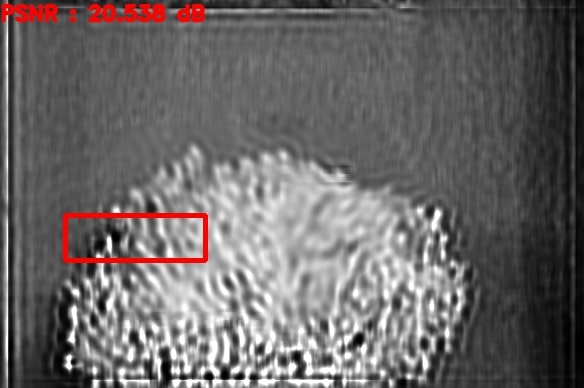}
\includegraphics[width=0.13\linewidth]{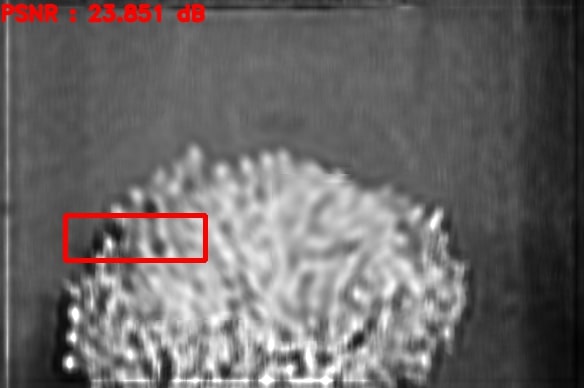}
\includegraphics[width=0.13\linewidth]{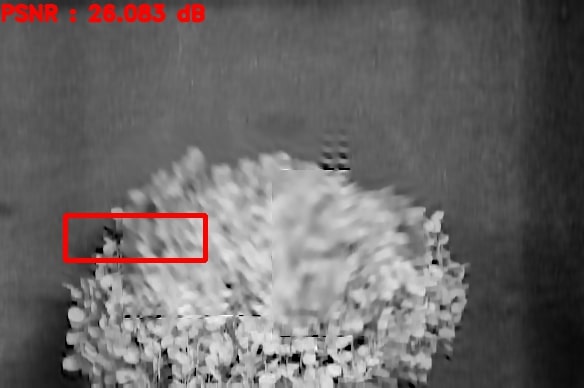}
\includegraphics[width=0.13\linewidth]{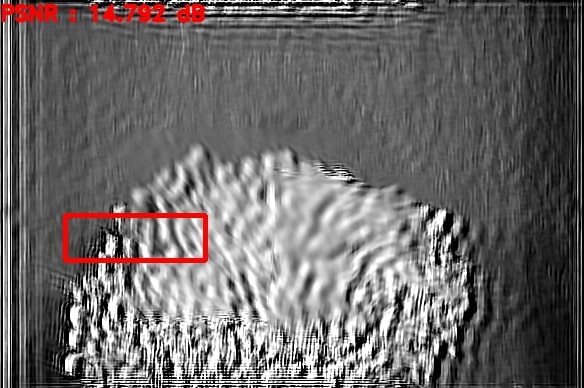}
\includegraphics[width=0.13\linewidth]{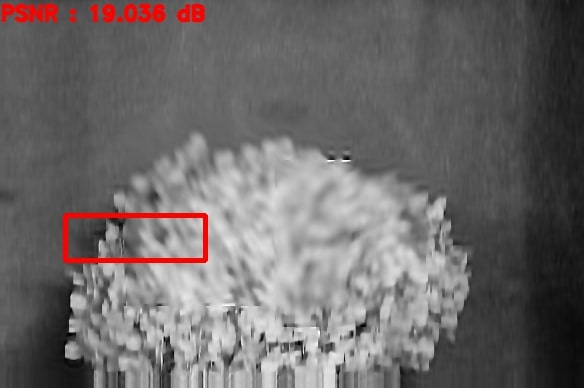}}
\\[-2.1ex]
\setcounter{subfigure}{0}
{\hspace{-1.2mm}\subfloat[Blurred image]{
\includegraphics[width=0.13\linewidth]{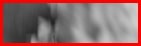}}
\subfloat[Noisy image]{
\includegraphics[width=0.13\linewidth]{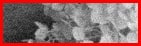}}
\subfloat[RL \cite{richardson1972bayesian}]{
\includegraphics[width=0.13\linewidth]{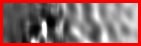}}
\subfloat[Deconvblind \cite{holmes1995light}]{
\includegraphics[width=0.13\linewidth]{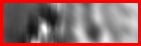}}
\subfloat[Whyte \cite{whyte2012non}]{
\includegraphics[width=0.13\linewidth]{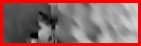}}
\subfloat[\gu{GFRS \cite{kheradmand2014general}}]{
\includegraphics[width=0.13\linewidth]{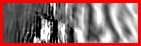}}
\subfloat[\gu{GBBD \cite{bai2018graph}}]{
\includegraphics[width=0.13\linewidth]{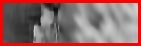}}
}
{
\includegraphics[width=0.13\linewidth]{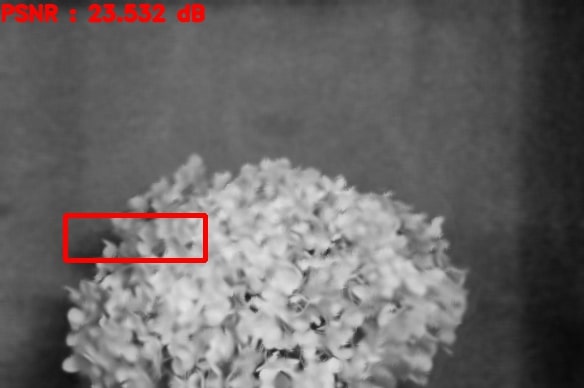}
\includegraphics[width=0.13\linewidth]{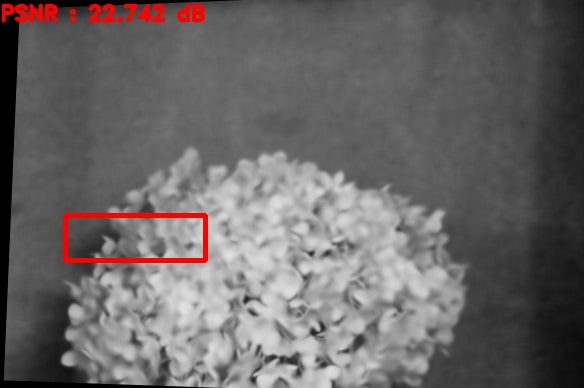}
\includegraphics[width=0.13\linewidth]{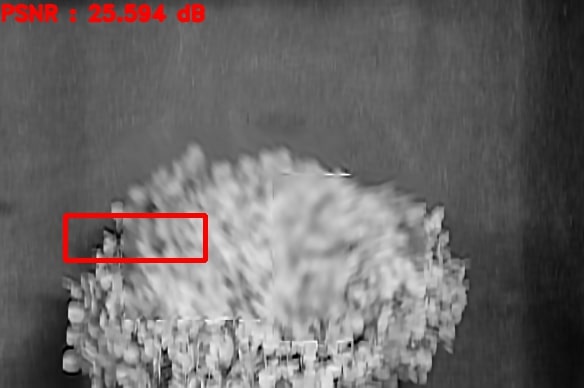}
\includegraphics[width=0.13\linewidth]{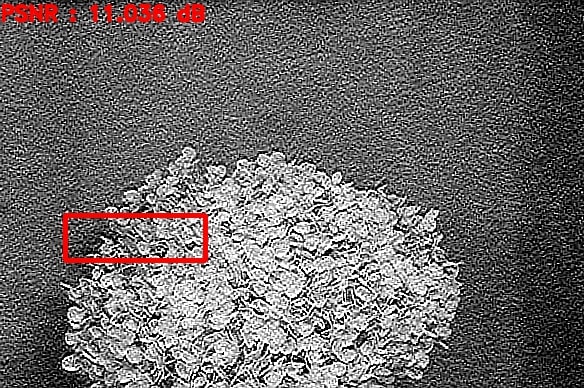}
\includegraphics[width=0.13\linewidth]{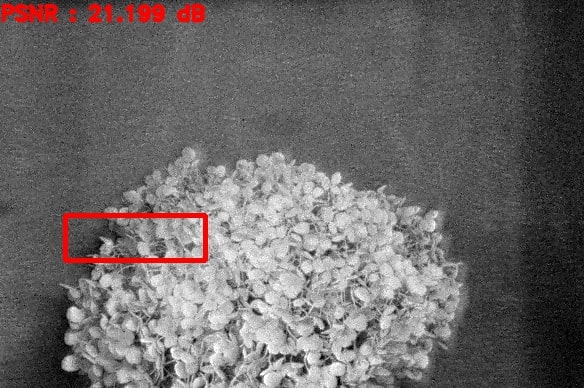}
\includegraphics[width=0.13\linewidth]{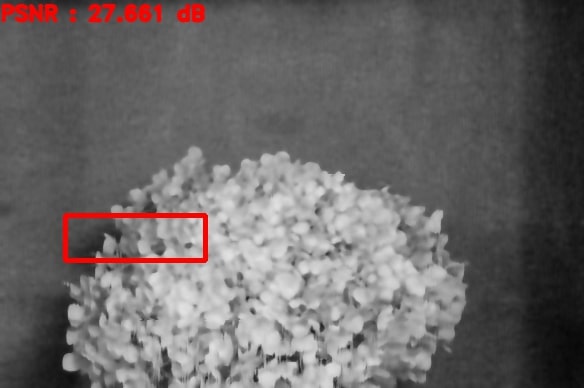}
\includegraphics[width=0.13\linewidth]{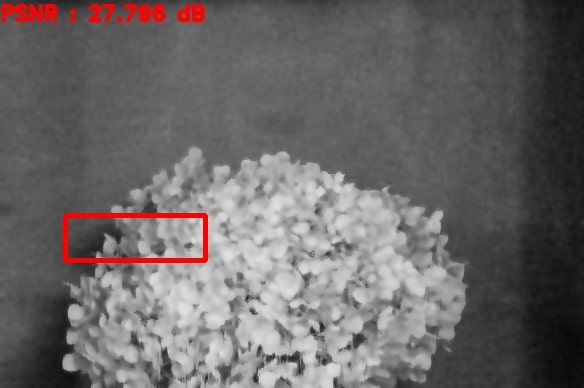}
}\\[-2.1ex]

\subfloat[\guThi{Registration-1}]{
\includegraphics[width=0.13\linewidth]{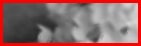}}
\subfloat[\guThi{Registration-2}]{
\includegraphics[width=0.13\linewidth]{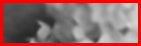}}
\subfloat[SBD-single \cite{zhang2013multi}]{
\includegraphics[width=0.13\linewidth]{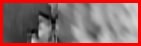}}
\subfloat[SBD-multi \cite{zhang2013multi}]{
\includegraphics[width=0.13\linewidth]{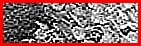}}
\subfloat[GCRL+DL \cite{yuan2007image}]{
\includegraphics[width=0.13\linewidth]{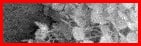}}
\subfloat[OGMM]{
\includegraphics[width=0.13\linewidth]{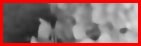}}
\subfloat[OGMM+DL]{
\includegraphics[width=0.13\linewidth]{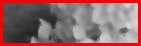}}\hspace{1.3mm}

   \caption{Visual comparison on image \textit{Hydrangea} in dataset \cite{baker2011database} with synthetic blur (\textit{BlurType4} in Fig. \ref{fig:BlurVisualization}). (b) is added with Gaussian noise ($\sigma = 10)$. }
\label{fig:ComplexBlurComparison2}
\end{figure*}

\begin{figure*}[tbh]
\centering
{
   \includegraphics[width=0.13\linewidth]{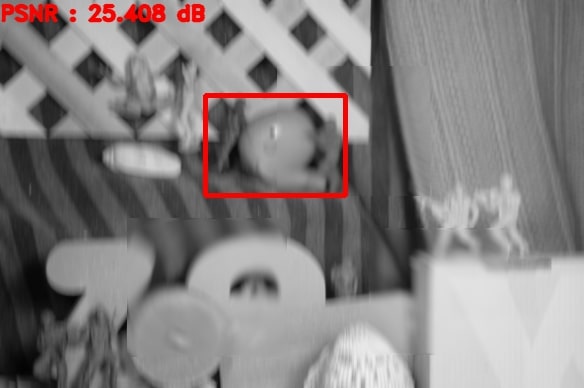}
   \includegraphics[width=0.13\linewidth]{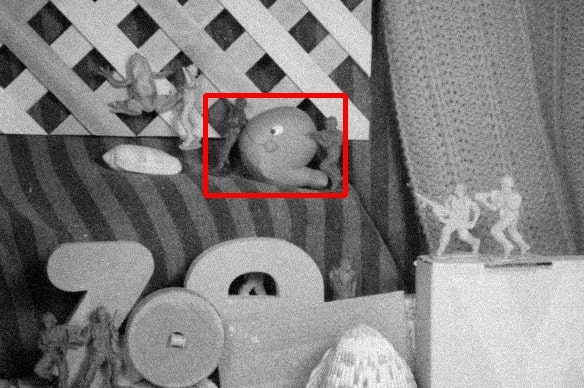}
   \includegraphics[width=0.13\linewidth]{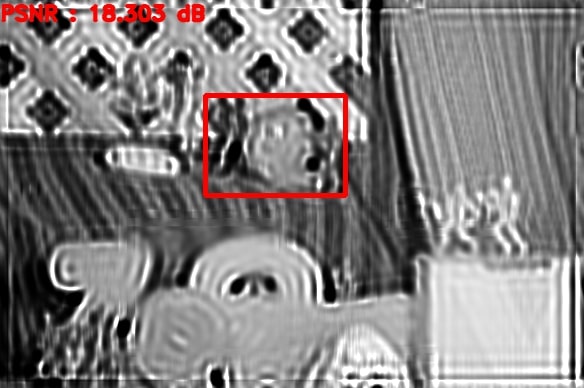}
   \includegraphics[width=0.13\linewidth]{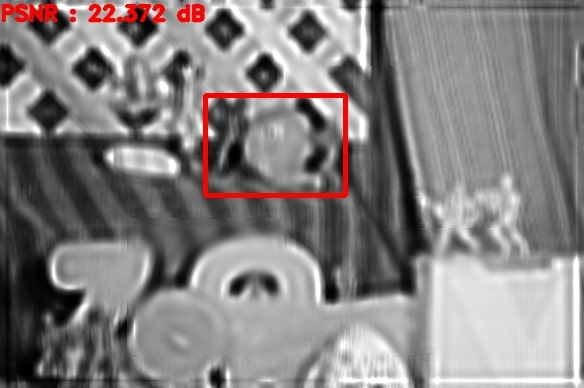}
   \includegraphics[width=0.13\linewidth]{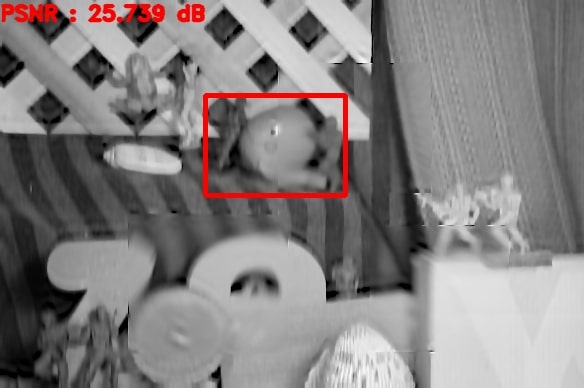}
   \includegraphics[width=0.13\linewidth]{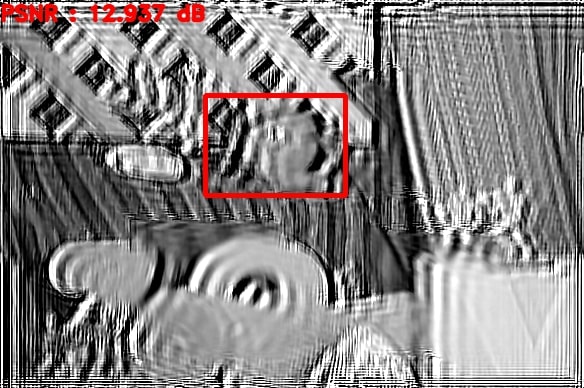}
   \includegraphics[width=0.13\linewidth]{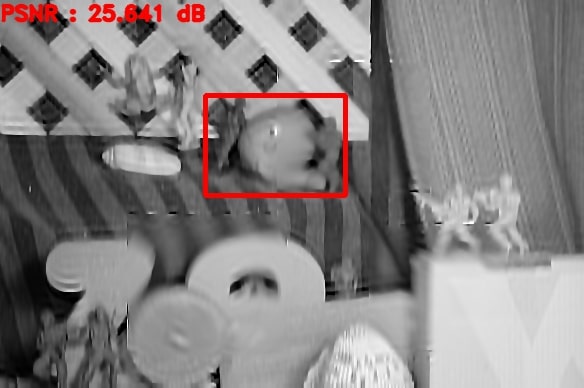}
}\\[-2.1ex]
\setcounter{subfigure}{0}
{\subfloat[Blurred image]{
\includegraphics[width=0.13\linewidth]{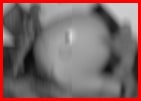}}
\subfloat[Noisy image]{
\includegraphics[width=0.13\linewidth]{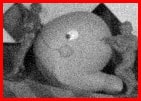}}
\subfloat[RL \cite{richardson1972bayesian}]{
\includegraphics[width=0.13\linewidth]{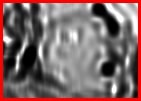}}
\subfloat[Deconvblind \cite{holmes1995light}]{
\includegraphics[width=0.13\linewidth]{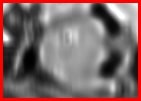}}
\subfloat[Whyte \cite{whyte2012non}]{
\includegraphics[width=0.13\linewidth]{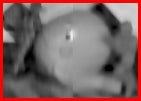}}
\subfloat[\gu{GFRS \cite{kheradmand2014general}}]{
\includegraphics[width=0.13\linewidth]{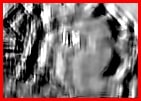}}
\subfloat[\gu{GBBD \cite{bai2018graph}}]{
\includegraphics[width=0.13\linewidth]{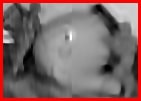}}\hspace{0.2mm}
}

{
\includegraphics[width=0.13\linewidth]{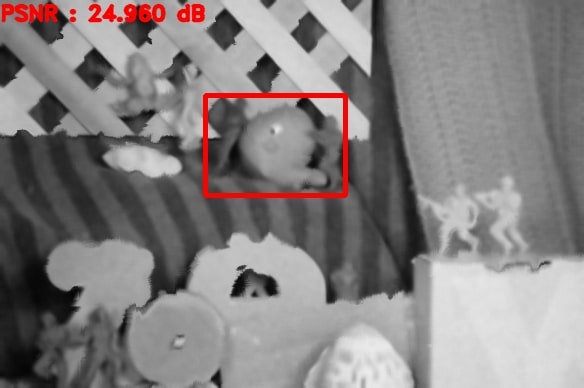}
\includegraphics[width=0.13\linewidth]{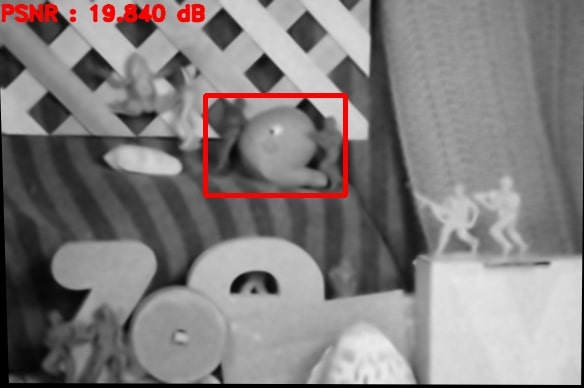}
\includegraphics[width=0.13\linewidth]{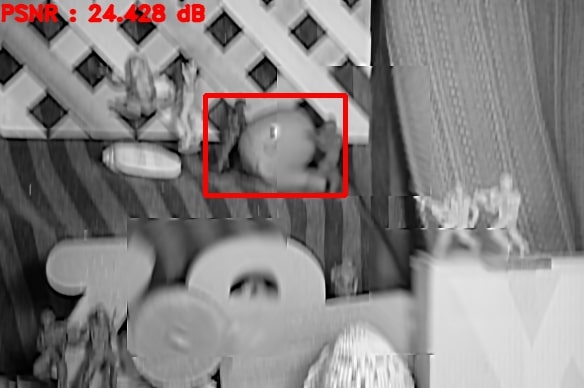}
\includegraphics[width=0.13\linewidth]{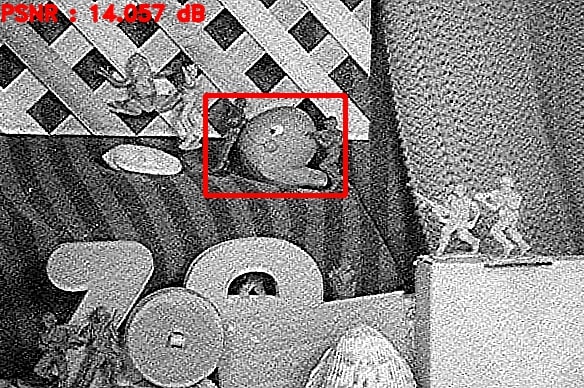}
\includegraphics[width=0.13\linewidth]{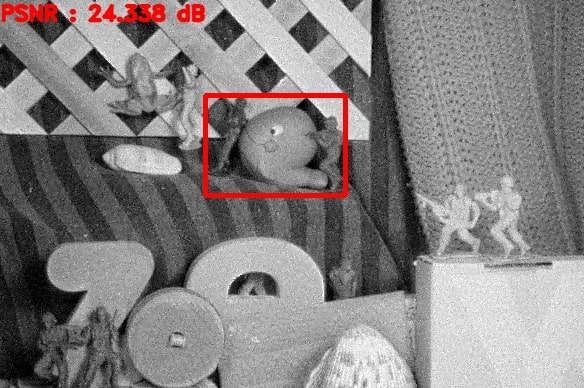}
\includegraphics[width=0.13\linewidth]{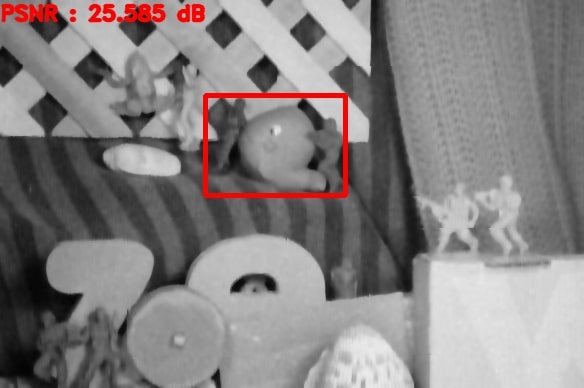}
\includegraphics[width=0.13\linewidth]{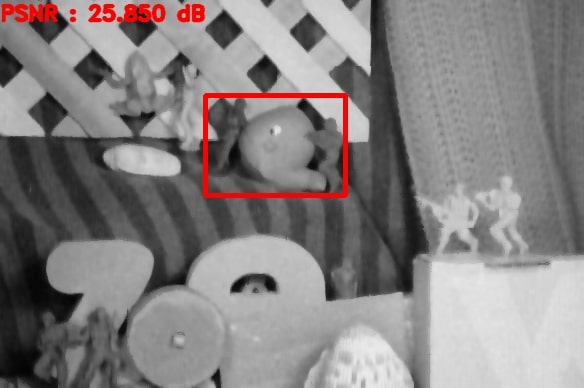}
}\\[-2.1ex]

\subfloat[\guThi{Registration-1}]{
\includegraphics[width=0.13\linewidth]{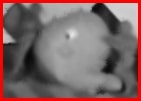}}
\subfloat[\guThi{Registration-2}]{
\includegraphics[width=0.13\linewidth]{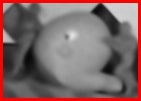}}
\subfloat[SBD-single \cite{zhang2013multi}]{
\includegraphics[width=0.13\linewidth]{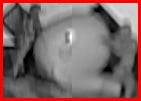}}
\subfloat[SBD-multi \cite{zhang2013multi}]{
\includegraphics[width=0.13\linewidth]{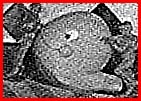}}
\subfloat[GCRL+DL \cite{yuan2007image}]{
\includegraphics[width=0.13\linewidth]{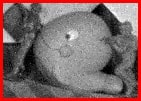}}
\subfloat[OGMM]{
\includegraphics[width=0.13\linewidth]{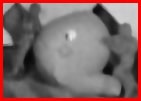}}
\subfloat[OGMM+DL]{
\includegraphics[width=0.13\linewidth]{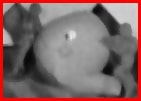}}\hspace{1.3mm}

   \caption{Visual comparison on image \textit{Army} in dataset \cite{baker2011database} with synthetic blur (\textit{BlurType5} in Fig. \ref{fig:BlurVisualization}). (b) is added with Gaussian noise ($\sigma = 10)$. }
\label{fig:ComplexBlurComparison3}
\end{figure*}

\begin{figure*}
\centering
{
\includegraphics[width=0.13\linewidth]{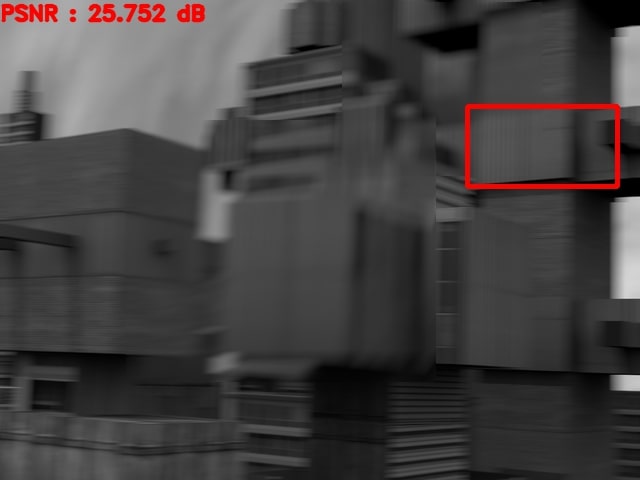}
\includegraphics[width=0.13\linewidth]{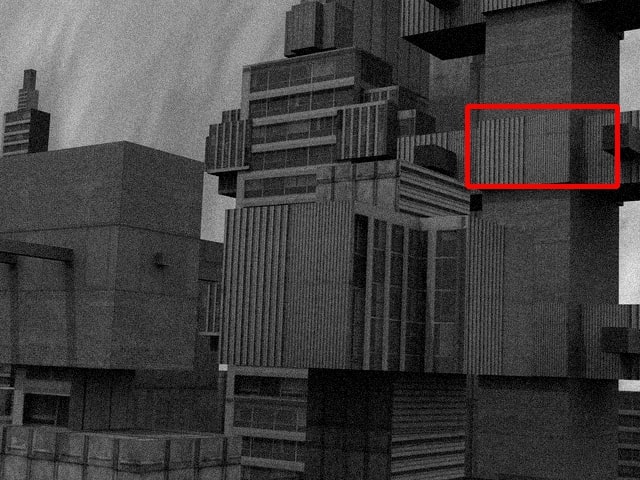}
\includegraphics[width=0.13\linewidth]{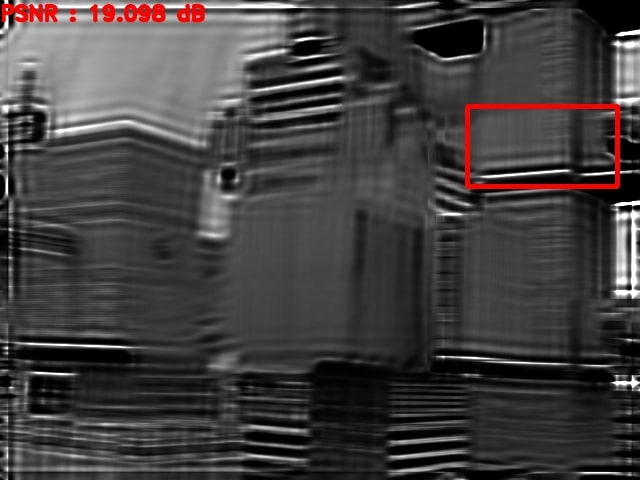}
\includegraphics[width=0.13\linewidth]{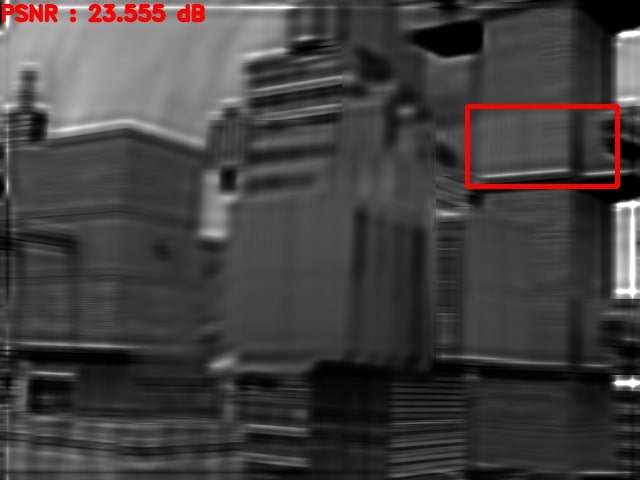}
\includegraphics[width=0.13\linewidth]{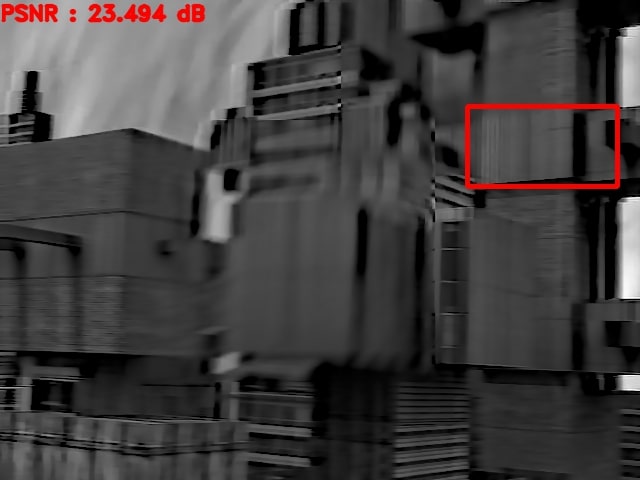}
\includegraphics[width=0.13\linewidth]{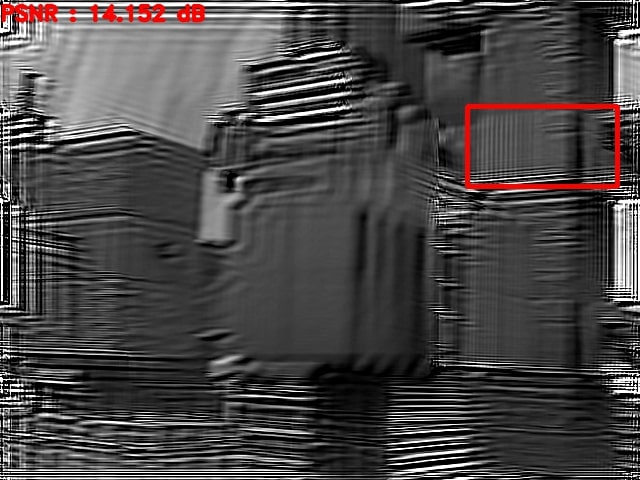}
\includegraphics[width=0.13\linewidth]{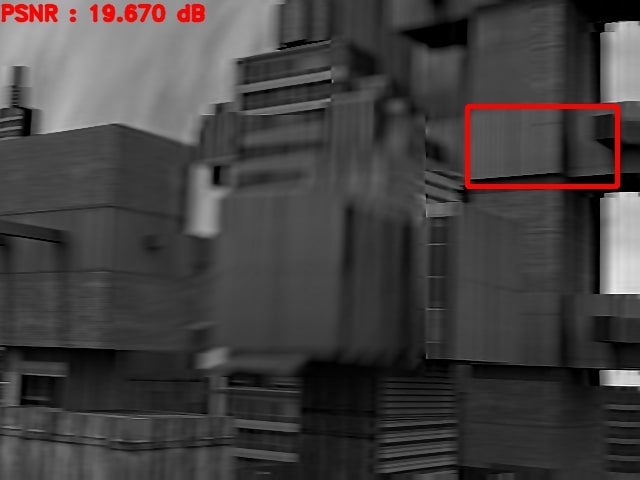}
}\\[-2.1ex]
\setcounter{subfigure}{0}
{\subfloat[Blurred image]{
\includegraphics[width=0.13\linewidth]{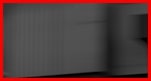}}
\subfloat[Noisy image]{
\includegraphics[width=0.13\linewidth]{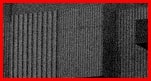}}
\subfloat[RL \cite{richardson1972bayesian}]{
\includegraphics[width=0.13\linewidth]{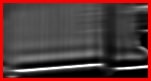}}
\subfloat[Deconvblind \cite{holmes1995light}]{
\includegraphics[width=0.13\linewidth]{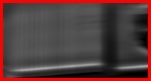}}
\subfloat[Whyte \cite{whyte2012non}]{
\includegraphics[width=0.13\linewidth]{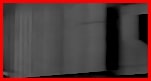}}
\subfloat[\gu{GFRS \cite{kheradmand2014general}}]{
\includegraphics[width=0.13\linewidth]{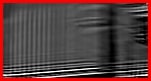}}
\subfloat[\gu{GBBD \cite{bai2018graph}}]{
\includegraphics[width=0.13\linewidth]{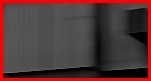}}\hspace{0.3mm}
}

{
\includegraphics[width=0.13\linewidth]{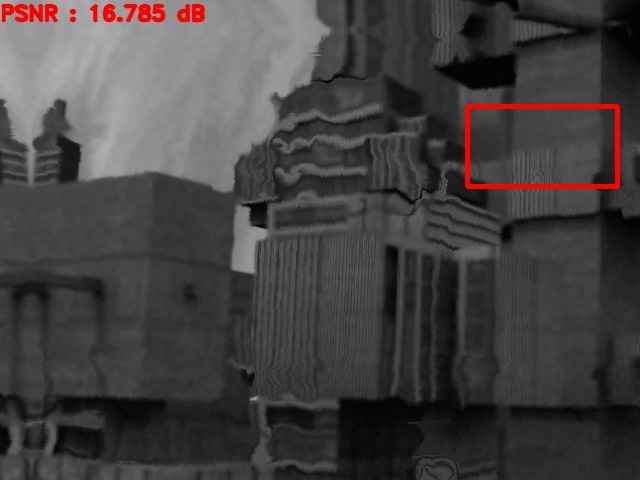}
\includegraphics[width=0.13\linewidth]{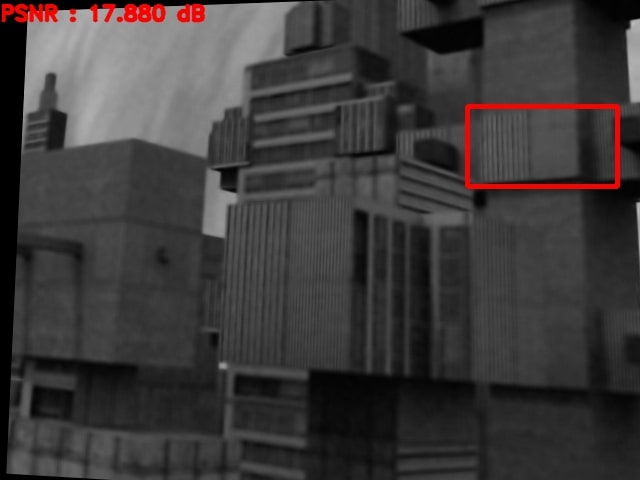}
\includegraphics[width=0.13\linewidth]{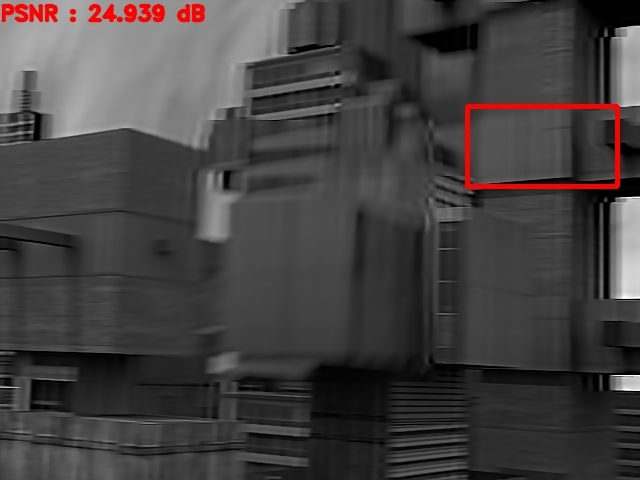}
\includegraphics[width=0.13\linewidth]{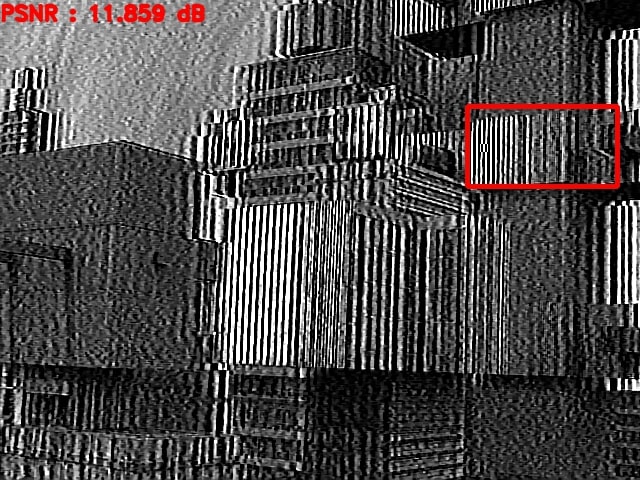}
\includegraphics[width=0.13\linewidth]{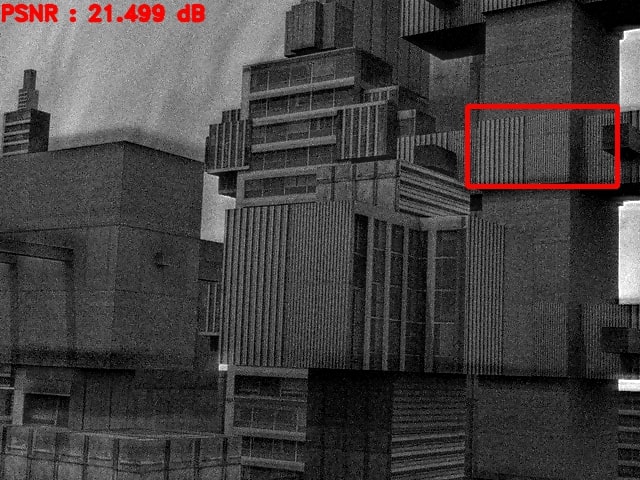}
\includegraphics[width=0.13\linewidth]{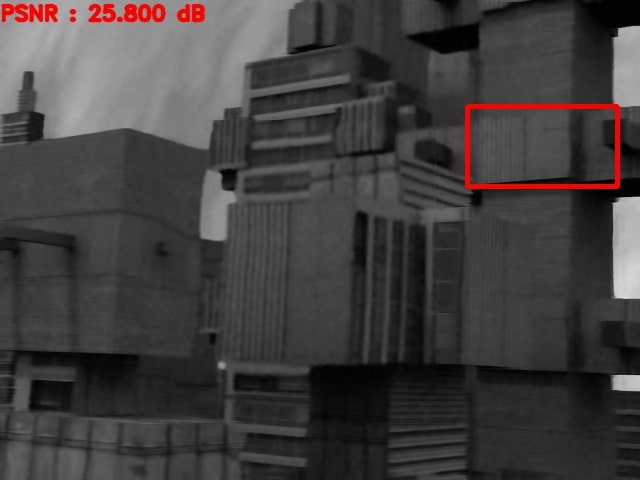}
\includegraphics[width=0.13\linewidth]{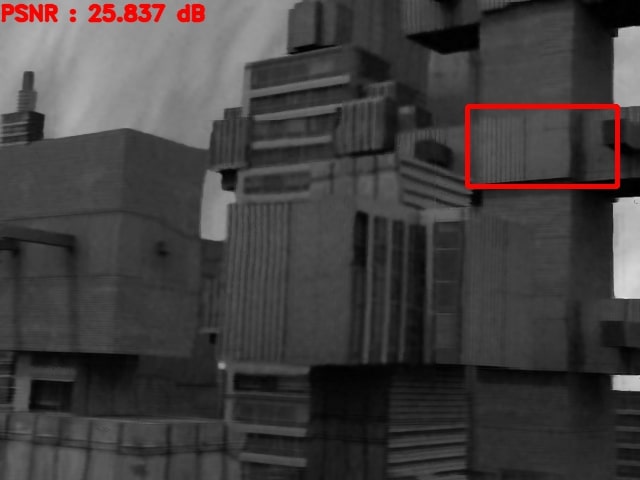}
}\\[-2.1ex]

\subfloat[\guThi{Registration-1}]{
\includegraphics[width=0.13\linewidth]{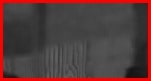}}
\subfloat[\guThi{Registration-2}]{
\includegraphics[width=0.13\linewidth]{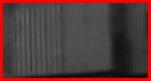}}
\subfloat[SBD-single \cite{zhang2013multi}]{
\includegraphics[width=0.13\linewidth]{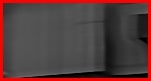}}
\subfloat[SBD-multi \cite{zhang2013multi}]{
\includegraphics[width=0.13\linewidth]{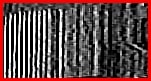}}
\subfloat[GCRL+DL \cite{yuan2007image}]{
\includegraphics[width=0.13\linewidth]{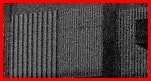}}
\subfloat[OGMM]{
\includegraphics[width=0.13\linewidth]{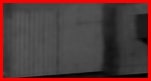}}
\subfloat[OGMM+DL]{
\includegraphics[width=0.13\linewidth]{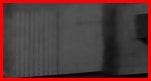}}\hspace{1.3mm}

   \caption{Visual comparison on image \textit{Urban} in dataset \cite{baker2011database} with synthetic blur (\textit{BlurType6} in Fig. \ref{fig:BlurVisualization}). (b) is added with Gaussian noise ($\sigma = 10)$. }
\label{fig:ComplexBlurComparison4}
\end{figure*}

\begin{figure*}[htb]
\begin{center}
\subfloat{
   \includegraphics[width=0.15\linewidth]{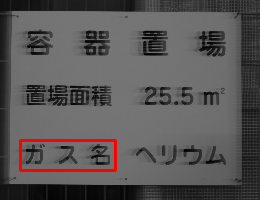}
   \includegraphics[width=0.15\linewidth]{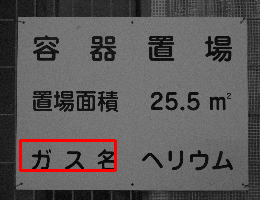}
   \includegraphics[width=0.15\linewidth]{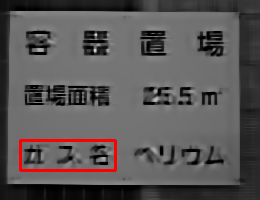}
   \includegraphics[width=0.15\linewidth]{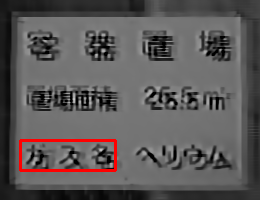}
   \includegraphics[width=0.15\linewidth]{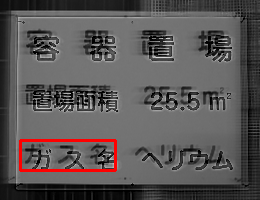}
   \includegraphics[width=0.15\linewidth]{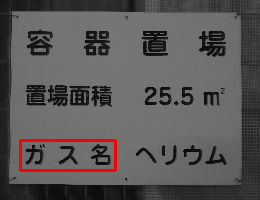}}\\[-2.1ex]
      
   \setcounter{subfigure}{0}
      
   \subfloat[Blurred IMG]{
   \includegraphics[width=0.15\linewidth]{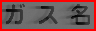}}
   \subfloat[Noisy IMG]{
   \includegraphics[width=0.15\linewidth]{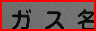}}
  \subfloat[SBD-single \cite{zhang2013multi}] {
   \includegraphics[width=0.15\linewidth]{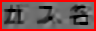}}
   \subfloat[SBD-multi \cite{zhang2013multi}]{
   \includegraphics[width=0.15\linewidth]{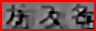}}
    \subfloat[GCRL+DL \cite{yuan2007image}]{
   \includegraphics[width=0.15\linewidth]{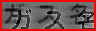}}
   \subfloat[OGMM+DL]{
   \includegraphics[width=0.15\linewidth]{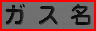}}
   
\end{center}
   \caption{Visual comparison on real-world data \textit{Sign}. (a) Blurred image taken with the shutter speed of 0.5 second and ISO of 100. (b) Noisy image taken with the shutter speed of 0.01 second and ISO of 3200, and further enhanced by gamma correction ($\gamma = 1.5$). 
   The close-up windows in (a) and (b) show different appearances because the two images are taken in different views. }
\label{fig:realData1}
\end{figure*}

\begin{figure*}[htb]
\centering

\subfloat{
   \includegraphics[width=0.15\linewidth]{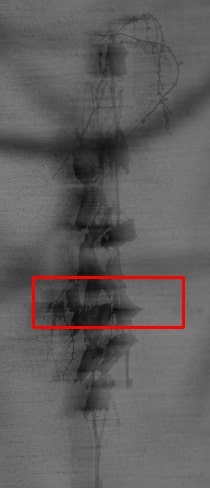}
   \includegraphics[width=0.15\linewidth]{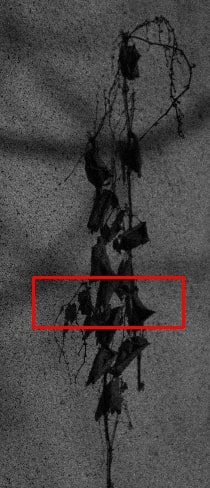}
   \includegraphics[width=0.15\linewidth]{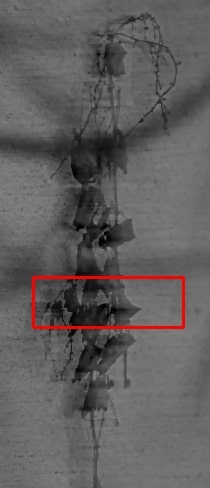}
   \includegraphics[width=0.15\linewidth]{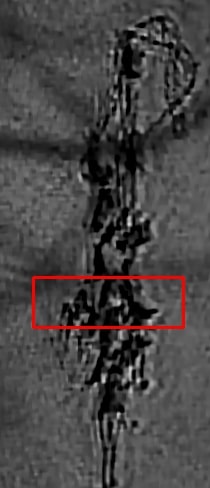}
   \includegraphics[width=0.15\linewidth]{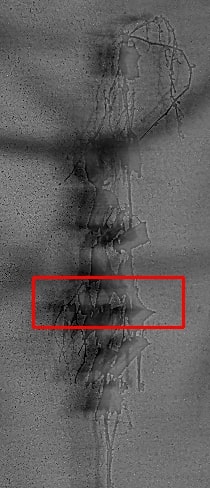}
   \includegraphics[width=0.15\linewidth]{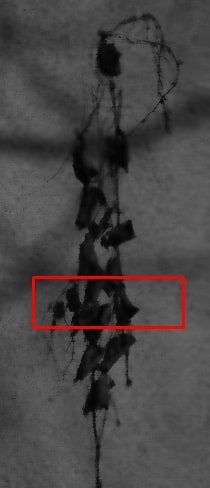}}\\[-2.1ex]
      
   \setcounter{subfigure}{0}
      
   \subfloat[Blurred IMG]{
   \includegraphics[width=0.15\linewidth]{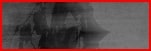}}
   \subfloat[Noisy IMG]{
   \includegraphics[width=0.15\linewidth]{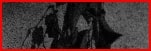}}
  \subfloat[SBD-single \cite{zhang2013multi}] {
   \includegraphics[width=0.15\linewidth]{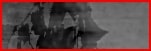}}
   \subfloat[SBD-multi \cite{zhang2013multi}]{
   \includegraphics[width=0.15\linewidth]{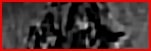}}
    \subfloat[GCRL+DL \cite{yuan2007image}]{
   \includegraphics[width=0.15\linewidth]{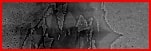}}
   \subfloat[OGMM+DL]{
   \includegraphics[width=0.15\linewidth]{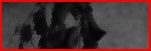}}
   
   \caption{Visual comparison on real-world data \textit{licorice}. (a) Blurred image taken with the shutter speed of 0.5 second and ISO of 100. (b) Noisy image taken with the shutter speed of 0.005 second and ISO of 3200, and further enhanced by gamma correction ($\gamma = 2$).
   The close-up windows in (a) and (b) show different appearances because the two images are taken in different views. }
\label{fig:realData2}
\end{figure*}

\begin{figure*}[htb]
\centering
\subfloat{
   \includegraphics[width=0.15\linewidth]{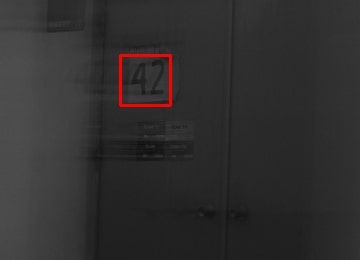}
   \includegraphics[width=0.15\linewidth]{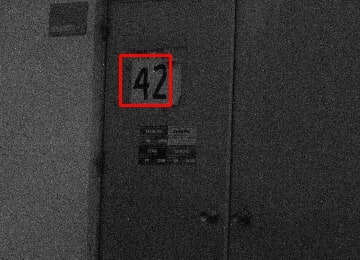}
   \includegraphics[width=0.15\linewidth]{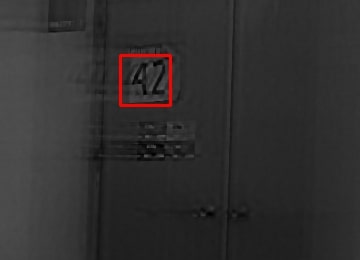}
   \includegraphics[width=0.15\linewidth]{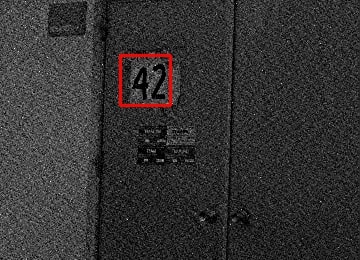}
   \includegraphics[width=0.15\linewidth]{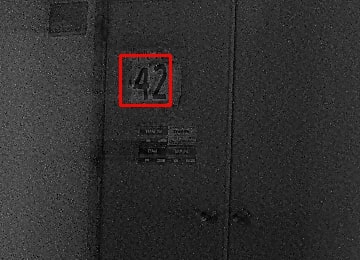}
   \includegraphics[width=0.15\linewidth]{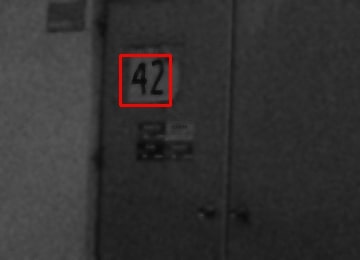}}\\[-2.1ex]
      
   \setcounter{subfigure}{0}
      
   \subfloat[Blurred IMG]{
   \includegraphics[width=0.15\linewidth]{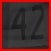}}
   \subfloat[Noisy IMG]{
   \includegraphics[width=0.15\linewidth]{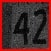}}
  \subfloat[SBD-single \cite{zhang2013multi}] {
   \includegraphics[width=0.15\linewidth]{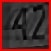}}
   \subfloat[SBD-multi \cite{zhang2013multi}]{
   \includegraphics[width=0.15\linewidth]{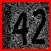}}
    \subfloat[GCRL+DL \cite{yuan2007image}]{
   \includegraphics[width=0.15\linewidth]{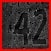}}
   \subfloat[OGMM+DL]{
   \includegraphics[width=0.15\linewidth]{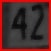}}

   \caption{Visual comparison on real-world data \textit{Gate}. (a) Blurred image taken with the shutter speed of 2 second and ISO of 100. (b) Noisy image taken with the shutter speed of 0.01 second and ISO of 3200, and further enhanced by Gain correction ($\alpha = 2$) and Gamma correction ($\gamma = 2$). 
   The close-up windows in (a) and (b) show different appearances because the two images are taken in different views. }
\label{fig:realData3}
\end{figure*}

\begin{figure*}
\centering
{
\includegraphics[width=0.15\linewidth]{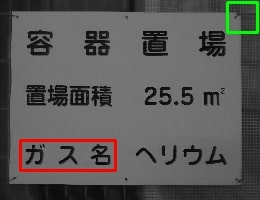}\hspace{0.9em}
\includegraphics[width=0.15\linewidth]{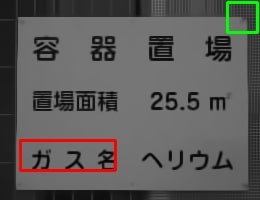}\hspace{0.9em}
\includegraphics[width=0.15\linewidth]{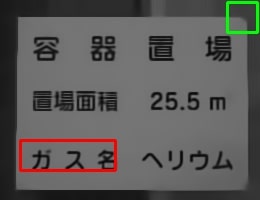}\hspace{0.9em}
\includegraphics[width=0.15\linewidth]{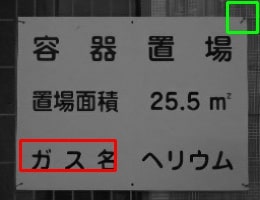}\hspace{0.9em}
\includegraphics[width=0.15\linewidth]{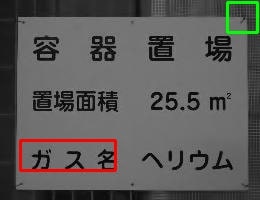}
}\\[-2.1ex]

{\subfloat[OGMM+DL]{
\includegraphics[width=0.15\linewidth]{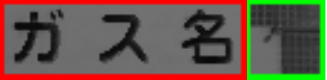}}\hspace{0.6em}
\subfloat[BM3D \cite{dabov2008image}]{
\includegraphics[width=0.15\linewidth]{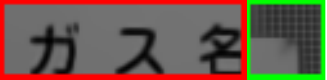}}\hspace{0.5em}
\subfloat[PCLR \cite{chen2015external}]{
\includegraphics[width=0.15\linewidth]{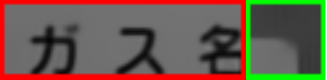}}\hspace{0.5em}
\subfloat[\guSec{MemNet \cite{tai2017memnet}}]{
\includegraphics[width=0.15\linewidth]{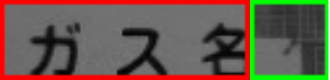}}\hspace{0.5em}
\subfloat[\guSec{CBDNet \cite{guo2019toward}}]{
\includegraphics[width=0.15\linewidth]{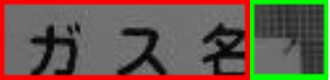}}
}

{
\includegraphics[width=0.15\linewidth]{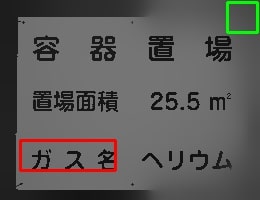}\hspace{0.9em}
\includegraphics[width=0.15\linewidth]{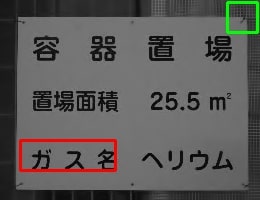}\hspace{0.9em}
\includegraphics[width=0.15\linewidth]{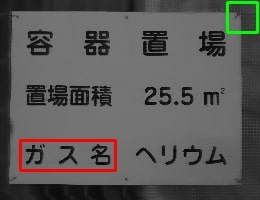}\hspace{0.9em}
\includegraphics[width=0.15\linewidth]{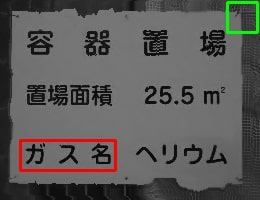}\hspace{0.9em}
\includegraphics[width=0.15\linewidth]{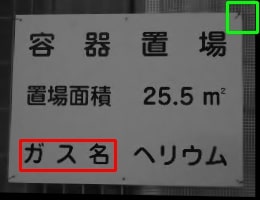}
}\\[-2.1ex]

\subfloat[Bilateral filter \cite{tomasi1998bilateral}]{
\includegraphics[width=0.15\linewidth]{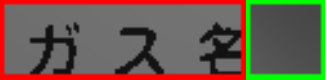}}\hspace{0.6em}
\subfloat[Fast NLM \cite{buades2005non}]{
\includegraphics[width=0.15\linewidth]{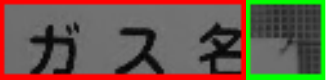}}\hspace{0.5em}
\subfloat[OGMM+DL using \protect\linebreak (g) as noisy input]{
\includegraphics[width=0.15\linewidth]{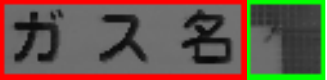}}\hspace{0.5em}
\subfloat[Registration result of (g) with optical flow]{
\includegraphics[width=0.15\linewidth]{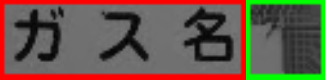}}\hspace{0.5em}
\subfloat[Registration result of (g) with homography]{
\includegraphics[width=0.15\linewidth]{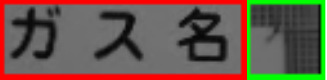}}

\caption{(a) and (h) are the deblurring results of the blurred input (Fig. \ref{fig:realData1}(a)) by OGMM without or with denoising the noisy input Fig. \ref{fig:realData1}(b). \guSec{ (b)$\sim$(g) are the denoised results of Fig. \ref{fig:realData1}(b).} (i) and (j) are the results by aligning (g) with Fig. \ref{fig:realData1}(a), by using two different correspondence estimation methods.}
\label{fig:DenoisingComparison}
\end{figure*}

\begin{figure*}
\centering
{
\includegraphics[width=0.14\linewidth]{figure4experiment/Gate/OGMMDL_rec2.jpg}\hspace{0.9em}
\includegraphics[width=0.14\linewidth]{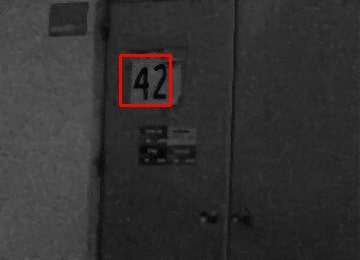}\hspace{0.9em}
\includegraphics[width=0.14\linewidth]{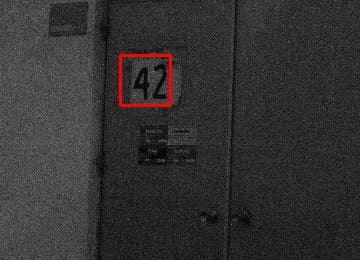}\hspace{0.9em}
\includegraphics[width=0.14\linewidth]{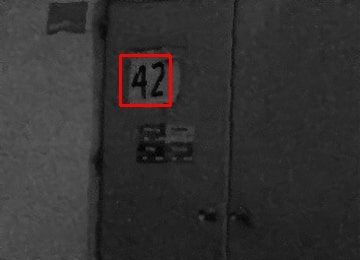}\hspace{0.9em}
\includegraphics[width=0.14\linewidth]{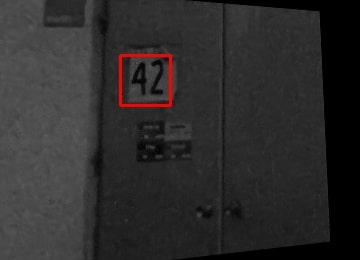}
}\\[-2.1ex]

\subfloat[\guSec{OGMM+DL}]{
\includegraphics[width=0.14\linewidth]{figure4experiment/Gate/OGMMDL_ROI2.jpg}}\hspace{0.5em}
\subfloat[\guSec{BM3D \cite{dabov2008image}}]{
\includegraphics[width=0.14\linewidth]{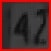}}\hspace{0.5em}
\subfloat[\guSec{PCLR \cite{chen2015external}}]{
\includegraphics[width=0.14\linewidth]{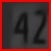}}\hspace{0.5em}
\subfloat[\guSec{MemNet \cite{tai2017memnet}}]{
\includegraphics[width=0.14\linewidth]{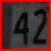}}\hspace{0.5em}
\subfloat[\guSec{CBDNet \cite{guo2019toward}}]{
\includegraphics[width=0.14\linewidth]{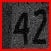}}\hspace{0.4em}

{
\includegraphics[width=0.14\linewidth]{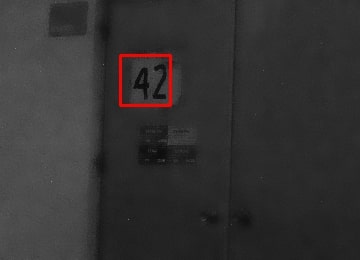}\hspace{0.9em}
\includegraphics[width=0.14\linewidth]{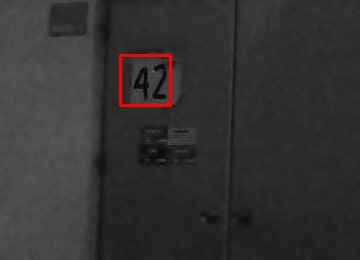}\hspace{0.9em}
\includegraphics[width=0.14\linewidth]{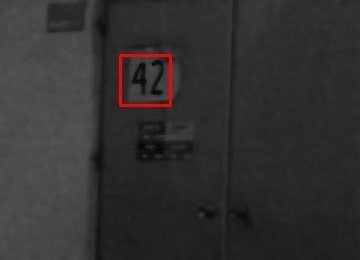}\hspace{0.9em}
\includegraphics[width=0.14\linewidth]{figure4experiment/AdditionalExp/GateRe/GateCorespondenceDenoised_Rec.jpg}\hspace{0.9em}
\includegraphics[width=0.14\linewidth]{figure4experiment/AdditionalExp/GateRe/homography_Memnet_Rec.jpg}
}\\[-2.1ex]

\subfloat[\guSec{Bilateral filter \cite{tomasi1998bilateral}}]{
\includegraphics[width=0.14\linewidth]{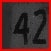}}\hspace{0.5em}
\subfloat[\guSec{Fast NLM \cite{buades2005non}}]{
\includegraphics[width=0.14\linewidth]{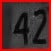}}\hspace{0.5em}
\subfloat[\guSec{OGMM+DL using \protect\linebreak (d) as noisy input}]{
\includegraphics[width=0.14\linewidth]{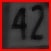}}\hspace{0.5em}
\subfloat[\guSec{Registration result of (d) with optical flow}]{
\includegraphics[width=0.14\linewidth]{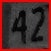}}\hspace{0.5em}
\subfloat[\guSec{Registration result of (d) with homography}]{
\includegraphics[width=0.14\linewidth]{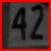}}\hspace{0.4em}

\caption{\guSec{(a) and (h) are the deblurring results of the blurred input (Fig. \ref{fig:realData3}(a)) by OGMM without or with denoising the noisy input Fig. \ref{fig:realData3}(b). (b)$\sim$(g) are the denoised results of Fig. \ref{fig:realData3}(b). (i) and (j) are the results by aligning (d) with Fig. \ref{fig:realData3}(a), by using two different correspondence estimation methods.}}
\label{fig:DenoisingComparison2}
\end{figure*}

\begin{figure*}[htb]
\centering
\subfloat{
   \includegraphics[width=0.2\linewidth]{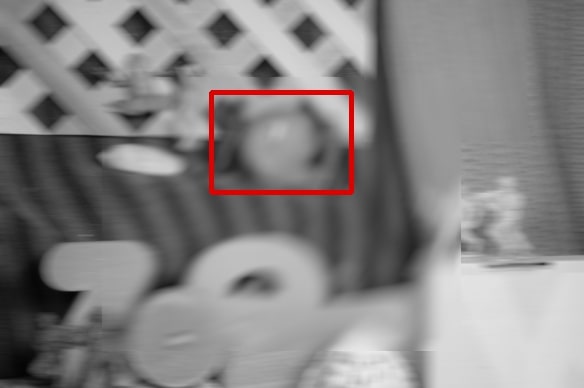}
   \includegraphics[width=0.15\linewidth]{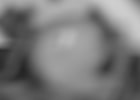}
   \includegraphics[width=0.15\linewidth]{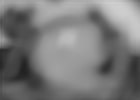}
   \includegraphics[width=0.15\linewidth]{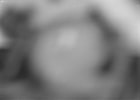}
   \includegraphics[width=0.15\linewidth]{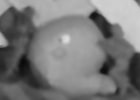}
   }\\[-2.1ex]
      
   \setcounter{subfigure}{0}
      
   \subfloat[\gu{Blurred IMG}]{
   \includegraphics[width=0.2\linewidth]{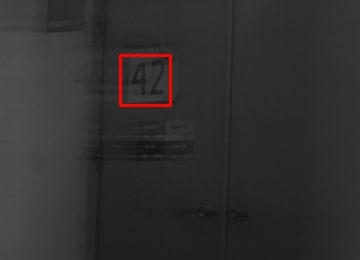}}
   \subfloat[\gu{Blurred area}]{
   \includegraphics[width=0.15\linewidth]{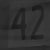}}
  \subfloat[\gu{DMC \cite{nah2017deep}}] {
   \includegraphics[width=0.15\linewidth]{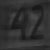}}
   \subfloat[\gu{DeblurGAN \cite{kupyn2018deblurgan}}]{
   \includegraphics[width=0.15\linewidth]{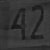}}
   \subfloat[\gu{OGMM+DL}]{
   \includegraphics[width=0.15\linewidth]{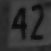}}

   \caption{\gu{Visual comparison with deep-learning based deblurring methods on both synthetic data \textit{Army} (row 1) and real-world data \textit{Gate} (row 2). Close-up views are shown for examining the details. }}
\label{fig:DeepComparison}
\end{figure*}

\begin{figure*}[tb]
\centering
   \setcounter{subfigure}{0}
      
   \subfloat[Blurred IMG]{
   \includegraphics[width=0.16\linewidth]{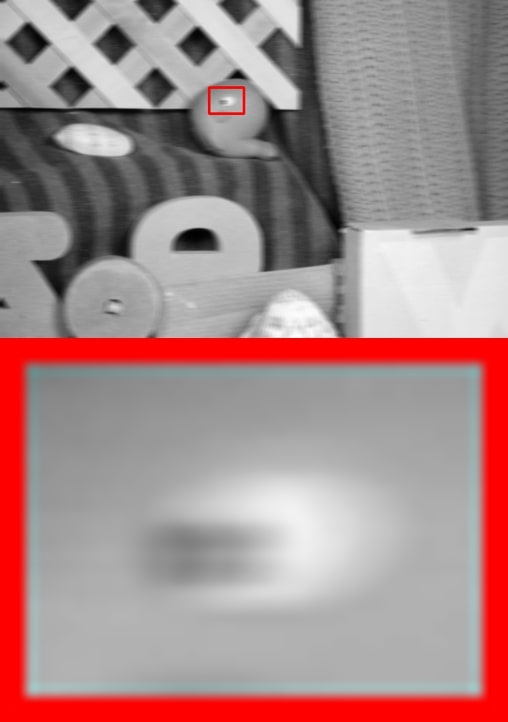}}\hspace{1.0em}%
   \subfloat[Noisy IMG ]{
   \includegraphics[width=0.16\linewidth]{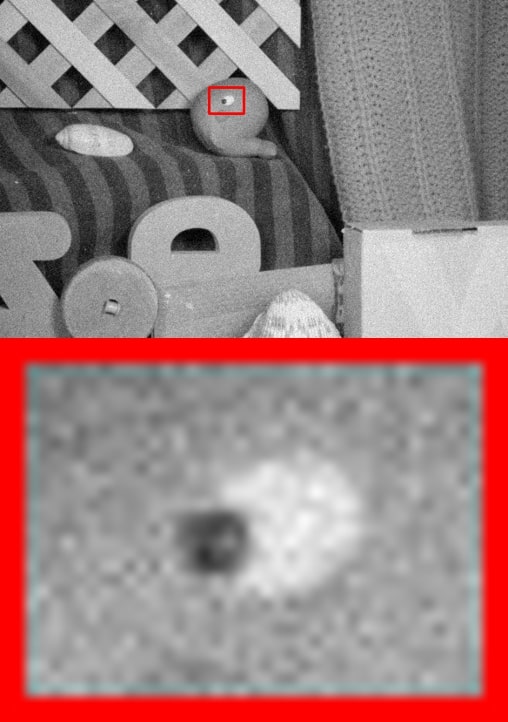}}\hspace{1.0em}%
  \subfloat[Deblurred IMG ($t=1$)] {
   \includegraphics[width=0.16\linewidth]{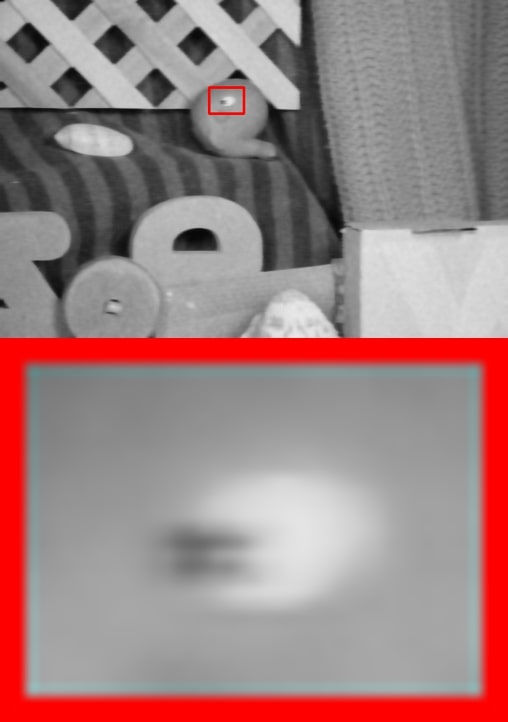}}\hspace{1.0em}%
   \subfloat[Deblurred IMG ($t=2$)]{
   \includegraphics[width=0.16\linewidth]{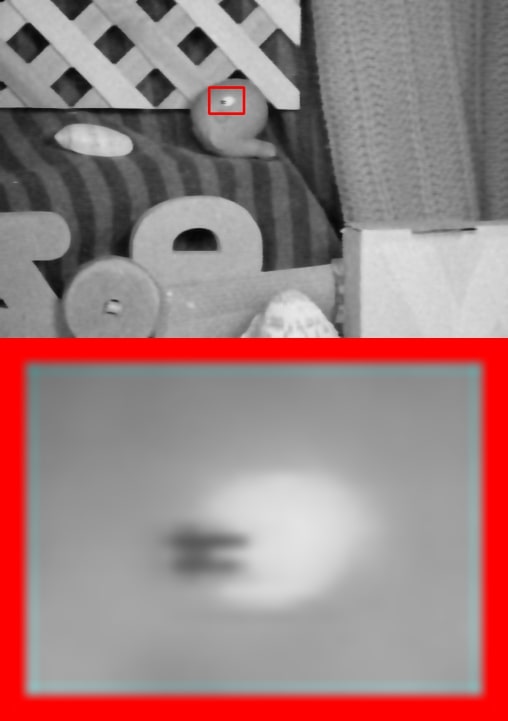}}\hspace{1.0em}%
    \subfloat[Deblurred IMG ($t=6$)]{
   \includegraphics[width=0.16\linewidth]{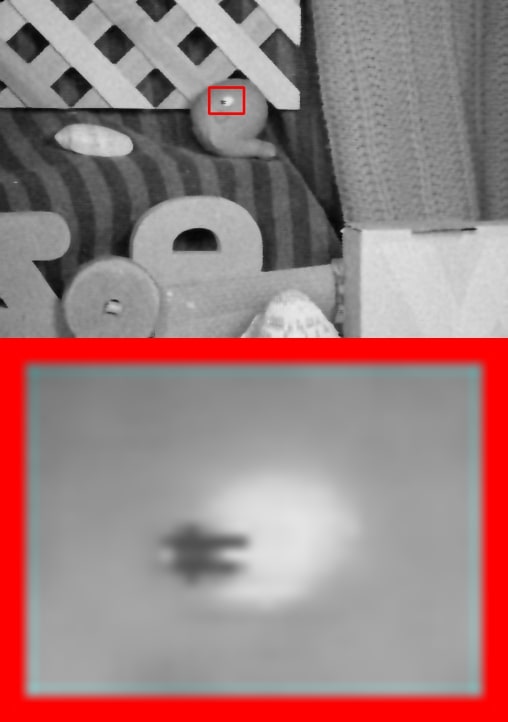}}\hspace{1.0em}%
    
   \subfloat[Ground truth optical flow between original (a) and (b)]{
   \includegraphics[width=0.16\linewidth]{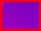}}\hspace{1.0em}%
   \subfloat[Optical Flow calculated between (a) and (b) \protect\linebreak AAE=0.87, AEE=0.60]{
   \includegraphics[width=0.16\linewidth]{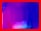}}\hspace{1.0em}%
   \subfloat[Optical flow calculated between (c) and (b) \protect\linebreak AAE=0.80, AEE=0.52]{
   \includegraphics[width=0.16\linewidth]{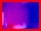}}\hspace{1.0em}%
  \subfloat[Optical flow calculated between (d) and (b) \protect\linebreak AAE=0.79, AEE=0.51] {
   \includegraphics[width=0.16\linewidth]{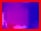}}\hspace{1.0em}%
   \subfloat[Optical flow calculated between (e) and (b) \protect\linebreak AAE=0.77, AEE=0.50]{
   \includegraphics[width=0.16\linewidth]{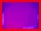}}\hspace{1.0em}%

   \caption{Qualitative and quantitative analysis of the optical flow on image \textit{RubberWhale} from dataset \cite{baker2011database}. (a) is with the synthetic blur \textit{BlurType1} in Fig. \ref{fig:BlurVisualization}.  (c)$\sim$(d) are deblurring results with respect to different numbers of iterations. (f)$\sim$(j) show the visualization of optical flow of the ROI area between the deblurring result and the noisy image using the color flow \cite{baker2011database}. The update of the optical flow is introduced in Sec. \ref{sec:opticalFLowUpdate}. Also, both the average absolute flow endpoint error (AEE) in pixel and the average angular error (AAE) in radian are shown with respect to the ground-truth flow. }
\label{fig:OpticalFlowEva}
\end{figure*}

\begin{figure}[htb]
	\centering
\setcounter{subfigure}{0}
\subfloat{
\includegraphics[width=25mm,scale=0.5]{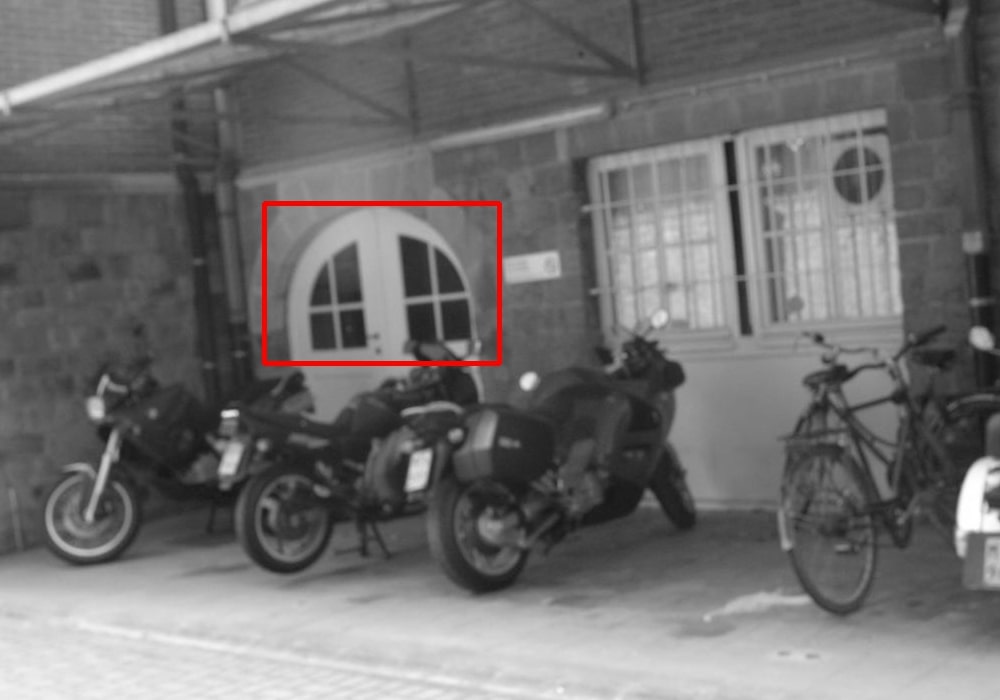}
\includegraphics[width=25mm,scale=0.5]{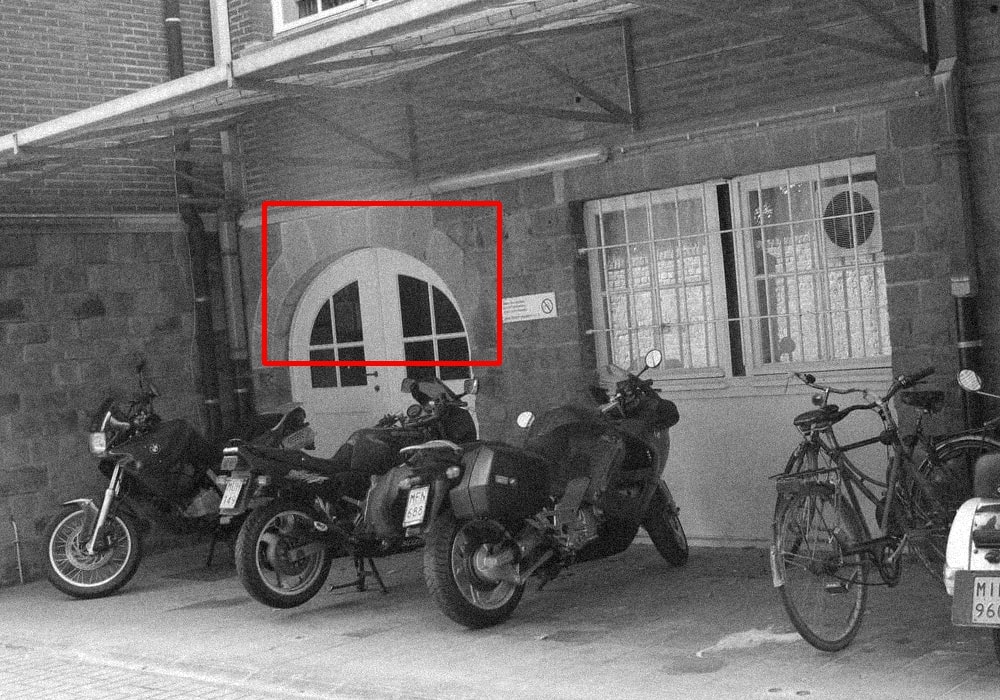}
\includegraphics[width=25mm,scale=0.5]{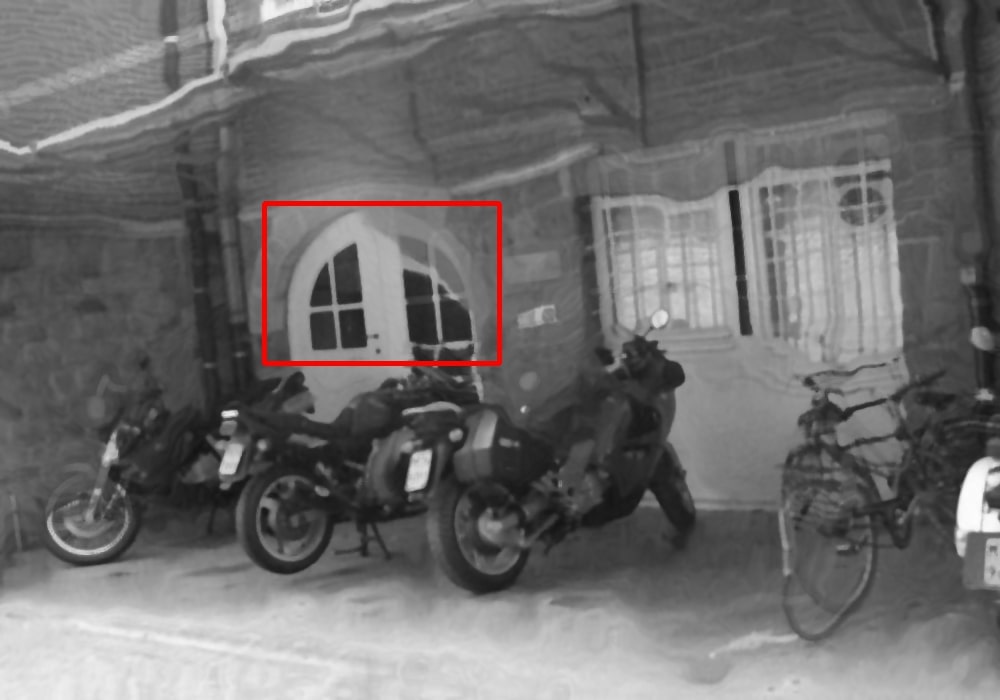}
}\\[-2.1ex]

\setcounter{subfigure}{0}
\subfloat[Blurred image]{
\includegraphics[width=25mm,scale=0.2]{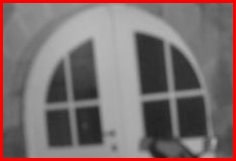}}
\subfloat[Noisy image]{
\includegraphics[width=25mm,scale=0.2]{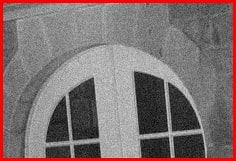}}
\subfloat[Deblurred (a)]{
\includegraphics[width=25mm,scale=0.2]{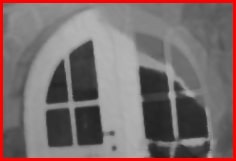}}

\caption{A failure example. A large view difference deteriorates the accuracy of optical flow, thus leading to a distorted result. }
\label{fig:Limitation}
\end{figure}

\section{Experimental Results}
We evaluate our approach on both synthetic data and real-world data. Quantitative comparisons on synthetic data with ground truth are also carried out. Besides traditional methods \gu{\cite{richardson1972bayesian,holmes1995light,whyte2012non,kheradmand2014general,bai2018graph,zhang2013multi,yuan2007image}}, we also evaluate the performance of our model by comparing it with the state-of-the-art deep learning based deblurring methods \cite{nah2017deep,kupyn2018deblurgan}. Additionally, since our method depends on optical flow, we perform an analysis for it.

\subsection{Synthetic Data}
We first assess the performance of some current deblurring methods and our approach on ten image pairs from the publicly available dataset \cite{baker2011database}. The dataset consists of multiple pairs of images taken from two different views in various scenes. To demonstrate the robustness of our method to different blur models, we synthetically generate six types of blur: (1) linear motion blur, (2) circular motion blur, (3) the mixture of circular motion blur, linear motion blur and Gaussian blur, (4) the mixture of two types of linear motion blur and circular motion blur, (5) the mixture of circular motion blur, zoom motion blur and two types of linear motion blur, and (6) the mixture of two types of linear motion blur and circular motion blur. The visualization of each type of blur is shown in Fig. \ref{fig:BlurVisualization}. The first image in each pair is blurred with these six types of blur, respectively. Gaussian noise is added to the second image to generate the noisy image.  

We compare our approach with the deblurring methods \cite{zhang2013multi,yuan2007image} which can also take a pair of such images as input. \textit{To our knowledge, deblurring using a pair of blurred/noisy images has been sparsely treated so far, and the method \cite{yuan2007image} is \guSec{highly close} to ours}. We also compare our method with \gu{five} single image deblurring methods \gu{\cite{kheradmand2014general, bai2018graph}},\cite{whyte2012non, richardson1972bayesian, holmes1995light}, including two baseline methods \cite{richardson1972bayesian, holmes1995light}. Note that the results of the single image blind deblurring methods are only reference results (produced with the available programs provided by authors), and they do not take advantage of the second image as input. \guThi{In addition, we include the comparison against the pipeline of denoising and further alignment strategies \cite{farneback2003two, buades2005non, artin2016geometric} which can alternatively generate results from the view of the blurred image (see Sec. \ref{sec:Real-World Data} for more information). } 

As suggested by previous works \cite{vankawala2015survey, vasu2018non}, we compute three metrics (in an average sense) for quantitative comparisons: peak signal-to-noise ratio (PSNR), structural similarity (SSIM) and mean square error (MSE). Tab. \ref{syntheticTable} displays the PSNR, SSIM and MSE, which are calculated between the deblurred images and the corresponding ground-truth image. We can see from Tab. \ref{syntheticTable} that our approach is more accurate than the two baseline methods \cite{richardson1972bayesian, holmes1995light} and the state-of-the-art techniques \cite{zhang2013multi,yuan2007image,whyte2012non},\gu{\cite{kheradmand2014general, bai2018graph}}. Fig. \ref{fig:ComplexBlurComparison1} to Fig. \ref{fig:ComplexBlurComparison4} show the deblurring results under the mixup of different types of blur, by different deblurring methods. Notice that the close-up views focus on the blurred areas that are mixed by at least two kinds of blur. We mainly discuss the results by \cite{zhang2013multi,whyte2012non,yuan2007image},\gu{\cite{kheradmand2014general, bai2018graph}} which are typically superior to \cite{richardson1972bayesian,holmes1995light}. \guThi{Fig. \ref{fig:ComplexBlurComparison1}(h,i),  Fig. \ref{fig:ComplexBlurComparison2}(h,i),  Fig. \ref{fig:ComplexBlurComparison3}(h,i) and  Fig. \ref{fig:ComplexBlurComparison4}(h,i) respectively show two types of registration results (Registration-1 and Registration-2) between the blurred images (Fig. \ref{fig:ComplexBlurComparison1}(a),  Fig. \ref{fig:ComplexBlurComparison2}(a),  Fig. \ref{fig:ComplexBlurComparison3}(a) and  Fig. \ref{fig:ComplexBlurComparison4}(a)) and the denoised versions of the noisy images (Fig. \ref{fig:ComplexBlurComparison1}(b),  Fig. \ref{fig:ComplexBlurComparison2}(b),  Fig. \ref{fig:ComplexBlurComparison3}(b) and  Fig. \ref{fig:ComplexBlurComparison4}(b)) with \cite{buades2005non}, using dense optical flow (DOF) \cite{farneback2003two} and homography \cite{artin2016geometric}. 
}

\gu{\textbf{Single image \cite{kheradmand2014general, bai2018graph}}.}
\gu{The method of \cite{kheradmand2014general} is a general kernel similarity-based non-blind single image deblurring approach which states that nonlinear motion blur can be addressed. Its deblurring results, presented in Fig. \ref{fig:ComplexBlurComparison1}(f), Fig. \ref{fig:ComplexBlurComparison2}(f), Fig. \ref{fig:ComplexBlurComparison3}(f), and Fig. \ref{fig:ComplexBlurComparison4}(f), contain a large amount of ringing artifacts, which is due to the difficulty in correctly estimating its kernel similarity matrix for highly complex blur kernels. A similar issue occurs in the method of \cite{bai2018graph}, which is a blind deblurring approach designed to deal with uniform blur via a combined regularization using re-weighted graph TV priors. Obviously, the mixup of different types of blurs is beyond its capacity. Tab. \ref{syntheticTable} reports the involved metric evaluation. }

\textbf{Single/Multiple images \cite{zhang2013multi}.} The method \cite{zhang2013multi} can handle both single image and multiple images, and the authors suggest that multiple input images can estimate the blur kernel more accurately. We thus show two versions of this method for comparison. \gu{Fig. \ref{fig:ComplexBlurComparison1}(j), Fig. \ref{fig:ComplexBlurComparison2}(j), Fig. \ref{fig:ComplexBlurComparison3}(j), and Fig. \ref{fig:ComplexBlurComparison4}(j)} show the version of single image deblurring results of method \cite{zhang2013multi}. Mixup of the space-variant and the space-invariant blur hinders the algorithm from estimating correct blur kernel, which is directly reflected by the blurry result and low PSNR values. This also accounts for that the single image deblurring version of \cite{zhang2013multi} performs well on the linear motion blur case, but fails to deal with the remaining five types of non-uniform blur. For the version with multiple input images, despite the fact that the method \cite{zhang2013multi} puts emphasis on automatically distinguishing blurred images from noisy images, it tends to mistake the noisy image for the blurred image and conduct deblurring to the noisy image (\gu{Fig. \ref{fig:ComplexBlurComparison1}(k), Fig. \ref{fig:ComplexBlurComparison2}(k), Fig. \ref{fig:ComplexBlurComparison3}(k), and Fig. \ref{fig:ComplexBlurComparison4}(k)}). This is because the proposed coupled penalty function in \cite{zhang2013multi} judges that the intense blur leads to a higher degradation of the blurred image than the noisy image, thus unwillingly treating the noisy image as the dominant image. 

\textbf{Non-uniform blur \cite{whyte2012non}.} The method of \cite{whyte2012non} stresses its advantage in handling non-uniform blur. Comparing to method \cite{zhang2013multi}, it produces clearer results and higher PSNR values. However, the mixup of blurs can cause obvious discontinuities, especially around the edges of dotted rectangles shown in Fig. \ref{fig:BlurVisualization}, which increases the difficulty in estimating the blur kernel. For example, the close-up view in Fig. \ref{fig:ComplexBlurComparison3}(e) shows the deblurring results close to the border of two different types of blurs. It is obvious that the border still clearly exists after deblurring by method \cite{whyte2012non}, which confirms the limitation.

\textbf{Blurred/noisy images pairs \cite{yuan2007image}.} The method of \cite{yuan2007image} utilizes paired blurred/noisy images for deblurring, requiring the same view for both images. As can be observed from \gu{Fig. \ref{fig:ComplexBlurComparison1}(l)}, the result of \cite{yuan2007image} mixes both appearances of the blurred and the noisy images since the noisy image dominates the final result, which makes the deblurring result appears quite similar to the noisy image in \gu{Fig. \ref{fig:ComplexBlurComparison1}(l)}. This may further result in ``ghost area'' when the difference of the capturing view angle gets larger (\gu{Fig. \ref{fig:realData1}(l)} and \gu{Fig. \ref{fig:realData2}(l)}). As a result, it has low PSNR and SSIM values, reflected in Tab. \ref{syntheticTable}.

Our method, in contrast, requiring no further adjustment or extra kernel modeling with respect to the blur types and the border-near areas, is able to generate more visually pleasing results with higher PSNR values, which shows robustness against complex blur. Also, it is worth pointing out that the final result can be further improved by adding a detail layer onto the output of OGMM, as shown in \gu{Fig. \ref{fig:ComplexBlurComparison1}(n), Fig. \ref{fig:ComplexBlurComparison2}(n), Fig. \ref{fig:ComplexBlurComparison3}(n), and Fig. \ref{fig:ComplexBlurComparison4}(n)}.

\subsection{Real-World Data}
\label{sec:Real-World Data}
We test our approach on various kinds of blurred/noisy image pairs which are captured in low light environments using an off-the-shelf camera. Also, we compare our method with the state-of-the-art techniques \cite{zhang2013multi,yuan2007image,kupyn2018deblurgan}. 

We adopt the following procedure to take a real-world photo pair. First, we set a low ISO and a low shutter speed to obtain the blurred image. In the process of capturing, we add a camera shake, or move the object on purpose to produce stronger blur. Secondly, we use a high ISO and a high shutter speed to obtain the noisy image. Different from the synthetic data, the captured noisy images are too dark to use directly. Before deblurring, the noisy image is enhanced by synchronizing its brightness with the blurred image. The enhancement is achieved via gain/bias change and gamma correction, which also amplifies noise. 

Fig. \ref{fig:realData1} to Fig. \ref{fig:realData3} exhibit visual comparisons on real-world data. 
The blur kernels estimated by \cite{zhang2013multi}, using single or multiple images, have difficulty in recovering the sharp image. The method \cite{yuan2007image} requires the same capturing view for the blurred/noisy image pair, which limits their applicability. As presented in the close-up view, it can be easily observed that heavy misalignment occurs when adding their generated detail layer back.
The result by our method, without the need of kernel estimation, enjoys significantly better visual quality than those by the state-of-the-art methods \cite{zhang2013multi,yuan2007image}. Also, in the case of relatively large view difference, our method bridges two images from different view angles and can correctly correspond the patches in the blurred and noisy images. The deblurred result has the same view with the blurred input.

\guSec{\textbf{Discussion on alternative denoising and alignment.} }
Applying denoising and further alignment algorithms on the noisy image of the pair is an alternative for achieving a clear image corresponding to the blurred image. In Fig. \ref{fig:DenoisingComparison} \guSec{and Fig. \ref{fig:DenoisingComparison2}}, we show some visual results for some well-known denoising methods \guSec{\cite{buades2005non, chen2015external,dabov2008image,tomasi1998bilateral,tai2017memnet,guo2019toward}}, and registration/alignment techniques \cite{rublee2011orb,farneback2003two}. As can be observed from Fig. \ref{fig:DenoisingComparison}(g) \guSec{ and Fig. \ref{fig:DenoisingComparison2}(g)}, the method \cite{buades2005non} induces ``mosaics'' around the character edges. In Fig. \ref{fig:DenoisingComparison}(b,c,f) \guSec{and Fig. \ref{fig:DenoisingComparison2}(b,c)}, methods \cite{chen2015external,dabov2008image,tomasi1998bilateral} suffer from the same issue that the fine structures near the edges 
are smoothed out, which is one potential phenomenon of traditional denoising methods. \guSec{The methods \cite{tai2017memnet,guo2019toward} are two state-of-the-art deep learning based denoising approaches. The method \cite{tai2017memnet} builds dense connections among several types of memory blocks. 
The method \cite{guo2019toward} introduced a sub-network with asymmetric and total variation (TV) losses to specifically deal with real-world situations such as noise arising from in-camera processing. It can be clearly observed in Fig. \ref{fig:DenoisingComparison}(d) and Fig. \ref{fig:DenoisingComparison}(e) that they perform well on the areas with plain textures; however, noise around the context areas remains unresolved (please enlarge the corresponding areas for clear observations). In addition, both methods fail to handle intense real-world noise (Fig. \ref{fig:DenoisingComparison2}(d,e)). }

As shown in Fig. \ref{fig:DenoisingComparison}(h) \guSec{ (green box) and Fig. \ref{fig:DenoisingComparison2}(h)}, in our method, denoising the noisy input inevitably leads to \guSec{over-smoothness.} Moreover, the view angle of the denoised image needs to be aligned with the blurred input. \guSec{Fig. \ref{fig:DenoisingComparison}(i,j) and Fig. \ref{fig:DenoisingComparison2}(i,j)} are the results aligned by the correspondence provided by optical flow and homography \cite{artin2016geometric}, respectively. 
Obviously, these final ``deblurred'' results involve defects, such as distortion \guSec{(Fig. \ref{fig:DenoisingComparison}(j) and Fig. \ref{fig:DenoisingComparison2}(j)), misalignments (Fig. \ref{fig:DenoisingComparison}(i) and Fig. \ref{fig:DenoisingComparison2}(i)) and remaining noise  (Fig. \ref{fig:DenoisingComparison2}(i,j))}. This is because the two input blurred and noisy images have limited quality and cannot provide sufficiently accurate correspondence estimation for alignment.

\gu{\subsection{Comparison with Deep Learning based Methods}
We also compare our approach with two major deep learning based deblurring methods \cite{nah2017deep,kupyn2018deblurgan} which claimed their capability in dealing with complex blur with suggested parameters and trained models.
We show the comparison results  in Fig. \ref{fig:DeepComparison}. } 

\gu{Nah et al. developed an end-to-end, multi-scale CNN for recovering sharp images without blur kernel estimation \cite{nah2017deep}. Yet, its results in Fig. \ref{fig:DeepComparison}(c) seem to be less visually pleasant, since blur is only partially removed. The method of \cite{kupyn2018deblurgan} resorts to a conditional adversarial network which is optimized using a
multi-component loss function to free kernel estimation. As shown in Fig. \ref{fig:DeepComparison}(d), it can hardly remove the strong blur with a single blurred image as the input, in both cases of synthetic complex blur and real-world blur. }

\subsection{Analysis on Guiding Optical Flow}
\gu{This subsection reports qualitative and quantitative analysis on guiding optical flow.
In our implementation, we simply adopt Farneback's original dense optical flow \cite{farneback2003two}, which performs well due to small change of viewing angle between the input image pair. }

Optical flow between the deblurred result and the noisy image is supposed to be refined iteratively. Fig. \ref{fig:OpticalFlowEva} shows an example of the update of optical flow along with the deblurring in Fig. \ref{fig:introduction}. In Fig. \ref{fig:OpticalFlowEva}(f), the ground-truth flow between the original Fig. \ref{fig:OpticalFlowEva}(a) without blur and Fig. \ref{fig:OpticalFlowEva}(b) without noise is taken as an evaluation basis. Fig. \ref{fig:OpticalFlowEva}(g)$\sim$(j) show the optical flow calculated between the noisy image and the deblurred result with respect to different iteration times. As can be observed, with the increase of iteration times, the EM optimization iteratively removes the blur, and the optical flow shows clear improvement. Moreover, we show quantitative results in terms of the average absolute flow endpoint error (AEE) \cite{otte199estimation} and the average angular error (AAE) \cite{barron1994performance}, with respect to the ground-truth flow. Both the errors decrease along with the increase of the iteration numbers. It demonstrates that optical flow acts as a positive role in deblurring, and in return the deblurred image can also improve optical flow.

\section{Conclusion}
We proposed a novel, robust image deblurring method with the use of a pair of blurred/noisy images. Our approach first builds patch correspondences between the blurred and noisy images, and then relates the latent pixel intensities with the noisy pixel intensities under the GMM framework. We  introduced a bilateral term for better features preservation. To refine the deblurred result, we extract and add a detail layer to it. Our approach is free of blur kernel estimation and robust to various types of blur. Extensive experiments over the synthetic and real-world data demonstrate that our method outperforms state-of-the-art techniques, in terms of both visual quality and quantity.

The major limitation of our method is its dependency on optical flow. If the motion gap between the two images is too large, the accuracy of optical flow deteriorates. As a result, this would alter object appearance or reshape some sharp features, as illustrated in Fig. \ref{fig:Limitation}. In the future, we would like to exploit more useful and effective relationship among patches to address the issue of undesired optical flow.


\ifCLASSOPTIONcaptionsoff
  \newpage
\fi

\bibliographystyle{abbrv}
\bibliography{reference}

\begin{IEEEbiography}[{\includegraphics[width=1in,height=1.25in,clip,keepaspectratio]{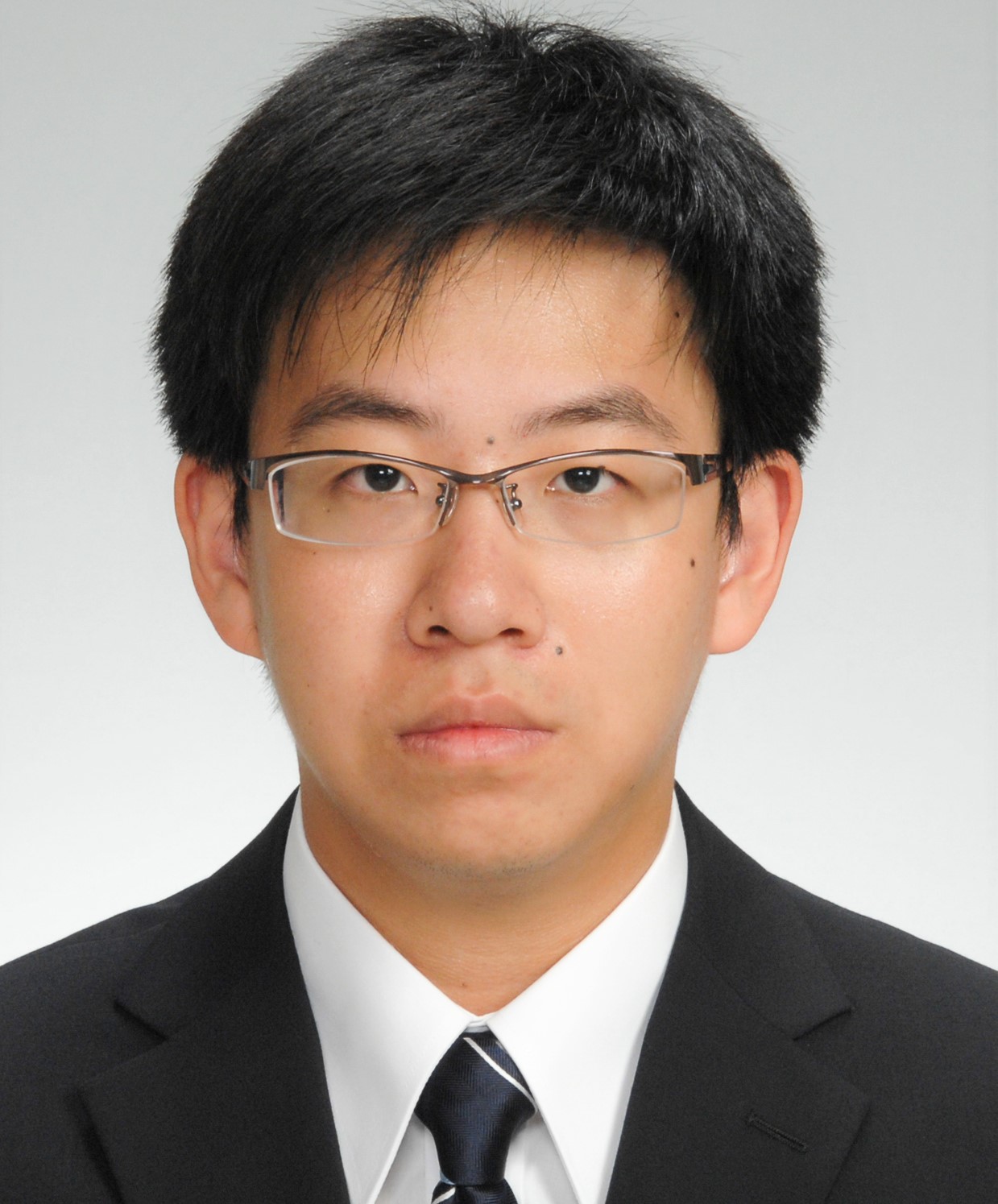}}]{Chunzhi Gu}
Received his B.E. degree in June 2018 at University of Shanghai for Science and Technology (Shanghai, China), and is now pursuing Ph.D. degree at University of Fukui (Fukui, Japan) since April 2018. His current research focuses on modeling visual computing problems with probabilistic models, followed by optimization methods to estimate the parameters.
\end{IEEEbiography}

\begin{IEEEbiography}[{\includegraphics[width=1in,height=1.25in,clip,keepaspectratio]{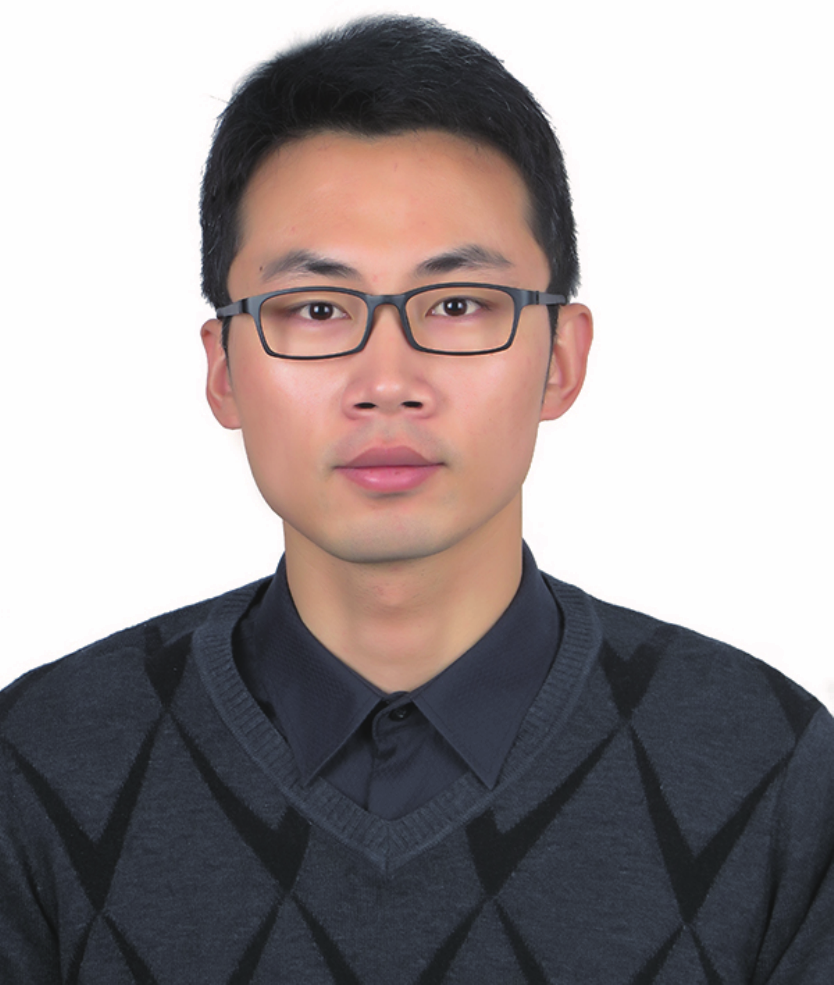}}]{Xuequan Lu}
is a Lecturer (Assistant Professor) at Deakin University, Australia. He spent more than two years as a Research Fellow in Singapore. Prior to that, he earned his Ph.D at Zhejiang University (China) in June 2016. His research interests mainly fall into the category of visual computing, for example, geometry modeling, processing and analysis, animation/simulation, 2D data processing and analysis. More information can be found at \url{http://www.xuequanlu.com}.
\end{IEEEbiography}

\begin{IEEEbiography}[{\includegraphics[width=1in,height=1.25in,clip,keepaspectratio]{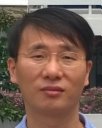}}]{Ying He}
is an associate professor at School of Computer Science and Engineering, Nanyang Technological University, Singapore. He received the BS and MS degrees in electrical engineering from Tsinghua University, China, and the PhD degree in computer science from Stony Brook University, USA. His research interests fall into the general areas of visual computing and he is particularly interested in the problems which require geometric analysis and computation. For more information, visit http://www.ntu.edu.sg/home/yhe/
\end{IEEEbiography}

\begin{IEEEbiography}[{\includegraphics[width=1in,height=1.25in,clip,keepaspectratio]{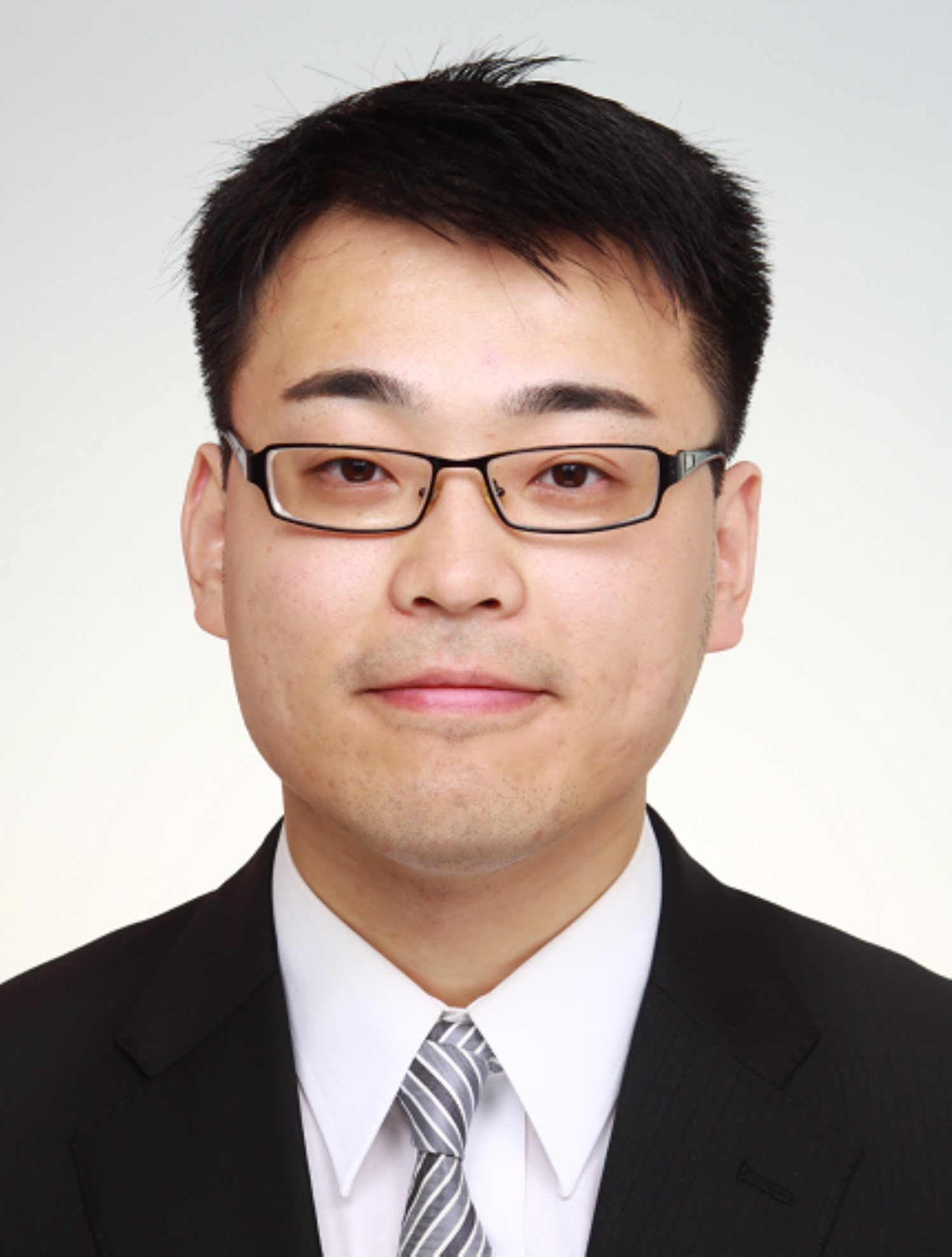}}]{Chao Zhang}
Received his Ph.D. degree at Iwate university (Japan) in March 2017. He is now a full-time assistant professor at department of engineering, university of Fukui (Japan) since April 2017. His research interests include computer vision , computer graphics, and evolutionary computing, mainly focused on applying optimization methods to solve visual computing problems.
\end{IEEEbiography}
\vfill 




\end{document}